\documentclass{article}

\PassOptionsToPackage{table,dvipsnames}{xcolor}
\PassOptionsToPackage{numbers,sort&compress}{natbib}

\usepackage{PRIMEarxiv}

\ifdefined\XeTeXversion
  \usepackage{fontspec}
\else
  \usepackage[utf8]{inputenc}
  \usepackage[T1]{fontenc}
\fi

\usepackage{xcolor}
\usepackage{colortbl}

\usepackage{amsmath,amsfonts,bm}

\def\eqref#1{equation~\ref{#1}}

\def\1{\bm{1}}

\DeclareMathAlphabet{\mathsfit}{\encodingdefault}{\sfdefault}{m}{sl}
\SetMathAlphabet{\mathsfit}{bold}{\encodingdefault}{\sfdefault}{bx}{n}

\usepackage{amsmath}
\usepackage{mathalpha}
\usepackage{isomath}        \usepackage{amsfonts}
\usepackage{amssymb}        \usepackage{amsthm}
\usepackage{nicefrac}

\newtheorem{proposition}{Proposition}

\usepackage{natbib}

\usepackage{graphicx}
\usepackage{pgfplots}
\usepackage{placeins}
\pgfplotsset{compat=1.18}
\usepgfplotslibrary{groupplots}
\usepackage{tikz}
\usetikzlibrary{arrows.meta, decorations.pathmorphing, calc, positioning}
\usepackage{subcaption}
\usepackage{rotating}

\usepackage{booktabs}
\usepackage{tabularx}
\usepackage{longtable}
\usepackage{multirow}
\usepackage{array}
\usepackage{threeparttable}
\usepackage{siunitx}
\usepackage{enumitem}
\usepackage[normalem]{ulem}
\usepackage{comment}
\usepackage{url}
\usepackage{microtype}
\ifdefined\XeTeXversion
  
  \microtypesetup{protrusion=false}
\fi
\ifdefined\XeTeXversion
  \usepackage{xeCJK}
  \IfFontExistsTF{Noto Sans CJK SC}{%
    \setCJKmainfont{Noto Sans CJK SC}%
    \setCJKfamilyfont{mj}{Noto Sans CJK SC}%
    \setCJKsansfont{Noto Sans CJK SC}%
    \setCJKmonofont{Noto Sans CJK SC}%
  }{%
    \setCJKmainfont{FandolSong-Regular.otf}%
    \setCJKsansfont{FandolHei-Regular.otf}%
    \setCJKmonofont{FandolHei-Regular.otf}%
    \IfFontExistsTF{UnBatang.ttf}{\setCJKfamilyfont{mj}{UnBatang.ttf}}%
      {\setCJKfamilyfont{mj}{batang.ttf}}%
  }
  \newenvironment{CJK*}[2]{}{}
\else
  \usepackage{CJKutf8}
\fi
\usepackage{listings}

\definecolor{codegreen}{rgb}{0,0.6,0}
\definecolor{codegray}{rgb}{0.5,0.5,0.5}
\definecolor{codepurple}{rgb}{0.58,0,0.82}
\definecolor{backcolour}{rgb}{0.97,0.97,0.97}

\usepackage{hyperref}
\hypersetup{
  pdftitle={Every Ablation Is a Dose: Counterweights and the Semblance of Self-Repair},
  pdfkeywords={mechanistic interpretability, self-repair, ablation, causal intervention, counterweights}
}

\definecolor{TrueC}{HTML}{1B7F3B}
\definecolor{FalseC}{HTML}{B3261E}
\definecolor{NullC}{HTML}{6B7280}
\definecolor{Junk}{HTML}{808080}
\newcommand{\tTr}[1]{\textcolor{TrueC}{\textbf{#1}}}
\newcommand{\tFa}[1]{\textcolor{FalseC}{\textbf{#1}}}

  \newcommand{\cjkjp}[2]{\begin{CJK*}{UTF8}{min}#1\end{CJK*}\,{\normalfont\scriptsize(#2)}} 
\newcommand{\jk}[1]{\textcolor{Junk}{#1}}

\newcommand{\edit}[1]{#1} \newcommand{\chg}[1]{#1}  \newcommand{\del}[1]{}

\definecolor{Navy}{HTML}{182840}
\definecolor{PGreen}{HTML}{288858}
\definecolor{PRed}{HTML}{D04028}

\title{\includegraphics[width=\textwidth]{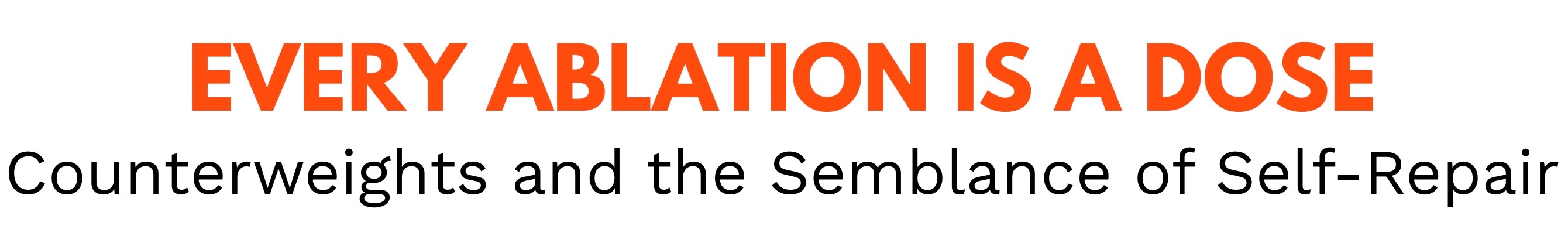}}

\author{%
  Areeb Ahmad \quad Pratinav Seth \quad Vinay Kumar Sankarapu \\[2pt]
  \affiliation{Lexsi Labs}\\[1pt]
  {\small\sffamily\mdseries \{areeb.ahmad, pratinav.seth, v.k\}@lexsi.ai}%
}

\setabstract{Ablate a component of the language model, and it's often observed that other
components adjust and appear to compensate. This phenomenon, termed self-repair,
has been observed repeatedly, but its mechanism remains unclear. The most
systematic study to date concluded that self-repair is noisy and unlikely to have
a single explanation. We argue that it has one, a gain already present before any
ablation. Any intervention on a causally important component can be viewed as a
point on a coordinate axis $\lambda$, the signed strength of a counterfactual
contrast. Hence, any conventional ablation methods are uncalibrated points on this
axis. We show that the causal repair response for a fine-grained unit $r$ is
governed by an affine law $E_r(\lambda)=own_{r} + \gamma_r\lambda$. The slope
$\gamma_r$ is a fixed coefficient that consistently influences the model, with or
without ablation, and its sign determines whether the unit counteracts or
reinforces the removed signal. On a factual-verdict task across four models from
distinct families (Gemma, Qwen, LLaMA, and Mistral), we identify components (MLP
neurons, OV neurons, and singular directions) that follow this affine law, $68$ of
$81$ downstream directions in all. Moreover, we can anticipate the magnitude of
$\gamma_r$ from the fixed weights. On the IOI circuit of GPT-2 Small, seven of the
ten heads the intervention can reach follow the law, and all seven are
counterweights. From this perspective, what may appear as self-repair is merely a
counterweight doing its usual operation when the contrastive signal emerges at the
core.}

\setkeywords{Mechanistic interpretability, self-repair, ablation, causal intervention, counterweights}

\runningtitle{Every Ablation Is a Dose}

\begin{document}
\maketitle

\section{Introduction}
\label{sec:intro}

\chg{Ablation, deleting a component and observing its effect, is a standard tool for understanding model internals. Circuit discovery, attribution, and unlearning evaluation all infer a component's contribution from ablations \citep{wang2022interpretability,conmy2023automatedcircuitdiscoverymechanistic,dai2022knowledgeneurons,hong2025intrinsic}. However, self-repair undermines this inference. Ablating} a component causes downstream components to adjust, so the behavior partly survives, and ablation can underestimate the component's importance while also misattributing what it did.

Backup name movers \citep{wang2022interpretability}, the Hydra effect \citep{mcgrath2023hydraeffectemergentselfrepair}, and \chg{anti-erasure neurons \citep{rushing2024explorationsselfrepairlanguagemodels} were each found the same way, by ablating first and sorting components by what moved, and the most systematic study concluded that the process is noisy and unlikely to have a single mechanism \citep{rushing2024explorationsselfrepairlanguagemodels}. It remains open why a model possesses this capacity and which components are predisposed to compensate.}

\chg{Prior work has documented instances without identifying the rule.} {\citet{mcdougall2023copysuppressioncomprehensivelyunderstanding} showed that one head in GPT-2 Small suppresses whatever token the model is currently predicting, so ablating it lifts the suppression, and \citet{rushing2024explorationsselfrepairlanguagemodels} found late-layer MLP neurons that oppose an upstream head and relax when it is ablated.} {\chg{Neither could tell the rule from the instance, because a deletion is a single point and cannot distinguish} a component that switches on from one that responds in proportion to the displacement of its input.} {Direct logit attribution, the readout of every prior self-repair study, compounds this, since it degrades with distance from the unembedding \citep{nostalgebraist2020,belrose2023tunedlens}, which is why every compensator found so far is a late-layer component.}

\begin{figure}[t]
  \centering
  \includegraphics[width=\linewidth]{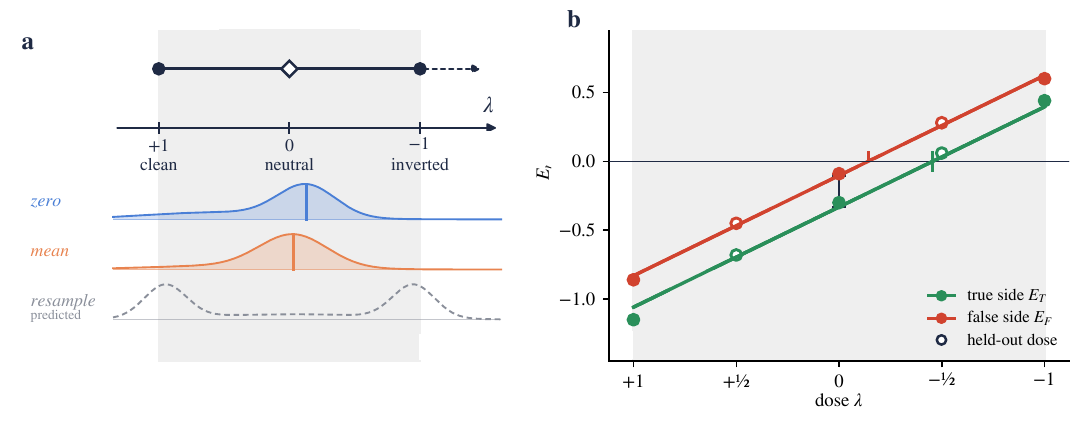}
  \caption{
  \textbf{(a)} Any intervention on the core \texttt{L28.n2046} (Gemma-2-9B-it) \chg{is a dose $\lambda$ in units of the pair's class contrast, with $+1$ clean, $0$ class signal removed, and $-1$ inverted. Zero and dataset-mean ablation land at per-prompt doses set by the data (true prompts shown), while resample} is predicted to land at the donor's dose.
  \textbf{(b)} Its counterweight \chg{\texttt{L38.n7088} ($\gamma_r < 0$)} \chg{has a contribution affine in $\lambda$, fitted on three doses} (filled), predicting held-out doses (hollow).}
  \label{fig:dose_axis}
\end{figure}

\chg{We claim that \textit{compensation is a standing coefficient}. A downstream direction $r$ (a computational unit in weights, e.g., a neuron) exerts a fixed opposition in the clean model before the upstream direction $c$ is deleted, and deletion records the release of that opposition. To see this, the intervention must be a continuous dose.} We study a True/False verdict task in which every prompt has a counterfactual partner of the opposite class, and we displace the signal in $c$ by a signed dose $\lambda$ in units of the pair's own class contrast. On this axis, \chg{every zero, mean, or resample ablation \citep{miller2024transformer,zhang2024bestpracticesactivationpatching,heimersheim2024use} is a point at a dose set by the data (Figure~\ref{fig:dose_axis}a).} Within the window of naturally occurring activations $|\lambda|\le 1$, $r$'s contribution to the verdict is affine in the dose,
\begin{equation}
  E_r(\lambda)=\mathrm{own}_r+\gamma_r\lambda,
  \label{eq:affine_law}
\end{equation}
\chg{with a fixed gain $\gamma_r$, which makes $r$ a \emph{counterweight} if $\gamma_r<0$ and a \emph{relay} if $\gamma_r>0$ (Figure~\ref{fig:dose_axis}b). The capacity is therefore standing, and compensation is its release as seen through a coarse intervention. The predisposed directions are the ones} the weights already couple to the core, and the magnitude of $\gamma_r$ is anticipated by the alignment of the core's write vector with $r$'s read weights in the static checkpoint (\S\ref{sec:coupling_mag_ground_in_weights}).

{We test this with a nested counterfactual intervention (\S\ref{sec:test}) in four instruction-tuned models of different families and on the IOI circuit of GPT-2 Small.} Scope conditions are stated in \edit{\S\ref{sec:conclusion}}.

\paragraph{Contributions.}
\begin{enumerate}[leftmargin=*, itemsep=2pt, topsep=2pt]
  \item \textbf{Compensation is standing.} The directions that self-repair \chg{already oppose the core in the clean model, so deletion releases} a gain rather than recruiting one. This holds across four instruction-tuned models \chg{and for the seven IOI heads that follow the law (\S\ref{sec:census}, \S\ref{sec:deletion})}.

  \item \textbf{A dose axis and a nested intervention.} Every conventional ablation is an uncontrolled point on a single axis, and a nested counterfactual intervention, read by total effect inside the dosed model, makes it observable  (\S\ref{sec:test}, \S\ref{sec:deletion}).

    \item \textbf{{The coupling has an affine form.}} {Within the on-manifold window $|\lambda|\le1$, a downstream direction's contribution to the verdict varies nearly linearly with the dose applied to the core, Eq.~\ref{eq:affine_law}.} {Lines fitted on three doses predict held-out intermediate doses with out-of-sample \chg{$R^2$ of $0.65$ to $0.91$ by family}} (\S\ref{sec:law}).

  \item \textbf{A population of named counterweights and relays.} Directions named from the weights alone are legible at the unembedding yet inert or opposing in the clean model, which is why deletion never surfaces them. They are certified by intervention rather than assumed from their labels, and their coupling magnitude is anticipated from the static checkpoint (\S\ref{sec:census}, \S\ref{sec:coupling_mag_ground_in_weights}).
\end{enumerate}

\section{Setup}
\label{sec:setup}

\paragraph{Weight-space bases.}

\chg{Decoder-only transformers add writes $o_c(x)\in\mathbb{R}^{d}$ from}
each attention head and MLP into a residual stream \citep{elhage2021mathematical}.
We decompose each write along two fixed sets of weight-space directions.
\chg{(I)~\textbf{Neuron directions} give $o_c=\sum_j z_{cj}\,w_{cj}$}, where $w_{cj}$ is a
column of the component's write matrix ($W^d_c$ for MLPs, $W_O$ for heads) and
$z_{cj}$ is its scalar activation.~\chg{(II)~\textbf{SVD basis} gives $o_c=\sum_k \alpha_{ck}\,v_{ck}$},
where $\{v_{ck}\}$ are the left-singular vectors of $W^d_c$ for MLPs
and of the OV circuit $W_OW_V$ for heads, and $\alpha_{ck}$ is an input dependent scalar \citep{ahmad2025componentssingularvectorbasedinterpretability,xue2026svd}.

\paragraph{A single interface.}
We unify these as a single \emph{direction} $r$, parameterized by a fixed write vector $\omega_r \in \mathbb{R}^{d}$ and an activation coefficient $a_r(x) \in \mathbb{R}$. This factorization \chg{isolates single-direction interventions, where $\mathrm{do}(a_r := v)$ overwrites} the coefficient of a specific direction  $\omega_r$. A direction's pole $\omega_r^{\top} W_U$ at the unembedding $W_U$ \chg{sets its \emph{direct effect}. The write moves} the logit of token $t$ by $a_r(x)\langle W_U[:,t], \omega_r\rangle$ {before the final norm}. We read these poles from the input-independent weights \citep{nostalgebraist2020, geva2021transformerfeedforwardlayerskeyvalue, dar2023analyzingtransformersembeddingspace, millidge_black_2022_svd, ahmad2025componentssingularvectorbasedinterpretability,hong2025intrinsic,xue2026svd}.

\paragraph{Interchange interventions.}
We treat the forward pass as a deterministic structural causal model \citep{pearl2009causality} in which the prompt is the exogenous variable and \chg{the basis coefficients $a_r(x)$ are the endogenous variables.} On this model, activation patching is the operation \chg{$\mathrm{do}(a_r := a_r(x'))$, which pins a coefficient} to the value it takes on a source input $x'$ \citep{vig2020causalmediationanalysisinterpreting, chan2022causal, geiger2020neuralnaturallanguageinference,geiger2021causalabstractionsneuralnetworks}. Every intervention in this paper, including the dose ladder of \S\ref{sec:test}, is built from this primitive.

\paragraph{Verdict margin.}
We evaluate the model's factual verdict with the \emph{verdict margin}, the gap in mean logits between faithful ($X$) and contrastive \chg{($Y$) answer tokens,}
\begin{equation}
  D_{XY}(x) = \frac{1}{|X|}\sum_{t\in X} \ell_t(x) - \frac{1}{|Y|}\sum_{t\in Y} \ell_t(x).
  \label{eq:margin}
\end{equation}
Evaluation is conditioned on clean-correct pairs, so $D_{XY} > 0$ \chg{on every clean run, and any intervention} that damages the verdict lowers $D_{XY}$.

\paragraph{Models, task, and data.}
\chg{We study four instruction-tuned models (Llama-3-8B-Instruct, Gemma-2-9B-it, Qwen2.5-7B-Instruct, and Mistral-7B-Instruct-v0.3).}
The task is a zero-shot factual verdict (true/false) on declarative statements (App.~\ref{app:prompts}). Contrastive pairs are synthesized from the Known-Facts corpus \chg{\citep{meng2022locating} so that the true and false variants} of a pair differ only in the asserted attribute, leaving all other phrasing identical (App.~\ref{app:data}).

\section{The Dose Axis}
\label{sec:test}

\chg{Every experiment in this paper measures what a downstream direction does when the core's class signal is disturbed. This requires}
two interventions composed in sequence, a \emph{primary} that disturbs the core
and a \emph{secondary} that measures a direction in the disturbed model. This
section defines both, puts every disturbance on a single axis, and states the
estimand.

\paragraph{A single coordinate for the core direction.}
Let $(x, x')$ denote a contrastive prompt pair (e.g., True vs.\ False), and let $a(x)$ and $a(x^\prime)$ be the corresponding activations along a core direction $c$. Decomposing these activations yields
\[
  m \;=\; \frac{a(x) + a(x^{\prime})}{2}, \qquad
  \Delta \;=\; a(x^{\prime}) - a(x),
\]
the \emph{mid-point} ($m$), which isolates the shared context across the prompt pair, and the \emph{contrast} ($\Delta$), which isolates the class-differential signal. Any clamp value $v$ applied to the core lies on the line connecting $a(x)$ and $a(x^\prime)$, \chg{parameterized by a single scalar,}
\begin{equation}
  a(\lambda) \;=\; m - \frac{\lambda}{2}\,\Delta ,
  \qquad
  \lambda(v) \;=\; \frac{2\,(m - v)}{\Delta} .
  \label{eq:dose}
\end{equation}
The continuous axis has three reference points that are intuitively conventional. At $\lambda=+1$ the core holds its clean activation for the given class. At $\lambda=-1$, it assumes the opposite-class activation, that is, the standard activation swap \citep{vig2020causalmediationanalysisinterpreting, geiger2020neuralnaturallanguageinference,geiger2021causalabstractionsneuralnetworks}.  At $\lambda=0$, the core rests at the class-neutral midpoint $m$, the unique point with no class differential.
Doses $\lambda<-1$ overshoot the counterfactual, injecting more than one unit of
the pair's own contrast, as in graded steering with a multiplier above one
\citep{zou2023representation,rimsky-etal-2024-steering,turner2024steeringlanguagemodelsactivation}. We refer to
$k=(1-\lambda)/2$ as the corruption fraction, \chg{where it aids reading, and $k$ runs}
$0\to1$ as $\lambda$ runs $+1\to-1$.

\paragraph{The primary intervention (clean world).}
\label{para:clean_world}
The primary is a dosed swap of the core toward its partner value,
\begin{equation}
  \pi_\lambda \;:=\; \mathrm{do}\Bigl(a_{c} := m - \tfrac{\lambda}{2}\Delta\Bigr)
  \;=\; \mathrm{do}\Bigl(a_{c} := a_c(x) + k\,[a_c(x') - a_c(x)]\Bigr) ,
  \label{eq:swap}
\end{equation}
applied to the core direction(s) $c$ certified in \S\ref{sec:census}, and to all
of them jointly, where a family has more than one. The clean model is the point \chg{$\lambda=+1$. We write $b=\varnothing$} for it and $b=\pi_\lambda$ for the dosed world. The ladder is $\lambda\in\{+1,\,+\tfrac12,\,0,\,-\tfrac12,\,-1,\,-3,\,-5,\,-7\}$.

\paragraph{The secondary intervention (dosed world).}
Within a world $b\in\{\varnothing,\pi_\lambda\}$, we read a downstream direction $r$ by patching its coefficient with the value it takes on the partner prompt in that same world
\begin{equation}
  \mathrm{sec}_r^{\,b} \;:=\; \mathrm{do}\bigl(a_r := a_r(x';\,b)\bigr).
  \label{eq:reference}
\end{equation}
\chg{Here $\pi_\lambda$ is a rule applied per prompt, so on $x'$ it clamps} the core to $m + \tfrac{\lambda}{2}\Delta$, the mirror image of Eq.~\ref{eq:swap} about $m$. At every dose the two prompts' cores sit equidistant from the midpoint on opposite sides, \chg{coinciding only at $\lambda = 0$. The coefficient $a_r(x';\pi_\lambda)$ is then} $r$'s activation on the partner prompt inside this mirror-dosed graph (Figure~\ref{fig:design}).

\definecolor{TrueC}{HTML}{1A7F37}
\definecolor{FalseC}{HTML}{B42318}

\definecolor{TrueC}{HTML}{1A7F37}
\definecolor{FalseC}{HTML}{B42318}
\begin{figure*}[t]
\centering
\resizebox{\textwidth}{!}{\begin{tikzpicture}[
  x=1cm, y=1cm, line cap=round, line join=round,
  every node/.style={font=\footnotesize},
  frame/.style={draw=black!30, line width=0.5pt, rounded corners=2pt},
  title/.style={font=\small, black!85, anchor=west},
  unit/.style={draw=black!75, fill=white, line width=0.6pt, rounded corners=1.5pt,
               minimum width=20pt, minimum height=14pt, inner sep=1pt},
  faded/.style={unit, draw=black!30, text=black!45},
  io/.style={unit, draw=TrueC, text=TrueC!70!black, fill=TrueC!6, minimum width=14pt},
  readout/.style={draw=black!75, double, circle, minimum size=16pt, inner sep=0pt, line width=0.6pt},
  value/.style={draw=FalseC, line width=0.6pt, rounded corners=1.5pt, fill=FalseC!4,
                text=FalseC!75!black, minimum height=14pt, inner sep=2.5pt},
  stream/.style={->, >={Stealth[length=5pt]}, black!30, line width=2.2pt},
  wire/.style={->, >={Stealth[length=4pt]}, black!65, line width=0.6pt},
  fwire/.style={->, >={Stealth[length=4pt]}, black!28, line width=0.6pt},
  inject/.style={->, >={Stealth[length=4pt]}, FalseC, line width=0.7pt, dash pattern=on 2.5pt off 1.5pt},
  role/.style={font=\footnotesize, black!55},
]

\newcommand{\openread}[2]{  \draw[black!65, line width=0.6pt] (#1,0) -- (#1,0.28*#2);
  \fill[black!75] (#1,0.28*#2) circle (1.1pt);
  \draw[black!75, line width=0.8pt] (#1,0.28*#2) -- ++(0.17,0.2*#2);
  \fill[black!75] (#1,0.52*#2) circle (1.1pt);
  \draw[wire] (#1,0.52*#2) -- (#1,0.75*#2);}

\newcommand{\panel}[4]{  \draw[frame] (0,-1.62) rectangle (6.6,2.02);
  \node[title] at (0.1,1.74) {#1};
  \node[io] (x) at (0.42,0) {$x$};
  \node[readout] (D) at (6.12,0) {$D$};
  \draw[stream] (x.east) -- (D.west);
  \node[unit, #2] (c) at (2.95,-1.0) {$c$};
  \node[role, anchor=north] at (c.south) {core};
  \draw[wire] (3.13,-0.75) -- (3.13,0);
  \node[unit, draw=FalseC, text=FalseC!80!black, fill=FalseC!5] (r) at (4.72,1.0) {$r$};
  \draw[wire, FalseC!80] (4.90,0.75) -- (4.90,0);
  \openread{4.54}{1}
  \node[value] (vr) at (2.95,1.0) {#3};
  \draw[inject] (vr.east) -- (r.west);
  \node[faded] (o1) at (4.1,-1.0) {$\cdots$};
  \draw[fwire] (3.92,0) -- (3.92,-0.75);
  \draw[fwire] (4.28,-0.75) -- (4.28,0);
  \node[faded] (o3) at (5.34,-1.0) {$\cdots$};
  \draw[fwire] (5.16,0) -- (5.16,-0.75);
  \draw[fwire] (5.52,-0.75) -- (5.52,0);
  \node[faded] (o2) at (1.65,1.0) {$\cdots$};
  \draw[fwire] (1.47,0) -- (1.47,0.75);
  \draw[fwire] (1.83,0.75) -- (1.83,0);
  \ifnum#4=1
    \openread{2.77}{-1}
    \node[value] (vc) at (1.2,-1.0) {$a_c := m-\tfrac{\lambda}{2}\Delta$};
    \draw[inject] (vc.east) -- (c.west);
  \else
    \draw[wire] (2.77,0) -- (2.77,-0.75);
  \fi
}

\begin{scope}
  \panel{(a) clean world, $b=\varnothing$}{}{$a_r(x';\varnothing)$}{0}
\end{scope}
\begin{scope}[shift={(7.2,0)}]
  \panel{(b) dosed world, $b=\pi_\lambda$}{draw=FalseC, fill=FalseC!10, text=FalseC!80!black}        {$a_r(x';\pi_\lambda)$}{1}
\end{scope}
\end{tikzpicture}}
\caption{\textbf{The nested counterfactual intervention}. \chg{{Open switches cut a read from the residual stream, and dashed arrows inject the replacement value.}} {\chg{\emph{(a) Clean world, $b=\varnothing$.} The downstream direction} $r$ is patched with $a_r(x';\varnothing)$, its value on the partner prompt $x'$ in the unmodified model.} {\chg{\emph{(b) Dosed world, $b=\pi_\lambda$.} The core $c$ is first clamped to dose $\lambda$ (Eq.~\ref{eq:swap}), and $r$ is then} patched with $a_r(x';\pi_\lambda)$, computed on $x'$ through this \emph{same} mirror-dosed graph.}}
\label{fig:design}
\end{figure*}

\paragraph{The estimand.}
Let $D(x;b)$ be the verdict margin {on $x$} in world $b$. The contribution of $r$ in world $b$ is the margin erased \chg{when its natural reading is intervened on,}
\begin{equation}
  E^{b}_{r}(x) \;=\; D(x;\,b) \;-\; D(x;\,b,\,\mathrm{sec}_r^{\,b}).
  \label{eq:contribution}
\end{equation}
$E>0$ means $r$ supports the verdict in that world, $E<0$ that it opposes. It is a total effect. Both passes run in world $b$, so a shift of the primary adds to both margins, which cancels in $E$.  In the clean world (with $b=\varnothing$) we write $E^{0}_r$.The point contrast
\begin{equation}
  R_r(\lambda) \;=\; E^{b}_r - E^{0}_r
  \label{eq:recruitment}
\end{equation}
measures how much $r$'s contribution changes between the clean and dosed worlds
at a single dose.

\section{\chg{Census of the Core and Its Counterweights}}
\label{sec:census}

A direction's label in the weights \chg{says what it writes, the tokens its output} promotes or suppresses at the unembedding. It says nothing about the direction's causal role \citep{bolukbasi2021interpretabilityillusionbert,huang2023rigorouslyassessingnaturallanguage}. This section separates the two. It certifies the core, the one or two directions per family that carry the verdict, and identifies the legible directions that do not carry it at rest yet respond once the core is intervened on. Both come from one bench, named from the weights alone.

\paragraph{A bench named from the weights.}
We scan the four families (\S\ref{sec:setup}) and label each direction $r$ by the tokens it promotes most and suppresses most on the unembedding. A direction enters the bench if a truth or falsity token, in English or another language, \chg{appears at either pole, and the suppressing pole} counts as much as the promoting one (App.~\ref{app:census-catalogues}).\footnote{The core neurons were first noticed as the sites at which the two statement classes separate most sharply in the clean forward pass, before the bench was constructed. The scan uses no activations and recovers them.} \chg{Selection uses only the checkpoint. The bench was fixed} before any dose was run and not revised after. Whether a direction affects the verdict is the causal question the census answers.

\begin{table}[t]
\centering
\small
\caption{\textbf{The core carries roughly half the total effect.} Total effect $E=E_T+E_F$ \chg{of the whole model on the verdict, where $E_T$ is} the True$-$False logit margin on the true statement and $E_F$ the False$-$True margin on the false statement, at the answer token. \chg{$E^0$ is the clean model, and $E^{\pi}$ is the same} with the core swapped to its full counterfactual ($k=1$, $\lambda=-1$). The last column is the share of the clean total effect the core is responsible for, $(E^0-E^{\pi})/E^0$. Standard errors in parentheses.}
\label{tab:core-ablation}
\begin{tabular}{llrrr}
\toprule
Model & Core(s) & $E^0$  & $E^{\pi}$  & core share \\
\midrule
Llama-3-8B-Instruct & \texttt{L18.n11065} & $7.78$ $(0.11)$ & $3.67$ $(0.05)$ & $52.7\%$ \\
Gemma-2-9B-it & \texttt{L28.n2046} $+$ \texttt{L33.n4294} & $9.50$ $(0.11)$ & $4.38$ $(0.05)$ & $53.9\%$ \\
Qwen2.5-7B-Instruct & \texttt{L22.n13149} $+$ \texttt{L24.n14758} & $18.76$ $(0.17)$ & $8.46$ $(0.07)$ & $54.9\%$ \\
Mistral-7B-Instruct-v0.3 & \texttt{L20.n14286} $+$ \texttt{L19.n8228} & $11.59$ $(0.15)$ & $5.03$ $(0.06)$ & $56.6\%$ \\
\bottomrule
\end{tabular}
\end{table}
\paragraph{One or two directions carry the verdict.}
In the clean world ($\lambda=+1$,\S\ref{para:clean_world}), we measure each direction's contribution $E^{0}_r$ (Eq.~\ref{eq:contribution}) and certify it as a core only if $E^{0}_r > 0$, reliably, on both classes. The core is small in every family, a single neuron in Llama (\texttt{L18.n11065}) and a pair in each of Gemma(\texttt{L28.n2046}, \texttt{L33.n4294}), Qwen (\texttt{L22.n13149}, \texttt{L24.n14758}), and Mistral (\texttt{L20.n14286}, \texttt{L19.n8228}), \chg{and it carries more than half of the clean total effect, $52.7$ to $56.6\%$ (Table~\ref{tab:core-ablation}).} Of the other directions, some are inert, others oppose the verdict on both classes despite legible truth or falsity poles.

\paragraph{The dismissed directions move against the core.}
Now intervene on the core (Eq.~\ref{eq:swap}) and re-measure
Eq.~\ref{eq:contribution} inside the dosed world. Most of these directions
move against it. As the core's class signal is weakened, their opposition to
\chg{the verdict shrinks. At some dose} it reaches zero, and past that dose they
support the verdict instead. We call such a direction a \emph{counterweight} (formally, $\gamma_r<0$ in Eq.~\ref{eq:affine_law}). Figure~\ref{fig:recruit_four} follows one \emph{counterweight} per family. Llama's \texttt{L23.n8972} goes from $-0.10$ to $+0.25$ nats, Gemma's \texttt{L38.n7088} from $-2.1$ to $+3.9$, Qwen's \texttt{L27.n13033} from
$-5.7$ to $+8.7$, and Mistral's \texttt{L31.n8773} from $-1.5$ to $+3.2$, each crossing zero on both classes. Together with the census, this shows that these directions write the truth vocabulary and pull against the core. \chg{A smaller set moves with the core, and \S\ref{sec:law} treats} these \emph{relays} on the same axis. The rest of the
bench is tabulated in App.~\ref{app:recruitment-tables}. The base checkpoints show the same response (App.~\ref{app:pretraining}). Renormalization, which \citet{rushing2024explorationsselfrepairlanguagemodels} credit with part of self-repair, explains at most $4.5\%$ \chg{of the audited recruitment (App.~\ref{app:renorm}).}

\definecolor{ETgreen}{RGB}{20,110,70}
\definecolor{EFred}{RGB}{178,34,52}
\definecolor{netink}{RGB}{28,28,34}
\definecolor{bandgray}{RGB}{237,237,240}

\begin{figure}[t]
\centering
\begin{tikzpicture}[
  every axis plot post/.append style={line join=round, line cap=round},
]
\begin{groupplot}[
  group style={group size=4 by 1, horizontal sep=0.85cm, ylabels at=edge left},
  width=0.29\linewidth, height=0.27\linewidth,
  x dir=reverse,
  xmin=-7.6, xmax=1.6,
  xtick={1,-1,-3,-5,-7},
  xticklabels={$+1$,$-1$,$-3$,$-5$,$-7$},
  ylabel={net $E^{\pi}_r$ (nats)}, ylabel style={font=\scriptsize, yshift=-4pt},
  tick label style={font=\tiny},
  tick align=outside, tick pos=left, major tick length=2pt,
  title style={font=\scriptsize\bfseries, yshift=-3pt},
  ymajorgrids=true, grid style={draw=black!10, line width=0.3pt},
  axis line style={draw=black!35, line width=0.4pt},
  axis on top, enlarge x limits=false, scaled y ticks=false,
  every axis plot/.append style={line width=0.9pt},
  legend columns=3,
  legend style={font=\scriptsize, draw=none, fill=none, column sep=10pt},
]

\nextgroupplot[
  title={Llama\ \ \texttt{L23.n8972}},
  ymin=-0.16, ymax=0.32,
  ytick={-0.1,0,0.1,0.2,0.3},
  yticklabels={$-0.1$,$0$,$0.1$,$0.2$,$0.3$},
  legend to name=recruitlegend]
  \fill[bandgray] (axis cs:-1,\pgfkeysvalueof{/pgfplots/ymin})
        rectangle (axis cs:1,\pgfkeysvalueof{/pgfplots/ymax});
  \draw[black!45, line width=0.4pt] (axis cs:-7.6,0) -- (axis cs:1.6,0);
  \addplot[ETgreen, densely dashed, line width=0.7pt]
    coordinates {(1,-0.04) (-1,0.02) (-3,0.06) (-5,0.11) (-7,0.16)};
  \addplot[EFred, densely dotted, line width=0.9pt]
    coordinates {(1,-0.06) (-1,0.02) (-3,0.04) (-5,0.06) (-7,0.09)};
  \addplot[netink, line width=1.0pt, mark=*, mark size=1.4pt,
           mark options={fill=netink, draw=white, line width=0.3pt}]
    coordinates {(1,-0.10) (-1,0.04) (-3,0.10) (-5,0.17) (-7,0.25)};
  \legend{$E_T$, $E_F$, net}

\nextgroupplot[
  title={Gemma\ \ \texttt{L38.n7088}},
  ymin=-2.8, ymax=4.6, ytick={-2,0,2,4}]
  \fill[bandgray] (axis cs:-1,\pgfkeysvalueof{/pgfplots/ymin})
        rectangle (axis cs:1,\pgfkeysvalueof{/pgfplots/ymax});
  \draw[black!45, line width=0.4pt] (axis cs:-7.6,0) -- (axis cs:1.6,0);
  \addplot[ETgreen, densely dashed, line width=0.7pt]
    coordinates {(1,-1.420) (-1,0.474) (-3,1.060) (-5,1.178) (-7,1.215)};
  \addplot[EFred, densely dotted, line width=0.9pt]
    coordinates {(1,-0.715) (-1,0.613) (-3,2.202) (-5,2.587) (-7,2.727)};
  \addplot[netink, line width=1.0pt, mark=*, mark size=1.4pt,
           mark options={fill=netink, draw=white, line width=0.3pt}]
    coordinates {(1,-2.135) (-1,1.087) (-3,3.262) (-5,3.765) (-7,3.942)};

\nextgroupplot[
  title={Qwen\ \ \texttt{L27.n13033}},
  ymin=-7, ymax=10, ytick={-5,0,5,10}]
  \fill[bandgray] (axis cs:-1,\pgfkeysvalueof{/pgfplots/ymin})
        rectangle (axis cs:1,\pgfkeysvalueof{/pgfplots/ymax});
  \draw[black!45, line width=0.4pt] (axis cs:-7.6,0) -- (axis cs:1.6,0);
  \addplot[ETgreen, densely dashed, line width=0.7pt]
    coordinates {(1,-2.86) (-1,-0.50) (-3,2.93) (-5,3.95) (-7,4.08)};
  \addplot[EFred, densely dotted, line width=0.9pt]
    coordinates {(1,-2.88) (-1,-0.51) (-3,3.22) (-5,4.45) (-7,4.65)};
  \addplot[netink, line width=1.0pt, mark=*, mark size=1.4pt,
           mark options={fill=netink, draw=white, line width=0.3pt}]
    coordinates {(1,-5.74) (-1,-1.01) (-3,6.15) (-5,8.40) (-7,8.73)};

\nextgroupplot[
  title={Mistral\ \ \texttt{L31.n8773}},
  ymin=-2.2, ymax=3.8, ytick={-2,0,2}]
  \fill[bandgray] (axis cs:-1,\pgfkeysvalueof{/pgfplots/ymin})
        rectangle (axis cs:1,\pgfkeysvalueof{/pgfplots/ymax});
  \draw[black!45, line width=0.4pt] (axis cs:-7.6,0) -- (axis cs:1.6,0);
  \addplot[ETgreen, densely dashed, line width=0.7pt]
    coordinates {(1,-0.771) (-1,0.024) (-3,1.088) (-5,1.489) (-7,1.438)};
  \addplot[EFred, densely dotted, line width=0.9pt]
    coordinates {(1,-0.748) (-1,0.050) (-3,1.240) (-5,1.754) (-7,1.732)};
  \addplot[netink, line width=1.0pt, mark=*, mark size=1.4pt,
           mark options={fill=netink, draw=white, line width=0.3pt}]
    coordinates {(1,-1.519) (-1,0.074) (-3,2.328) (-5,3.243) (-7,3.170)};

\end{groupplot}
\path (group c2r1.outer south) -- (group c3r1.outer south)
      node[midway, below=2pt, font=\scriptsize] {dose $\lambda=1-2k$};
\end{tikzpicture}
\vspace{3pt}
\pgfplotslegendfromname{recruitlegend}
\caption{\textbf{{The featured counterweight in each family} moves against the core.} Net contribution $E^{\pi}_r = E_T + E_F$ of one direction per family against the dose $\lambda$ on that family's certified core(s). \chg{The shaded band is the
on-manifold window $|\lambda|\le 1$, where $+1$ is the clean model and $-1$ the
counterfactual.} Each opposes the verdict at rest and crosses zero on both classes at a sufficient dose. \chg{Points are per-dose means, and standard errors are smaller} than the markers (paired $t$,
App.~\ref{app:recruitment-tables})}
\label{fig:recruit_four}

\end{figure}

\paragraph{The response tracks the amount of signal removed.}
In families with two cores, dosing either core alone \chg{moves part of the bench, and dosing both} moves more of it than either alone, and at low dose by more than the sum of the two (App.~\ref{app:recruitment-svj}). Two independent couplings \chg{would add exactly. The excess is} the share of the signal that the free core carries when only one is clamped.

\paragraph{\chg{The response follows the signal across interventions.}}
The same directions respond to a procedurally unrelated primary intervention, a learned
label-free perturbation of the singular values, and to its conjunction with the swap (App.~\ref{app:perturbation}). We can see that recruitment/compensation is robustly present in the model, irrespective of the disruption method.

\begin{figure}
    \centering
    \includegraphics[width=\linewidth]{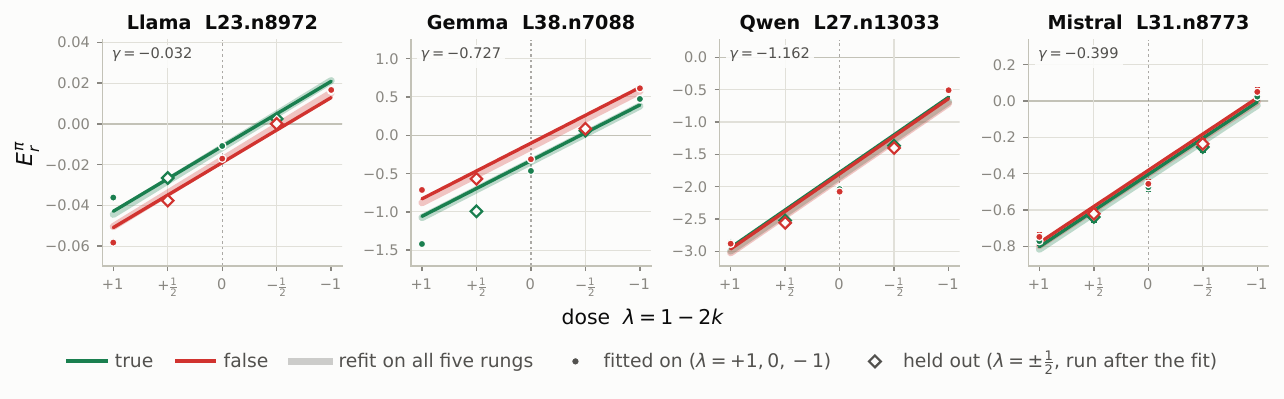}
    \caption{\textbf{The Affine Law}, four model families. For one exemplar direction per model, the erased margin $E_r^{\pi}$ against the dose $\lambda$. \chg{Lines are fit on the three filled rungs, with the hollow rungs run afterward and held out.} The band is the refit on all five.}
    \label{fig:law-exemplars}
\end{figure}

\begin{figure}[t]
\centering
\includegraphics[width=\linewidth]{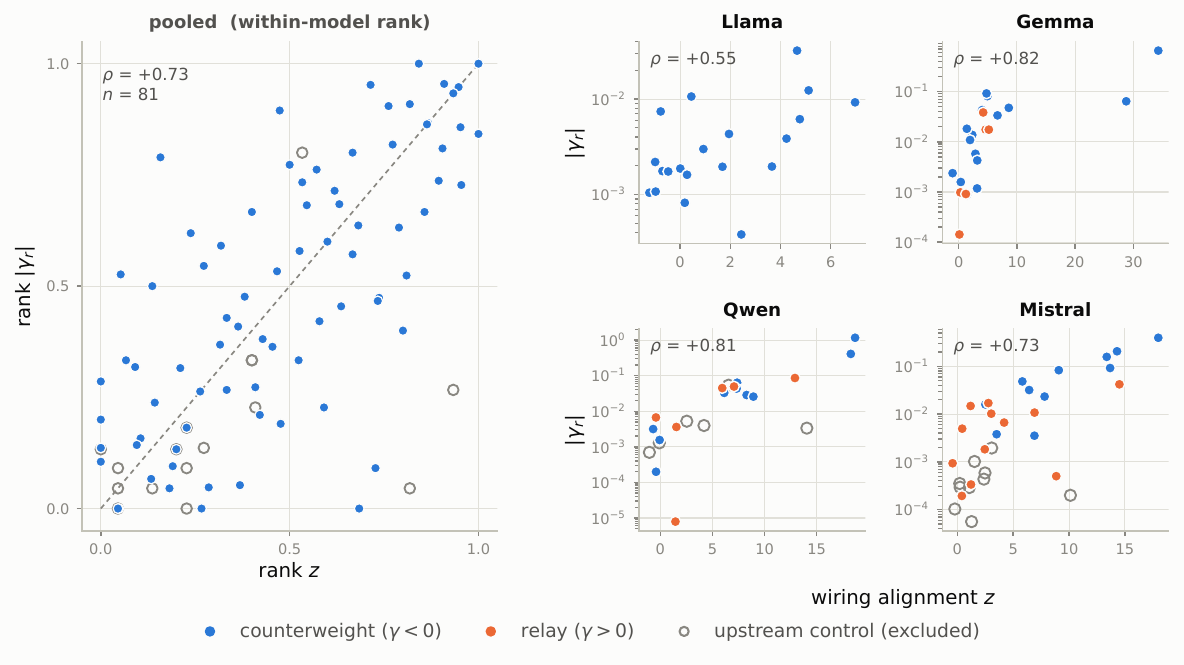}
\caption{\textbf{Coupling strength is legible in the weights}. Weight-space alignment $z$ between the core's write and each direction's read, computed from the checkpoint alone, against the intervention-measured coupling \chg{$\vert{}\gamma_r\vert{}$, pooled on a within-model rank scale (left, $\rho=+0.73$}, $n=81$) and per model (right). Hollow markers are upstream components with no causal path to the core.}
\label{fig:weight-visibility}
\end{figure}

\section{The Coupling Law}
\label{sec:law}

\S\ref{sec:census} \chg{fixed the sign of the response. As the core's} signal is withdrawn, most of the bench moves toward supporting the verdict.  It could not fix the shape, because only two of its doses, the clean model and full inversion, lie inside the on-manifold window ($|\lambda|\le1$, where the natural activation values lie). Shape is what separates the two readings. A component that switches on when the core's signal is lost, sits at its resting value until some level, and rises after it, one that responds in proportion to the displacement gives a line.Across the dose ladder, the response is a
line. For each bench direction $r$ and class $c\in\{True,False\}$, on $|\lambda|\le1$,
\begin{equation}
  E_r^{c}(\lambda) \;=\; \mathrm{own}_r^{c} \;+\; \gamma_r\,\lambda ,
  \label{eq:law}
\end{equation}
where $E_r^{c}(\lambda)$ is the contribution of Eq.~\ref{eq:contribution} on class $c$ \chg{in the world $\pi_\lambda$, in which the core is held at dose $\lambda$} in both passes the contrast compares, the measured pass on $x$ and the partner pass on $x'$ that supplies $r$'s counterfactual value
(Figure~\ref{fig:design}). We fit Eq.~\ref{eq:law} by weighted least squares on both classes at once, with weights from the per-rung standard errors, on the three rungs \chg{$\lambda\in\{+1,0,-1\}$, and we hold out} the rungs at $\pm\tfrac12$ and beyond $-1$ to test the line.

\paragraph{What the two parameters are.}
$\gamma_r$ is the coupling, how much of the core's signal reaches the verdict
through $r$ and \chg{with what sign. A negative slope} marks a
\emph{counterweight}, a positive one a \emph{relay}. One slope serves both classes (fitted separately, they agree at $\rho=0.83$) and \chg{one intercept serves each. Their mean} is $r$'s baseline, and their difference $\Delta\mathrm{own}_r$ is the class information $r$ gathers through routes other than the core, constant in dose, so the two class lines run parallel. The split raises the median in-sample $R^2$ in every family, \chg{from $[0.50,0.76]$ to $[0.82,0.91]$, and without it the law fails Mistral out of sample ($R^2=-0.41$, App.~\ref{app:law-tests})}.

\paragraph{Most of the bench couples to the core.}
Of the 81 bench directions the core can reach, 68 have a slope reliably different from zero \chg{under both fits, with the false discovery rate controlled at} $q=0.05$ across each family's full bench, including upstream controls (App.~\ref{app:law-tests}). Of these, 52 are counterweights, and 16 are relays (Table~\ref{tab:family-breakdown}), \chg{a ratio specific to a bench selected for legible poles.} \chg{The remaining 13 fail the significance gate. Seven have near-zero slopes, and six sit narrowly outside the correction.} Figure~\ref{fig:law-exemplars} shows one direction per family with its fitted lines.

The lines hold on doses they never saw. After freezing the fits we ran the half rungs \chg{at $\lambda=\pm\tfrac12$. The line predicts} them to within a median $9\%$ of each direction's response range, with out-of-sample $R^2$ between $0.65$ and $0.91$ by family. These doses lie inside the window, so the test shows that the line tracks a smooth response, \chg{not that no other smooth curve would. A quadratic term changes} $\gamma_r$ by less than its standard error (App.~\ref{app:law-tests}). Past full inversion the response saturates, so we certify Eq.~\ref{eq:law} on $|\lambda|\le1$ only. The same law resolves on the IOI circuit of GPT-2 Small, where every certified head is a counterweight (App.~\ref{app:ioi}).

\section{The Coupling Magnitude Is Grounded in the Weights}
\label{sec:coupling_mag_ground_in_weights}
$\gamma_r$ could be viewed as an artifact of the experimental design rather than an intrinsic network property. To address this, we define a static measure computed from the model weights. For downstream direction $r$, we compute \chg{the wiring alignment $z_r$, the cosine similarity} between the core's write vector $\omega_c$ and the rows where $r$ reads the residual stream, \chg{$z$-scored against $2{,}000$ random readers in the same layer (App.~\ref{app:wiring})}. Figure~\ref{fig:weight-visibility} shows that $\vert{}\gamma_r\vert{}$  correlates with $z_r$ across all model families \chg{(Spearman $\rho \in [0.55, 0.82]$, one-sided permutation $p \le 0.01$ in every model, Table~\ref{tab:onehop-perm}).} Given that four models were trained independently, \chg{Fisher's method combines them to $p \approx 10^{-10}$, an upper bound because Gemma's $p$ sits at the permutation floor.} This relationship remains robust when $\gamma_r$ is refitted exclusively on the intervened rungs $\lambda \in \{0, -1\}$, excluding clean-run data entirely ($\rho \in [0.43, 0.83]$, combined $p < 10^{-9}$). This statistic is ordinal and unsigned, accounts only for the direct value pathway, and \edit{tracks} the clean effect $|E_r^{0}|$ \edit{($\rho\in[0.68,0.86]$) about as strongly as $|\gamma_r|$, since} $E_r^{0}=\mathrm{own}_r+\gamma_r$ at $\lambda=+1$ (Eq.~\ref{eq:law})\edit{, a link the refit does not remove}. Consequently, we claim only that the checkpoint anticipates the \emph{magnitude} of the coupling, not the exact coefficient (App.~\ref{app:wiring-limits}).

\begin{figure}[t]
\centering
\includegraphics[width=\linewidth]{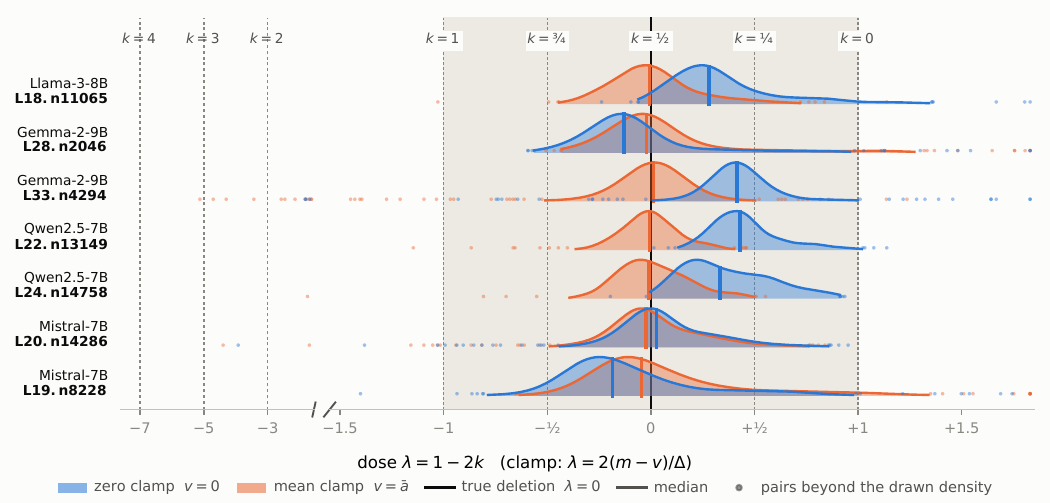}
\caption{\textbf{Deletion on the dose axis.} ~For each certified core, the
distribution over pairs of the dose $\lambda(v)=2(m-v)/\Delta$ administered by
a zero clamp ($v=0$, blue) and a dataset-mean clamp ($v=\bar a$, orange), with
medians marked. The vertical line is $\lambda=0$, the shaded band is the on-manifold window, the broken axis is the off-manifold tail.}
\label{fig:deletion_dose_a}
\end{figure}

\section{Deletion on the Dose Axis}
\label{sec:deletion}

Self-repair has only ever been \chg{measured by deletion, whether by mean ablation}
\citep{wang2022interpretability}, task-matched resampling
\citep{mcgrath2023hydraeffectemergentselfrepair}, or open-distribution
resampling \citep{rushing2024explorationsselfrepairlanguagemodels}. All three assume that the clamp strictly removes the component's contribution, interpreting whatever follows as the network's response to the loss. In our perspective, any clamp $a_c:= v$ is a point on the axis of \S\ref{sec:test}. It lands at dose ${\lambda}{(v) = 2(m - v)/\Delta}$, where Eq.~\ref{eq:affine_law}  governs the behavior of every certified counterweight and relay. Figure~\ref{fig:deletion_dose_a} shows how each clamp $v$ distributes on the dose axis. Zero ablation lands at $2m/\Delta$. Mean ablation is neutral on average because a balanced reference set puts its mean $\bar{a}$ near $m$, \chg{but not per pair. Since the dose} divides by the pair's contrast, pairs with a small contrast receive a large dose. Resample ablation lands wherever its donor pool places it \chg{(App.~\ref{app:realised} and Table~\ref{tab:clamp})}. In none of these three methods does the experimenter choose the dose. The data chooses it, assigning a different dose to every pair.

\paragraph{What the founding literature's observations saw.}
\chg{A deletion study records} the rise of the counterweights it evaluates, scaling with $|\gamma_r|$ and an uncalibrated dose that varies across studies and random draws. For IOI, \chg{the heads of \citet{wang2022interpretability} that obey Eq.~\ref{eq:law} are all counterweights ($\gamma_r < 0$)} (App.~\ref{app:ioi}). The reported backup is the release of a coupling already active in the clean model.

\section{Related Work}
\label{sec:related}
\paragraph{Self-repair.}
In the IOI circuit, the backup name mover head engages, and the negative name movers' effect weakens when the name movers are ablated \citep{wang2022interpretability}. Layer-level compensation in Chinchilla~7B was referred to as the Hydra effect by  \citet{mcgrath2023hydraeffectemergentselfrepair}, where late MLPs erase the top token.   \citet{rushing2024explorationsselfrepairlanguagemodels} attributes the self-repair partly to LayerNorm and partly to sparse anti-erasure neurons. \citet{patrawala2025llm} shows that adjacent layers reduce the effect of predecessors. The OR-gate redundancy hides components from noising-based discovery \citep{chen2025rethinkingcircuitcompletenesslanguage}. Concurrently, \citet{gong2026conditional} retrieves backup components by measuring the increase in ablation after the primary set is removed.
These works mostly locate \chg{compensators after deletion. We show} the opposition is a standing coupling in the clean model, \chg{and that on IOI the seven reachable heads that obey our dose law are all counterweights.}
\paragraph{Suppressive components.}
Copy suppression \citep{mcdougall2023copysuppressioncomprehensivelyunderstanding}, suppression neurons \citep{gurnee2024universalneuronsgpt2language} and function-vector cancellers \citep{wang2026function} write against the output, and entropy neurons damp confidence \chg{\citep{gurnee2024universalneuronsgpt2language,stolfo2024confidenceregulationneuronslanguage}. Concurrently,} \citet{oskin2026lookingglassdirectlyreading} reports that across 18 models, the signed logit mass opposing the predicted token is a median of seven times the mass carrying it.
\chg{These are characterized by} their clean-pass sign, often via \chg{direct logit attribution. We measure} how a direction's \emph{total} effect varies with the dose of a specific core.
\paragraph{Interventions and weights.}
The choice of ablation changes localization and faithfulness \citep{zhang2024bestpracticesactivationpatching,heimersheim2024use,miller2024transformer}, motivating optimal ablation \citep{li2024optimal}\chg{. Subspace patching may act} through dormant parallel pathways \edit{\citep{makelov2023subspacelookingforinterpretability}}, a claim contested by \edit{\citep{wu2024reply}}.
\chg{Concurrently, \citet{vaidyanathan2026curse} show} that patching effects contain interaction terms that, on IOI, grow linearly with patch distance with \chg{role-dependent sign, and $\gamma_r$ is} a related, per-direction coupling along a controlled dose.
Composition scores \citep{elhage2021mathematical}, SVD communication channels \edit{\citep{merullo2024talking}}, \chg{and a concurrent weights-only map \citep{oskin2026lookingglassdirectlyreading} chart which components \emph{can} interact.}
We place every clamp of the core, including standard ablations, on one dose axis, and show that a weight-only statistic anticipates the rank of $|\gamma_r|$.

\section{Conclusion}
\label{sec:conclusion}

\paragraph{Findings.}In this work, we investigated the phenomenon of self-repair that confounds the interpretation of ablations. We uncovered weight-space components that are finer-grained than heads or MLPs and are coupled to one another along a class-contrast dose axis, thereby responsible for the compensatory effect. Conventional ablation methods, such as zero, mean, or resampling, are unpicked points on this axis. Therefore, what is regarded as self-repair is the passive release of a pre-existing coupling rather than a dynamic, reactive process.  Furthermore, this interaction is governed by an approximate affine law, whose slope can be anticipated from the alignment between downstream read and upstream write vectors. \chg{Finally, we showed that this coupling is robust, present in the base counterparts of the instruction models, and active under a procedurally unrelated, learned perturbation.}

\paragraph{Limitations.}
\chg{We studied one task at one answer position, a True/False verdict read at a single token.} How it may translate to other tasks, and especially generative ones, remains an open question. \chg{We found the cores, counterweights and relays among legible monosemantic weight-space directions, although nothing requires them to be legible.} \chg{Our experiments also suggest that a counterweight may respond to coupling from distributed components. In models with two cores,} the downstream counterweight responded more strongly to joint intervention than either core (App.~\ref{app:recruitment-svj}). \chg{Mapping these interactions exhaustively is left for future work.}

\paragraph{Reflection} \chg{The methodology underlying the critique our findings motivate generalizes further than the findings themselves.} The meta-conclusion of this work focuses on our instruments, \chg{which can become our own Eddington's net \citep{eddington2021philosophy}.} Our very tools and methods may be entangled with an existing computational structure within the model that no one yet knows, which eventually gives rise to a semblance of a phenomenon, as ablations did with self-repair. This applies to this work, too. We have used a tool to uncover what artifacts another tool can produce, and our findings are subject to the same scrutiny. We offer this work as a ladder to be climbed, for what it lets you see, and once you have seen, it should be thrown away \citep{wittgenstein2003tractatus}.

\section{Ethical Concerns}
\label{sec:ethics}
This work studies open-weight models on synthetic statements from a public corpus
and involves no human subjects or personal data.

\section*{Reproducibility Statement}
All models and the fact corpus are public. Setup and data are in
\S\ref{sec:setup} and App.~\ref{app:prompts}, the intervention in
\S\ref{sec:test}, and fitting and tests in App.~\ref{app:law-tests}. We will
release code and per-run outputs with the camera-ready version.

\appendix
\clearpage
\FloatBarrier
\section*{Contents}
\label{app:toc}
\addcontentsline{toc}{section}{Appendix}
\begingroup
\small
\noindent\textbf{\S\ref{app:guide}\quad Reader's Guide and Notation}\dotfill\pageref{app:guide}\\
\hspace{1.5em}\S\ref{app:guide-map}\quad What each section is evidence for\dotfill\pageref{app:guide-map}\\
\hspace{1.5em}\S\ref{app:guide-notation}\quad Notation\dotfill\pageref{app:guide-notation}\\
\vspace{3pt}
\noindent\textbf{\S{\ref{app:models_and_bases}}\quad Models and Weight-Space Bases}\dotfill{\pageref{app:models_and_bases}}\\
\vspace{3pt}
\noindent\textbf{\S\ref{app:task-data}\quad Task, Prompts, and Data}\dotfill\pageref{app:task-data}\\
\hspace{1.5em}\S\ref{app:prompts}\quad Template and answer tokens\dotfill\pageref{app:prompts}\\
\hspace{1.5em}\S\ref{app:data}\quad Data\dotfill\pageref{app:data}\\
\hspace{1.5em}\S\ref{app:sample-sizes}\quad Sample sizes across experiments\dotfill\pageref{app:sample-sizes}\\
\vspace{3pt}
\noindent\textbf{\S{\ref{app:census}}\quad {Weight-Space Census of the Bench}}\dotfill{\pageref{app:census}}\\
\hspace{1.5em}\S\ref{app:census-catalogues}\quad Catalogues\dotfill\pageref{app:census-catalogues}\\
\vspace{3pt}
\noindent\textbf{\S\ref{app:procedure}\quad {Regularity and Certification of the Second-Order Intervention}}\dotfill\pageref{app:procedure}\\
\hspace{1.5em}\S\ref{app:dose-regularity}\quad The dose axis is smooth across the full ladder\dotfill\pageref{app:dose-regularity}\\
\hspace{1.5em}\S\ref{app:certification-stats}\quad Certification statistics\dotfill\pageref{app:certification-stats}\\
\vspace{3pt}
\noindent\textbf{\S\ref{app:recruitment}\quad Recruitment Under the Dosed Swap}\dotfill\pageref{app:recruitment}\\
\hspace{1.5em}\S\ref{app:recruitment-tables}\quad Per-model recruitment tables\dotfill\pageref{app:recruitment-tables}\\
\hspace{1.5em}\S\ref{app:recruitment-figures}\quad Dose-resolved recruitment\dotfill\pageref{app:recruitment-figures}\\
\hspace{1.5em}\S\ref{app:recruitment-svj}\quad Single-core versus joint-core recruitment\dotfill\pageref{app:recruitment-svj}\\
\vspace{3pt}
\noindent\textbf{\S\ref{app:perturbation}\quad {The Learned Perturbation as an Independent Primary}}\dotfill\pageref{app:perturbation}\\
\hspace{1.5em}\S\ref{app:perturbation-objective}\quad Objective and training\dotfill\pageref{app:perturbation-objective}\\
\hspace{1.5em}\S\ref{app:perturbation-results}\quad Results\dotfill\pageref{app:perturbation-results}\\
\hspace{1.5em}\S\ref{app:convergence}\quad Three-route convergence\dotfill\pageref{app:convergence}\\
\vspace{3pt}
\noindent\textbf{\S\ref{app:pretraining}\quad {Pretraining-Native Recruitment in Base versus Instruct}}\dotfill\pageref{app:pretraining}\\
\vspace{3pt}
\noindent\textbf{\S\ref{app:renorm}\quad Recruitment Is Not Renormalization}\dotfill\pageref{app:renorm}\\
\hspace{1.5em}\S\ref{app:renorm-objection}\quad The objection\dotfill\pageref{app:renorm-objection}\\
\hspace{1.5em}\S\ref{app:renorm-predicts}\quad What the objection predicts\dotfill\pageref{app:renorm-predicts}\\
\hspace{1.5em}\S\ref{app:renorm-measurement}\quad Measurement\dotfill\pageref{app:renorm-measurement}\\
\hspace{1.5em}\S\ref{app:renorm-worlds}\quad {Class-resolved margins in the worlds the audit indexes}\dotfill\pageref{app:renorm-worlds}\\
\hspace{1.5em}\S\ref{app:renorm-refutation}\quad The refutation\dotfill\pageref{app:renorm-refutation}\\
\hspace{1.5em}\S\ref{app:renorm-expansion}\quad Why an inversion-type primary need not shrink the stream\dotfill\pageref{app:renorm-expansion}\\
\hspace{1.5em}\S\ref{app:renorm-scope}\quad What this establishes, and what it leaves open\dotfill\pageref{app:renorm-scope}\\
\vspace{3pt}
\vspace{3pt}
\noindent\textbf{\S\ref{app:coupling-fits}\quad {Fits and Diagnostics of the Coupling Law}}\dotfill\pageref{app:coupling-fits}\\
\hspace{1.5em}\S\ref{app:fits}\quad Estimation procedure\dotfill\pageref{app:fits}\\
\hspace{1.5em}\S\ref{app:coupling-family}\quad Family-level summary\dotfill\pageref{app:coupling-family}\\
\hspace{1.5em}\S\ref{app:coupling-perdirection}\quad Per-direction fits\dotfill\pageref{app:coupling-perdirection}\\
\hspace{1.5em}\S\ref{app:fit-diagnostics}\quad Fit diagnostic figures\dotfill\pageref{app:fit-diagnostics}\\
\hspace{1.5em}\S\ref{app:coupling-holdout}\quad Exemplar fits and the half-rung holdout\dotfill\pageref{app:coupling-holdout}\\
\vspace{3pt}
\noindent\textbf{\S\ref{app:law-tests}\quad {Tests of the Coupling Law}}\dotfill\pageref{app:law-tests}\\
\vspace{3pt}
\noindent\textbf{\S\ref{app:ioi}\quad {Out-of-Domain Test on GPT-2 Small and Indirect-Object Identification}}\dotfill\pageref{app:ioi}\\
\vspace{3pt}
\noindent\textbf{\S\ref{app:wiring}\quad {The Wiring Statistic as a Weight-Space Predictor of Coupling}}\dotfill\pageref{app:wiring}\\
\hspace{1.5em}\S\ref{app:wiring-family}\quad Family breakdown\dotfill\pageref{app:wiring-family}\\
\hspace{1.5em}\S\ref{app:wiring-placement}\quad {The placement control on directions the core cannot reach}\dotfill\pageref{app:wiring-placement}\\
\vspace{3pt}
\noindent\textbf{\S\ref{app:deletion}\quad {Derivations Showing That Every Clamp Is a Dose}}\dotfill\pageref{app:deletion}\\
\vspace{3pt}
\noindent\textbf{\S\ref{app:general}\quad {Beyond a Binary Contrast}}\dotfill\pageref{app:general}\\
\vspace{3pt}
\noindent\textbf{\S\ref{app:conjecture}\quad {A Conjecture That Every Design Is a Point in the $(u,v)$ Plane}}\dotfill\pageref{app:conjecture}\\
\endgroup
\vspace{8pt}
\hrule
\vspace{10pt}

\clearpage
\FloatBarrier
\section{Reader's Guide and Notation}
\label{app:guide}

\subsection{What each section is evidence for}
\label{app:guide-map}

\begin{table}[!htbp]
\centering\small
\begin{tabular}{@{}p{0.30\linewidth}p{0.62\linewidth}@{}}
\toprule
Main-text claim & Supporting appendix section \\
\midrule
{Contribution 1 (compensation is standing)} & {\S\ref{app:recruitment} (recruitment), \S\ref{app:perturbation} (other primaries), \S\ref{app:pretraining} (base checkpoints), \S\ref{app:renorm} (not a LayerNorm artefact)} \\
{Contribution 2 (a dose axis and a nested intervention)} & {\S\ref{app:procedure}, \S\ref{app:deletion}} \\
{Contribution 3 (the coupling has an affine form)} & {\S\ref{app:coupling-fits} and \S\ref{app:law-tests}, with IOI in \S\ref{app:ioi}} \\
{Contribution 4 (named counterweights and relays, coupling legible in the weights)} & {\S\ref{app:census} (catalogues, clean census), \S\ref{app:wiring}} \\
\bottomrule
\end{tabular}
\caption{Claim-to-evidence map. Each row names the section that carries the primary {burden of proof. Supporting figures/tables} for a claim may also appear incidentally elsewhere.}
\label{tab:claim-map}
\end{table}

\subsection{Notation}
\label{app:guide-notation}

\begin{table}[!htbp]
\centering\small
\begin{tabular}{@{}lp{0.75\linewidth}@{}}
\toprule
Symbol & Meaning \\
\midrule
$x$ & The prompt (SCM exogenous variable). \\
$x'$ & The matched counterfactual partner of $x$ (opposite class, identical phrasing). \\
$b$ & World index: $b=\varnothing$ (clean), $b=\pi_k$ (core dosed to $k${; the main text writes $\pi_\lambda$ for the same world}), $b=\delta$ (learned perturbation). \\
$r$ & A weight-space direction (an \texttt{mlp\_neuron}, \texttt{ov\_neuron}, \texttt{ov\_svd}, or \texttt{mlp\_svd} unit; \S{\ref{app:models_and_bases}}). \\
$\omega_r$ & $r$'s fixed write vector. \\
$a_r(x;b)$ & $r$'s activation coefficient on prompt $x$ in world $b$. \\
$m, \Delta$ & Pair mean $\tfrac12(a_T+a_F)$ and contrast $a_F-a_T$ of the core's clean activations on a matched pair. \\
$\lambda$ & \textbf{The dose coordinate.} $\lambda(v) = 2(m-v)/\Delta$; $\lambda=+1$ clean, $\lambda=-1$ full counterfactual inversion, $\lambda=0$ class-neutral midpoint. \\
$k$ & Corruption fraction, $k=(1-\lambda)/2$. {The ladder of \S\ref{sec:test} is $k\in\{0,\tfrac14,\tfrac12,\tfrac34,1,2,3,4\}$; recruitment tables label $k=n$ as $\times n$.} $k\le1$ is on-manifold, $k>1$ is overshoot/extrapolation. \\
$E^b_r(x)$ & $r$'s causal contribution in world $b$: the verdict margin erased by overriding $r$'s activation with its regime-consistent counterfactual (main text Eq.~{\ref{eq:reference}}). \\
$R_r$ & Recruitment, $E^{\pi}_r - E^0_r$. \\
{$D_{\mathrm{TF}}$} & {Class-fixed verdict margin, True minus False tokens; $D_{XY}$ of Eq.~\ref{eq:margin} equals $D_{\mathrm{TF}}$ on true prompts and $-D_{\mathrm{TF}}$ on false ones.} \\
$\mathrm{own}_r, \gamma_r$ & Intercept and slope of the affine law $E^c_r(\lambda) = \mathrm{own}^c_r + \gamma_r\lambda$ (main text Eq.~{\ref{eq:law}}). $\gamma_r<0$: counterweight; $\gamma_r>0$: relay. \\
$z_r$ & Weight-only wiring-alignment statistic (\S\ref{app:wiring}) predicting $|\gamma_r|$ from the static checkpoint. \\
$\rho_b(x)$ & Answer-position RMS of the pre-unembedding stream in world $b$ (\S\ref{app:renorm}). \\
$g_k$ & Final-norm gain, $\rho_\varnothing/\rho_{\pi_k}$. \\
\bottomrule
\end{tabular}
\caption{Notation used throughout the appendix. {The dose is always $\lambda$, and $x$ always denotes a prompt.}}
\label{tab:notation}
\end{table}

\clearpage
\FloatBarrier
\section{Models and Weight-Space Bases}
\label{app:models_and_bases}

\begin{table}[!htbp]
\centering
\small
\setlength{\tabcolsep}{8pt}
\renewcommand{\arraystretch}{1.25}
\begin{tabular}{@{}l l l l@{}}
\toprule
\multicolumn{4}{@{}l@{}}{  \footnotesize Gated MLP, neuron $j$ of layer $\ell$:\quad
  $g_j=\mathbf{w}^{\mathrm{gate}}_j\!\cdot x$\quad
  $u_j=\mathbf{w}^{\mathrm{up}}_j\!\cdot x$\quad
  $\mathrm{MLP}(x)=\sum_j a_j(x)\,W_{\mathrm{down}}[:,j]$}\\
\addlinespace[2pt]
\midrule
Model & MLP input $x$ & Neuron scalar $a_j(x)$ & Residual write \\
\midrule
Llama-3-8B-Instruct & $\mathrm{RMSNorm}_{\mathrm{post\text{-}attn}}(h)$ & $\mathrm{SiLU}(g_j)\,u_j$ & $W_{\mathrm{down}}[:,j]$ \\
Gemma-2-9b-it & $\mathrm{RMSNorm}_{\mathrm{pre\text{-}ffw}}(h)$ & $\mathrm{GELU}_{\tanh}(g_j)\,u_j$ & $\boldsymbol{\gamma}_{\mathrm{post}}\odot W_{\mathrm{down}}[:,j]$ \\
Qwen2.5-7B-Instruct & $\mathrm{RMSNorm}_{\mathrm{post\text{-}attn}}(h)$ & $\mathrm{SiLU}(g_j)\,u_j$ & $W_{\mathrm{down}}[:,j]$ \\
Mistral-7B-Instruct-v0.3 & $\mathrm{RMSNorm}_{\mathrm{post\text{-}attn}}(h)$ & $\mathrm{SiLU}(g_j)\,u_j$ & $W_{\mathrm{down}}[:,j]$ \\
\bottomrule
\end{tabular}
\caption{\textbf{The \texttt{mlp\_neuron} scalar, per model.} {Every intervention in this paper acts on $a_j$, which is coordinate $j$ of the \texttt{down\_proj} input and the scalar multiplying that neuron's fixed write vector.} All four models use a gated MLP, and in all four, {the up branch is \emph{linear}. Only the gate} passes through the nonlinearity, so the scalar is $\sigma(g_j)\,u_j$ and never $\sigma(g_j)\,\sigma(u_j)$.}
\label{tab:mlp-neuron-forms}
\end{table}

The other three weight-space bases used to assemble the bench {(\texttt{ov\_neuron} is a single output channel of an attention head's $W_{O}$, \texttt{ov\_svd} a left-singular direction of a head's OV circuit, and \texttt{mlp\_svd} a left-singular direction of an MLP's down-projection)} follow the same interface{ (}a fixed write vector $\omega_r$ and a scalar activation coefficient $a_r(x)${)} with the coefficient read from the corresponding decomposition rather than a single \texttt{down\_proj} coordinate. Main text \S\ref{sec:setup} {gives the shared interface, and Table~\ref{tab:mlp-neuron-forms} exists} because the neuron basis is the one whose scalar is model-family-dependent, per the divergence above.

\FloatBarrier
\section{Task, Prompts, and Data}
\label{app:task-data}

\subsection{Template and answer tokens}
\label{app:prompts}

Every prompt is built by {substituting the statement into the fixed template below.}

\begin{quote}\ttfamily
Given the following statement, answer if it is True or False.\\[2pt]
Statement: \{q\}\\
Answer:
\end{quote}

\noindent
where \texttt{\{q\}} is the statement and the literal newlines are as shown (a
blank line after the instruction, none before \texttt{Answer:}). The prompt is
tokenised with the model's own tokenizer, with its BOS token but \emph{without}
{a chat template. The instruct models} are used as plain conditional language
models, so that the evaluation regime and the intervention regime coincide. All
readouts are taken at the final position, i.e.\ immediately after
\texttt{Answer:}, so the next token is the verdict word.

The verdict is read from two sets of vocabulary items rather than a single
token, so that the measurement does not depend on which capitalisation or
leading-space form a given tokenizer {happens to prefer. The two sets are}
{\begin{align*}
  \mathcal{T}&=\{\texttt{"\_true"},\ \texttt{"\_True"},\ \texttt{"\_TRUE"},\
                 \texttt{"true"},\ \texttt{"True"}\},\\
  \mathcal{F}&=\{\texttt{"\_false"},\ \texttt{"\_False"},\ \texttt{"\_FALSE"},\
                 \texttt{"false"},\ \texttt{"False"}\},
\end{align*}}
writing \texttt{\_} for a leading space. Two conventions matter for
reproducibility. {\textbf{Single-token filtering.} A surface form is retained
only if the model's tokenizer encodes it as exactly one token. Multi-token
forms} are dropped, since a form with no column in the unembedding cannot
contribute a logit. The retained subset is therefore tokenizer-dependent and
recorded in each run's metadata. {\textbf{Means rather than sums.}} $D_{\mathrm{TF}}$
averages within each set before subtracting, so the scale is comparable across
models even when the two sets survive filtering at different sizes. Logits are used raw {(no softmax, no temperature)} because the interventions are
linear perturbations of the residual stream and a normalisation would make the
measured effect depend on the rest of the vocabulary.

\subsection{Data}
\label{app:data}

\paragraph{Source.}
Statements derive from the \textbf{Known-Facts} corpus released with ROME
\citep{meng2022locating}, obtained from
\url{https://rome.baulab.info/data/dsets/known_1000.json} (1209 records, 1208
usable). Each record is a (subject, relation, attribute) triple with a
{cloze template, and we render} a claim as
\texttt{template.format(subject) + " " + attribute + "."}, i.e.\ a declarative
sentence with a terminal period. The corpus ships a pre-curated \texttt{prompt}
field that extends some templates {with additional text. We do not use it,}
because those extensions are not well-formed for every record.

\paragraph{{Synthesized false counterparts.}}
Known-Facts contains \emph{only true} facts, so it supplies no negative side.
For each record, we synthesise a false claim by drawing a type-matched
alternative attribute from the pool of attributes attested for that same
relation, excluding the record's own attribute. The true and false members of a
pair therefore share subject, relation, and surface template, and differ only in
the asserted attribute{,} the minimal edit the contrastive interventions
require. Sampling is deterministic given the build seed, so the pairing is
reproducible. Only the true side is {attributable to \citet{meng2022locating}.
The counterfactual side} is constructed here, and any error in it is ours.

\paragraph{Splits.}
Records are partitioned 70/10/20 into train/validation/test, stratified by
relation so that every split contains every relation, and with no record
appearing in more than one split (Splits are only used in learned steering perturbation  \ref{app:perturbation}).

\paragraph{Filtering to clean-correct pairs.}
The corpus is not filtered by model behaviour at build time. The restriction
used throughout the paper is applied afterward, {and per model. A pair is}
retained iff $D_{\mathrm{TF}}>0$ on its true member \emph{and}
$D_{\mathrm{TF}}<0$ on its false member, evaluated zero-shot with the template
of \S\ref{app:prompts}. Because this depends on the model, the retained
subset differs across the four models.

\subsection{Sample sizes across experiments}
\label{app:sample-sizes}

The source appendix reported $n$ in each table/figure caption individually,
which makes it hard to see at a glance whether a given comparison used a
matched subsample or the full retained pool. Table~\ref{tab:sample-sizes}
consolidates every $n$ that appears in this appendix.

\begin{table}[!htbp]
\centering\small
\begin{tabular}{@{}l r l@{}}
\toprule
Experiment & $n$ & Section \\
\midrule
Llama single-core swap (\texttt{L18.n11065}) & 532 & \S\ref{app:recruitment} \\
Gemma joint-core swap (\texttt{L28.n2046+L33.n4294}) & 583 & \S\ref{app:recruitment} \\
Qwen joint-core swap (\texttt{L22.n13149+L24.n14758}) & 478 & \S\ref{app:recruitment} \\
Mistral joint-core swap (\texttt{L20.n14286+L19.n8228}) & 598 & \S\ref{app:recruitment} \\
Gemma / Qwen dose-resolved single-core comparisons & 100 & \S\ref{app:recruitment} \\
Learned perturbation (\texttt{perfect\_on\_pairs}, all models) & -- (full pool) & \S\ref{app:perturbation} \\
Base-vs-instruct, Llama & 240 (shared, both checkpoints correct) & \S\ref{app:pretraining} \\
Base-vs-instruct, Gemma & 535 & \S\ref{app:pretraining} \\
Base-vs-instruct, Qwen & 470 & \S\ref{app:pretraining} \\
Base-vs-instruct, Mistral & 423 & \S\ref{app:pretraining} \\
Final-norm gain audit, per family & 50 matched pairs & \S\ref{app:renorm} \\
Half-rung holdout ($\lambda=\pm\tfrac12$), per family & 478--598 (same pop.\ as ladder) & \S\ref{app:coupling-fits} \\
\bottomrule
\end{tabular}
\caption{Sample sizes referenced in this appendix, consolidated from individual table/figure captions.}
\label{tab:sample-sizes}
\end{table}
\begin{CJK*}{UTF8}{gbsn}
\FloatBarrier
\section{{Weight-Space Census of the Bench}}
\label{app:census}

\subsection{Scan protocol}
Directions are named from the weights {before any intervention. Scanning the}
four model families (main text \S{\ref{sec:setup}}), each direction $r$ is labelled by the top
and bottom tokens of $\omega_r^\top W_U$ and retained if its pole represents
words corresponding to truth or falsity (\emph{lie, truth, wrong, sure,
illusion}), including cross-lingual matches pairing English true/false with
other counterparts. Selection relies {solely on static weights. Whether a}
retained direction {contributes}, opposes, or is dormant for the
realised verdict is a separate, causal question, answered in
\S\ref{app:recruitment}.

\subsection{Catalogues}
\label{app:census-catalogues}

{Tables~\ref{tab:llama_poles} to \ref{tab:mistral_verdict_catalogue} give the}
weight-fixed, intervention-independent poles for the benched directions in
each model.

\begin{table}[!htbp]\centering\footnotesize\setlength{\tabcolsep}{5pt}
\rowcolors{3}{gray!4}{white}
\caption{\textbf{Vocabulary poles of the recruited directions in Llama-3-8B-Instruct}
($\omega_r^{\!\top} W_U$, weight-fixed, {intervention-independent, with the top-$4$ promoted and the
bottom-$4$ suppressed)}. \textcolor{TrueC}{\textbf{Green}} marks truth-coding tokens,
\textcolor{FalseC}{\textbf{red}} falsity-coding tokens, \textcolor{Junk}{gray}
non-interpretable subword fragments.}
\label{tab:llama_poles}
\begin{tabular}{>{\ttfamily}l >{\ttfamily\footnotesize\raggedright\arraybackslash}p{4.8cm}
                   >{\ttfamily\footnotesize\raggedright\arraybackslash}p{4.8cm}}
\toprule
\normalfont Axis & \normalfont Promoted (top-4) & \normalfont Suppressed (bottom-4)\\
\midrule
L23.n8972     & \jk{iller, rello, LC, etz}              & \tFa{lie}, \tFa{lies}, \tFa{lying}, \tFa{Lie}\\
L19.n2738     & \jk{ledged, Alo, ADDE}, \cjkjp{くだ}{Jp.}  & \tFa{illusion}, \tFa{illusions}, \tFa{false}, \tFa{illusion}\\
L23.n9811     & \tTr{truth}, \tTr{Truth}, \tTr{truth}, \tTr{Truth} & \jk{oby, imore, Tell, tell}\\
L19.n3754     & \jk{Edition, onte}, \cjkjp{フ}{Jp.}, \jk{rale} & \tTr{sure}, \tTr{yes}, \tTr{yes}, \tTr{Sure}\\
L22.h1.svd117 & \jk{miss, missing, chances, misses}     & \tTr{true}, \tTr{true}, \jk{Important}, \tTr{TRUE}\\
L24.n4543     & \tTr{yes}, \tTr{true}, \tTr{True}, \tTr{Yes} & \jk{OrNil, uelle, illard, izh\,{\text{\scriptsize(Cyr.)}}}\\
L21.n798      & \jk{Arena, quer, KER, .BorderFactory}   & \tTr{true}, \jk{is}, \tTr{True}, \tTr{true}\\
L21.n4049     & \jk{642, upe, airo}, \cjkjp{立て}{Jp.}     & \tFa{wrong}, \tFa{wrong}, \tFa{Wrong}, \tFa{WRONG}\\
L20.h11.c95   & \jk{reversal, reversed, loss, ercul}    & \tFa{false}, \tFa{false}, \tFa{False}, \cjkjp{释}{explain}\\
L20.n14044    & \jk{lagod\,\text{\scriptsize(Cyr.)}, IGGER, /generated} & \tTr{truly}, \tTr{true}, \tTr{really}, \cjkjp{真正}{\textbf{genuine}}\\
L22.h1.svd116 & \jk{fun, Needs, needs, hopes}           & \tTr{correct}, \tTr{legitimate}, \cjkjp{正确}{\textbf{correct}}, \tTr{legitimacy}\\
L28.h27.c34   & \jk{total, total, Total, TOTAL}         & \tTr{True}, \tTr{true}, \jk{fully}, \tTr{True}\\
L28.h27.c120  & \jk{about, about}, \tTr{True}, \jk{truly} & \tTr{real}, \jk{near, Near}, \tTr{Real}\\
L21.n14165    & \tTr{Genuine}, \tTr{authentic}, \tTr{genuine}, \tTr{pure} & \jk{atorio, čet, strcasecmp, yh}\\
\midrule
\rowcolor{black!6}
L18.n11065\,$^{\dagger}$ & \tTr{true}, \tTr{yes}, \tTr{YES}, \jk{.YES} & \tFa{false}, \tFa{False}, \tFa{false}, \tFa{False}\\
\bottomrule
\end{tabular}
\\[2pt]
{\footnotesize {$^{\dagger}$Core. Its full $k\!=\!6$ lists are \texttt{true, yes, YES, .YES, Yes, TRUE} (promoted)
and \texttt{false, False, false, False, FALSE, (false} (suppressed).}}
\end{table}

\begin{table}[!htbp]\centering\footnotesize\setlength{\tabcolsep}{5pt}
\rowcolors{2}{gray!4}{white}
\caption{{\textbf{Logit-lens poles of the recruited Gemma-2-9b-it directions and the two cores} ($\omega_r\!\cdot\!W_U$, weight-fixed, intervention-independent, with the top-4 promoted and the bottom-4 suppressed)}. \textcolor{TrueC}{\textbf{Green}}$=$truth-coding, \textcolor{FalseC}{\textbf{red}}$=$falsity-coding, \textcolor{Junk}{gray}$=$non-interpretable subword. {Most recruited anti-directions carry} a clean English truth/falsity pole on one side against subword noise {on the other. The core \texttt{L28.n2046} carries} the cross-lingual \cjkjp{否}{no} beside English \emph{yes/true}.}
\label{tab:gemma_swap_poles}
\begin{tabular}{l >{\ttfamily\footnotesize\raggedright\arraybackslash}p{5.0cm} >{\ttfamily\footnotesize\raggedright\arraybackslash}p{5.0cm}}
\toprule
\normalfont Axis & \normalfont Promoted (top-4) & \normalfont Suppressed (bottom-4)\\
\midrule
\texttt{L38.n7088} & \tFa{wrong}, \tFa{wrong}, \tFa{Wrong}, \tFa{Wrong} & \jk{\_\_).}, \jk{BarStyle}, \jk{preferably}, \jk{AutoSize}\\
\texttt{L39.n10085} & \tFa{fake}, \tFa{Fake}, \tFa{pseudo}, \tFa{pseud} & \jk{argout}, \jk{cyd}, \jk{MessageOf}, \jk{MigrationBuilder}\\
\texttt{L38.n854} & \jk{ModelExpression}, \jk{\_\_":}, \jk{ituary}, \jk{orghini} & \tTr{true}, \tTr{true}, \tTr{True}, \tTr{True}\\
\texttt{L41.n8771} & \jk{SOUNDBITE}, \jk{estekak}, \jk{protoimpl}, \jk{ChrTalk} & \tTr{True}, \tTr{true}, \tTr{True}, \tTr{true}\\
\texttt{L39.n10210} & \tFa{incorrect}, \tFa{incorrect}, \tFa{Incorrect}, \tFa{WRONG} & \jk{IBOutlet}, \jk{setopt}, \jk{openzeppelin}, \jk{Geos}\\
\texttt{L41.n3789} & \jk{haikusbot}, \jk{writeFieldEnd}, \jk{complexType}, \jk{getDescription} & \tTr{real}, \jk{skuto}, \tTr{truly}, \tTr{genuine}\\
\texttt{L32.n8200} & \jk{LookAnd}, \jk{GEBURTS}, \jk{ItemBackground}, \jk{\textless{}bos\textgreater{}} & \tFa{False}, \tFa{false}, \tFa{False}, \tTr{True}\\
\texttt{L33.n1466} & \tFa{deception}, \tFa{fake}, \tFa{deceiving}, \tFa{deceptive} & \jk{rungsseite}, \jk{createSlice}, \jk{surate}, \jk{photobucket}\\
\texttt{L34.n4784} & \jk{Хьажоргаш\,{\scriptsize(Cyr.)}}, \jk{TypedDataSet}, \jk{RegressionTest}, \jk{pasto} & \jk{tagHelperRunner}, \jk{nahilalakip}, \tTr{truth}, \tTr{truths}\\
\texttt{L40.h11.svd18} & \tTr{truth}, \jk{tissue}, \jk{traffic}, \jk{bishop} & \jk{GEBURTSDATUM}, \jk{StoryboardSegue}, \jk{ModelExpression}, \jk{SourceChecksum}\\
\texttt{L40.h7.c177} & \jk{brekdaun\,{\scriptsize(Dev.)}}, \jk{Upper}, \jk{Upper}, \jk{Continuous} & \jk{DebuggerNonUser}, \jk{$\varnothing$}, \jk{StoryboardSegue}, \jk{perfect}\\
\texttt{L40.h7.c105} & \tTr{real}, \tTr{Real}, \tTr{Real}, \tTr{real} & \jk{principalColumn}, \jk{déf}, \jk{PerformLayout}, \jk{\textless{}\textgreater{}",}\\
\midrule
\rowcolor{black!6}
\texttt{L33.n4294} & \jk{InputDecoration}, \tTr{true}, \jk{jMenuItem}, \tTr{true} & \tFa{wrong}, \tFa{wrong}, \tFa{Wrong}, \tFa{Wrong}\\
\rowcolor{black!6}
\texttt{L28.n2046} & \cjkjp{否}{no}, \tFa{Nope}, \tFa{nope}, \tFa{NOPE} & \tTr{yes}, \tTr{true}, \tTr{TRUE}, \tTr{Yes}\\
\bottomrule\end{tabular}\end{table}

\begin{table}[!htbp]\centering\footnotesize\setlength{\tabcolsep}{5pt}
\rowcolors{2}{gray!4}{white}
\caption{{\textbf{A catalogue of interpretable truth/falsity directions in Qwen2.5-7B-Instruct} ($\omega_r\!\cdot\!W_U$, weight-fixed, intervention-independent, with the top-4 promoted and the bottom-4 suppressed)}. \textcolor{TrueC}{\textbf{Green}}$=$truth-coding, \textcolor{FalseC}{\textbf{red}}$=$falsity-coding, \textcolor{Junk}{gray}$=$non-interpretable subword. As in Gemma-2-9b and Llama-3-8b, Qwen carries a large population of directions whose unembedding poles {encode} the True/False verdict for this task ($6{,}310$ marked across the four {component families). A representative sample spanning MLP neurons (layers L20 to L27)} and attention (OV) axes is listed by $|\!\cos(\omega_r,w_{TF})|$. Coding is markedly cross-lingual, pairing Chinese \cjkjp{真}{true}/\cjkjp{错}{wrong}/\cjkjp{假}{fake}/\cjkjp{虚假}{false} with English \emph{true/false/fake}. Whether each contributes to, opposes, or is dormant for the realized verdict is measured separately ( \S\ref{app:recruitment}).}
\label{tab:qwen_verdict_catalogue}
\begin{tabular}{l >{\ttfamily\footnotesize\raggedright\arraybackslash}p{5.0cm} >{\ttfamily\footnotesize\raggedright\arraybackslash}p{5.0cm}}
\toprule
\normalfont Axis & \normalfont Promoted (top-4) & \normalfont Suppressed (bottom-4)\\
\midrule
\texttt{L27.n4228} & \jk{$\varnothing$}, \jk{-Headers}, \jk{ómo}, \jk{'\}\}\textgreater{}} & \tTr{truly}, \cjkjp{真正}{genuine}, \tTr{true}, \cjkjp{真正的}{genuine}\\
\texttt{L22.n13149} & \tTr{true}, \tTr{True}, \jk{\_true}, \tTr{true} & \tFa{false}, \tFa{False}, \tFa{false}, \tFa{False}\\
\texttt{L20.n2073} & \tFa{False}, \tFa{false}, \tFa{False}, \tFa{false} & \tTr{true}, \tTr{True}, \tTr{true}, \tTr{True}\\
\texttt{L22.n4538} & \tFa{false}, \tFa{false}, \tFa{False}, \tFa{False} & \tTr{true}, \tTr{True}, \tTr{True}, \tTr{true}\\
\texttt{L27.n13033} & \jk{fa}, \jk{ase}, \jk{л\,{\scriptsize(Cyr.)}}, \jk{fa} & \tFa{false}, \tFa{False}, \tFa{False}, \tFa{false}\\
\texttt{L24.n3278} & \tTr{true}, \tTr{true}, \cjkjp{真实}{real}, \tTr{True} & \jk{ulfill}, \jk{.Gray}, \jk{naz}, \jk{泄露\,{\scriptsize(Zh.)}}\\
\texttt{L25.n4929} & \jk{assertNotNull}, \jk{seedu}, \jk{reeNode}, \jk{addtogroup} & \tFa{false}, \tFa{false}, \tFa{False}, \tFa{False}\\
\texttt{L25.n11565} & \tTr{true}, \tTr{true}, \cjkjp{真}{true}, \tTr{True} & \jk{commodo}, \jk{情人\,{\scriptsize(Zh.)}}, \jk{ollah}, \jk{蘼\,{\scriptsize(Zh.)}}\\
\texttt{L22.n9820} & \tTr{yes}, \tTr{sure}, \tTr{Yes}, \tTr{Sure} & \jk{YO}, \jk{áll}, \jk{ublish}, \jk{.Itoa}\\
\texttt{L24.n14758} & \tTr{right}, \tTr{right}, \tTr{Right}, \tTr{Right} & \cjkjp{错}{wrong}, \cjkjp{錯}{wrong}, \jk{$\varnothing$}, \cjkjp{错误}{error}\\
\texttt{L27.n6739} & \tFa{none}, \tFa{none}, \tFa{None}, \tFa{None} & \jk{;line}, \jk{这两个\,{\scriptsize(Zh.)}}, \jk{oriously}, \jk{odox}\\
\texttt{L23.n3002} & \jk{,\},}, \jk{(',',\$}, \jk{akest}, \jk{TMPro} & \tTr{true}, \tTr{true}, \jk{(true}, \tTr{True}\\
\texttt{L24.n9321} & \jk{ORITY}, \jk{twor}, \jk{kowski}, \jk{akthar\,{\scriptsize(Ar.)}} & \cjkjp{真正的}{genuine}, \cjkjp{真正}{genuine}, \tTr{true}, \tTr{genuine}\\
\texttt{L22.n9609} & \tFa{False}, \tFa{false}, \tFa{false}, \tFa{False} & \jk{glEnable}, \cjkjp{阳性}{positive}, \cjkjp{正品}{genuine}, \jk{顺\,{\scriptsize(Zh.)}}\\
\texttt{L23.n7028} & \jk{.Itoa}, \jk{机动\,{\scriptsize(Zh.)}}, \jk{{\CJKfamily{mj}싼}}, \jk{StringBuffer} & \tTr{yes}, \tTr{Yes}, \tTr{Yes}, \tTr{yes}\\
\texttt{L25.n8444} & \jk{seedu}, \jk{Decompiled}, \jk{anes}, \jk{.uf} & \cjkjp{假}{fake}, \tFa{fake}, \cjkjp{伪}{pseudo}, \jk{pseud}\\
\texttt{L23.n14476} & \jk{ournemouth}, \jk{ulfill}, \jk{indsight}, \jk{になれ\,{\scriptsize(Jp.)}} & \cjkjp{真正的}{genuine}, \cjkjp{真实}{real}, \tTr{real}, \jk{真实的\,{\scriptsize(Zh.)}}\\
\texttt{L22.n13025} & \jk{:".\$}, \jk{㎞}, \jk{.IsNullOr}, \jk{:normal} & \cjkjp{真正}{genuine}, \cjkjp{真正的}{genuine}, \tTr{truly}, \tTr{true}\\
\texttt{L23.n8341} & \jk{.getOwnProperty}, \jk{Hath}, \jk{眉\,{\scriptsize(Zh.)}}, \jk{{\CJKfamily{mj}렵}} & \tFa{illusion}, \jk{Ill}, \tFa{illusions}, \jk{幻\,{\scriptsize(Zh.)}}\\
\texttt{L24.n8308} & \tTr{yes}, \tTr{Yes}, \tTr{yes}, \tTr{Yes} & \jk{uk}, \jk{@\$}, \jk{究竟\,{\scriptsize(Zh.)}}, \jk{到底\,{\scriptsize(Zh.)}}\\
\texttt{L24.n12848} & \cjkjp{虚假}{false}, \tFa{false}, \cjkjp{假}{fake}, \tFa{fake} & \jk{苡\,{\scriptsize(Zh.)}}, \jk{nieu}, \jk{Ổ}, \jk{[Byte}\\
\texttt{L24.n5510} & \tFa{none}, \tFa{none}, \tFa{None}, \tFa{None} & \jk{这两个\,{\scriptsize(Zh.)}}, \tFa{no}, \jk{,no}, \jk{不再\,{\scriptsize(Zh.)}}\\
\texttt{L23.h27.a9} & \tTr{correct}, \cjkjp{正确}{correct}, \jk{$\varnothing$}, \cjkjp{正确的}{correct} & \jk{.Empty}, \cjkjp{无关}{unrelated}, \jk{empty}, \jk{Empty}\\
\texttt{L23.h27.c88} & \cjkjp{正确}{correct}, \tTr{correct}, \jk{:[[}, \tTr{Correct} & \jk{不可以\,{\scriptsize(Zh.)}}, \jk{可以直接\,{\scriptsize(Zh.)}}, \jk{egot}, \jk{不可\,{\scriptsize(Zh.)}}\\
\texttt{L26.h17.{svd}117} & \jk{.Active}, \jk{active}, \cjkjp{真}{true}, \jk{active} & \jk{base}, \jk{internal}, \cjkjp{内部}{internal}, \jk{基\,{\scriptsize(Zh.)}}\\
\texttt{L21.n17369} & \jk{"Yes}, \tTr{Yes}, \jk{“Yes}, \tTr{Yes} & \jk{Nor}, \tFa{Nope}, \jk{-not}, \jk{nor}\\
\texttt{L23.h27.c82} & \tTr{correct}, \tTr{true}, \cjkjp{正确}{correct}, \jk{know} & \jk{al-ithn\,{\scriptsize(Ar.)}}, \jk{.Undef}, \jk{"nil}, \jk{\_TypeInfo}\\
\bottomrule\end{tabular}\end{table}

\begin{table}[!htbp]\centering\footnotesize\setlength{\tabcolsep}{5pt}
\rowcolors{2}{gray!4}{white}
\caption{{\textbf{A catalogue of interpretable truth/falsity directions in Mistral-7B-Instruct-v0.3} ($\omega_r\!\cdot\!W_U$, weight-fixed, intervention-independent, with the top-4 promoted and the bottom-4 suppressed)}. \textcolor{TrueC}{\textbf{Green}}$=$truth-coding, \textcolor{FalseC}{\textbf{red}}$=$falsity-coding, \textcolor{Junk}{gray}$=$non-interpretable subword. All 35 bench directions are listed, {ordered by layer. Of these, 14 carry} a clean bipolar verdict axis and the rest carry a clean pole on one side against subword noise on the other. $\dagger$ marks the {core(s)} used as the primary intervention.}
\label{tab:mistral_verdict_catalogue}
\begin{tabular}{l >{\ttfamily\footnotesize\raggedright\arraybackslash}p{5.0cm} >{\ttfamily\footnotesize\raggedright\arraybackslash}p{5.0cm}}
\toprule
\normalfont Axis & \normalfont Promoted (top-4) & \normalfont Suppressed (bottom-4)\\
\midrule
\texttt{L15.n11835} & \tFa{fake}, \tFa{invalid}, \tFa{false}, \tFa{False} & \jk{riel}, \jk{lib}, \jk{trends}, \jk{HAL}\\
\texttt{L16.n1170} & \jk{ensures}, \jk{ensuring}, \jk{confirmed}, \jk{ingo} & \jk{STO}, \tFa{FALSE}, \tFa{false}, \jk{$\varnothing$}\\
\texttt{L16.n5090} & \jk{ugno}, \jk{iten}, \jk{Graf}, \jk{rolog} & \jk{pret}, \jk{pretend}, \tFa{fake}, \jk{nomin}\\
\texttt{L16.n9991} & \tFa{false}, \jk{fals}, \tFa{False}, \tFa{false} & \jk{$\varnothing$}, \jk{amon}, \jk{bert}, \jk{bis}\\
\texttt{L17.n11073} & \jk{integrity}, \jk{respons}, \tTr{genuine}, \jk{proper} & \jk{artificial}, \jk{stere}, \jk{unsafe}, \tFa{false}\\
\texttt{L17.n13852} & \tTr{true}, \tTr{real}, \tTr{True}, \tTr{TRUE} & \jk{asp}, \jk{ape}, \jk{dom}, \jk{Mend}\\
\texttt{L17.n8203} & \tFa{fake}, \jk{fals}, \jk{fraud}, \tFa{false} & \jk{aju}, \jk{Nu}, \jk{$\varnothing$}, \jk{pn}\\
\texttt{L18.n10672} & \jk{odi}, \jk{Roose}, \jk{berger}, \tTr{Verify} & \jk{mock}, \tFa{fake}, \jk{Mock}, \jk{mock}\\
\texttt{L18.n8574} & \jk{anymore}, \jk{excessive}, \jk{too}, \jk{unnecessary} & \jk{properly}, \jk{rather}, \jk{proper}, \jk{hopefully}\\
\texttt{L19.n11056} & \tTr{genuinely}, \tTr{truly}, \tTr{genuine}, \tTr{actual} & \jk{$\varnothing$}, \jk{indirect}, \jk{ester}, \jk{oll}\\
\texttt{L19.n8228}$^{\dagger}$ & \tTr{true}, \tTr{truth}, \tTr{True}, \tTr{TRUE} & \tFa{false}, \tFa{False}, \tFa{False}, \tFa{false}\\
\texttt{L20.n14286}$^{\dagger}$ & \jk{shake}, \jk{shook}, \tFa{False}, \jk{shaking} & \tTr{yes}, \tTr{Yes}, \tTr{true}, \tTr{Yes}\\
\texttt{L20.n4786} & \tTr{right}, \jk{achuset}, \tTr{TRUE}, \jk{Seth} & \tFa{wrong}, \jk{mistaken}, \jk{mis}, \tFa{incorrect}\\
\texttt{L22.h20.c93} & \jk{perfection}, \jk{exact}, \jk{exactly}, \jk{mainstream} & \jk{unexpected}, \jk{new}, \tFa{false}, \tFa{incorrect}\\
\texttt{L22.h20.svd0} & \jk{poorly}, \tFa{wrong}, \jk{poor}, \jk{incons} & \tTr{correctly}, \jk{properly}, \tTr{correct}, \jk{proper}\\
\texttt{L22.h20.svd40} & \jk{slightly}, \jk{slight}, \jk{missing}, \jk{remaining} & \jk{various}, \jk{increasingly}, \tTr{actual}, \tTr{real}\\
\texttt{L22.h23.svd2} & \jk{unknown}, \tFa{wrong}, \jk{absent}, \jk{isEmpty} & \tTr{correctly}, \tTr{valid}, \jk{viable}, \tTr{legitimate}\\
\texttt{L22.n1070} & \jk{reliable}, \jk{properly}, \jk{efficient}, \tTr{accurate} & \jk{unsafe}, \jk{poorly}, \jk{unfair}, \jk{undefined}\\
\texttt{L22.n5778} & \tFa{wrong}, \jk{wr}, \tFa{incorrect}, \jk{Wr} & \jk{properly}, \jk{proper}, \jk{gc}, \jk{olia}\\
\texttt{L22.n8735} & \jk{sek}, \jk{pred}, \jk{ork}, \jk{otal} & \jk{fals}, \tFa{false}, \tFa{incorrect}, \tFa{False}\\
\texttt{L24.n1021} & \jk{Strip}, \jk{Tib}, \jk{endo}, \jk{strip} & \tFa{false}, \tFa{fake}, \tFa{False}, \tFa{false}\\
\texttt{L24.n4907} & \jk{poorly}, \jk{poor}, \tFa{incorrect}, \tFa{wrong} & \tTr{correctly}, \jk{ben}, \jk{adequate}, \tTr{correct}\\
\texttt{L24.n5072} & \tTr{true}, \tTr{True}, \tTr{TRUE}, \tTr{true} & \jk{zero}, \jk{Zero}, \jk{Zero}, \jk{zero}\\
\texttt{L25.h8.svd45} & \jk{wild}, \tFa{false}, \jk{artificial}, \jk{amateur} & \tTr{real}, \jk{normal}, \jk{Private}, \jk{private}\\
\texttt{L27.n11713} & \tTr{truly}, \tTr{true}, \tTr{genuinely}, \tTr{true} & \jk{oft}, \cjkjp{宮}{palace}, \jk{uvud}, \jk{ço}\\
\texttt{L28.n4079} & \jk{inet}, \jk{stract}, \jk{èse}, \jk{teger} & \tTr{true}, \tTr{True}, \tTr{true}, \tTr{True}\\
\texttt{L28.n9060} & \tTr{truly}, \tTr{true}, \tTr{True}, \jk{Tru} & \tTr{accurate}, \jk{accur}, \jk{accurately}, \jk{accuracy}\\
\texttt{L29.n3831} & \tTr{real}, \tTr{real}, \tTr{Real}, \tTr{Real} & \jk{esser}, \jk{gorith}, \jk{otta}, \tFa{False}\\
\texttt{L29.n8705} & \tTr{true}, \tTr{True}, \tTr{true}, \tTr{True} & \jk{illet}, \jk{agr}, \jk{ícula}, \jk{ől}\\
\texttt{L30.h3.c8} & \jk{Complete}, \jk{complete}, \jk{Complete}, \jk{formal} & \jk{exact}, \jk{extract}, \jk{aver}, \jk{Extract}\\
\texttt{L30.n11434} & \tFa{false}, \tFa{fake}, \tFa{False}, \tFa{false} & \jk{F}, \jk{Bog}, \jk{chen}, \jk{icc}\\
\texttt{L30.n11755} & \tTr{real}, \tTr{real}, \tTr{actual}, \tTr{true} & \jk{ally}, \jk{ALL}, \jk{benfalls}, \jk{iously}\\
\texttt{L30.n498} & \tTr{genuine}, \tTr{authentic}, \tTr{accurate}, \tTr{true} & \jk{arbitrary}, \tFa{False}, \jk{idle}, \jk{lazy}\\
\texttt{L31.n13669} & \jk{xF}, \jk{stract}, \jk{icer}, \jk{kör} & \tTr{true}, \tTr{True}, \tTr{True}, \tTr{true}\\
\texttt{L31.n8773} & \tFa{false}, \tFa{False}, \tFa{False}, \jk{fals} & \jk{ulo}, \jk{heng}, \jk{Ä}, \jk{kennis}\\
\bottomrule\end{tabular}\end{table}

\end{CJK*}
\FloatBarrier
\section{{Regularity and Certification of the Second-Order Intervention}}
\label{app:procedure}

\subsection{The dose axis is smooth across the full ladder}
\label{app:dose-regularity}

Certification, the coupling fits of \S\ref{app:coupling-fits}, and every
{claim anchor at $k\le1$. The regime $k>1$ ($\lambda=1-2k$) is read} as extrapolation along a regular response surface rather than as an artefact of an off-manifold
$do(\cdot)$, and that reading rests on the surface {being} smooth
through the transition. Table~\ref{tab:margin_worlds} (\S\ref{app:renorm})
{gives the evidence. At $\times1$ every family still answers correctly on both
classes, every family-class pair} crosses zero between $\times1$ and $\times2$,
and no family shows a discontinuity there. Llama's per-dose steps on the true
side are near-constant ($-2.07,-2.27,-2.48,-2.70$), while the two joint-core
families decelerate sharply at depth {(Gemma $-2.85,-2.38,-1.09,-0.54$ and Qwen
$-5.26,-5.07,-3.56,-2.17$}). {Compensatory recruitment predicts a marginal effect that flattens as the intervention
deepens, because the reserve's contribution grows with dose (\S\ref{app:recruitment}) as the primary's marginal effect shrinks. Llama, whose reserve is by far the smallest of the three (\S\ref{app:pretraining}), is the family whose steps do not flatten.} This co-occurs with the true-side norm growth documented
in \S\ref{app:renorm} (Table~\ref{tab:gain}), consistent with
Eq.~{\ref{eq:normexpansion}}.

Magnitudes at $k>1$ are nonetheless {read as ordinal rather than calibrated, because the
per-direction} fits place the linear boundary at or near full inversion for
most directions (\S\ref{app:coupling-fits}), and \S\ref{app:renorm} locates the
dose at which the intervention itself leaves the linear regime of
Eq.~{\ref{eq:normexpansion}}.

\subsection{Certification statistics}
\label{app:certification-stats}

A direction is scored per side (true-statement, false-statement) with a paired
$t=|\mathrm{mean}|/\mathrm{sem}$ against {a threshold of $t\ge2$. Cells below}
threshold are treated as not reliably different from zero. \emph{Certified
contributor} (main text \S{\ref{sec:census}}) requires $E^0_r>0$, reliably, on both sides in
the clean model. \emph{Recruited} requires the same sign-and-reliability test
to hold under the primary intervention for a direction that was clean-null or
clean-anti. {\texttt{flip@} in Tables~\ref{tab:llama_regime_recruit} to \ref{tab:mistral_regime_recruit}}
records the first dose at which that happens. Evaluation throughout uses pairs on which the model is correct, zero-shot, on both the true and false members (\S\ref{app:data}), so that recruitment is measured against a background where the clean verdict
{holds}.

\FloatBarrier
\section{Recruitment Under the Dosed Swap}
\label{app:recruitment}

\subsection{Per-model recruitment tables}
\label{app:recruitment-tables}
{{Tables~\ref{tab:llama_regime_recruit} to \ref{tab:mistral_regime_recruit} give, per model,} each benched direction's clean and dosed contributions on both classes and the dose at which it flips.}

\begin{table*}[!htbp]\centering\footnotesize\setlength{\tabcolsep}{3pt}
\caption{\textbf{Backup recruitment across dose under ablation of the Llama-3-8B-Instruct verdict core} (\texttt{L18.n11065}, contrastive swap, \texttt{sec\_ref=regime}, $n{=}532$). Per side, the clean erased margin $E^0_T=D_{TF}(T)-D_{TF}(T,\mathrm{sec})$ and $E^0_F=D_{TF}(F,\mathrm{sec})-D_{TF}(F)$ {($+$ supports the verdict, with colour giving the clean class as \textcolor{FalseC}{anti}/\textcolor{TrueC}{contributor}/\textcolor{black!55}{null})}, then $E^{\pi}_T$ and $E^{\pi}_F$ under the ablation at each swap dose. \textcolor{TrueC}{Green}$=$reliably $>0$ (now a contributor on that side), \textcolor{FalseC}{red}$=${reliably $<0$. The superscript} is the paired $t=|\mathrm{mean}|/\mathrm{sem}$. Uncoloured cells are below threshold ($t<2$). {Row shade $=$ clean class, and `flip@'}$=$first dose a clean null/anti axis is reliably $>0$ on both sides. Across the coloured entries the median $t\approx8$ (up to $t\approx56$). The pair-averaged verdict gap runs $-7.8$ (clean) $\to -3.7, +1.4, +6.8, +12.3$ {across $\times1..\times4$, so the verdict gives way} between $\times1$ and $\times2$.}
\label{tab:llama_regime_recruit}
\resizebox{\textwidth}{!}{\begin{tabular}{lll rr rr rr rr rr c}
\toprule
 & & & \multicolumn{2}{c}{$E^0$ (clean)} & \multicolumn{2}{c}{$E^{\pi}\times1$} & \multicolumn{2}{c}{$E^{\pi}\times2$} & \multicolumn{2}{c}{$E^{\pi}\times3$} & \multicolumn{2}{c}{$E^{\pi}\times4$} & \\
\cmidrule(lr){4-5}\cmidrule(lr){6-7}\cmidrule(lr){8-9}\cmidrule(lr){10-11}\cmidrule(lr){12-13}
secondary & family & detects & $T$ & $F$ & $T$ & $F$ & $T$ & $F$ & $T$ & $F$ & $T$ & $F$ & flip@ \\
\midrule
\rowcolor{FalseC!8}
\texttt{L23.n8972} & mlp neuron & lie & \textcolor{FalseC}{$-0.04$} & \textcolor{FalseC}{$-0.06$} & \textcolor{TrueC}{$+0.02^{\scriptscriptstyle 13}$} & \textcolor{TrueC}{$+0.02^{\scriptscriptstyle 13}$} & \textcolor{TrueC}{$+0.06^{\scriptscriptstyle 34}$} & \textcolor{TrueC}{$+0.04^{\scriptscriptstyle 32}$} & \textcolor{TrueC}{$+0.11^{\scriptscriptstyle 47}$} & \textcolor{TrueC}{$+0.06^{\scriptscriptstyle 41}$} & \textcolor{TrueC}{$+0.16^{\scriptscriptstyle 56}$} & \textcolor{TrueC}{$+0.09^{\scriptscriptstyle 52}$} & $\times1$ \\
\rowcolor{FalseC!8}
\texttt{L23.n9811} & mlp neuron & truth & \textcolor{FalseC}{$-0.02$} & \textcolor{FalseC}{$-0.02$} & \textcolor{TrueC}{$+0.00^{\scriptscriptstyle 3}$} & \textcolor{TrueC}{$+0.00^{\scriptscriptstyle 4}$} & \textcolor{TrueC}{$+0.05^{\scriptscriptstyle 25}$} & \textcolor{TrueC}{$+0.03^{\scriptscriptstyle 22}$} & \textcolor{TrueC}{$+0.10^{\scriptscriptstyle 37}$} & \textcolor{TrueC}{$+0.05^{\scriptscriptstyle 34}$} & \textcolor{TrueC}{$+0.11^{\scriptscriptstyle 48}$} & \textcolor{TrueC}{$+0.02^{\scriptscriptstyle 8}$} & $\times1$ \\
\rowcolor{FalseC!8}
\texttt{L19.n3754} & mlp neuron & sure & \textcolor{FalseC}{$-0.01$} & \textcolor{FalseC}{$-0.02$} & \textcolor{TrueC}{$+0.00^{\scriptscriptstyle 3}$} & \textcolor{TrueC}{$+0.00^{\scriptscriptstyle 4}$} & \textcolor{TrueC}{$+0.01^{\scriptscriptstyle 11}$} & \textcolor{TrueC}{$+0.02^{\scriptscriptstyle 15}$} & \textcolor{TrueC}{$+0.02^{\scriptscriptstyle 18}$} & \textcolor{TrueC}{$+0.02^{\scriptscriptstyle 13}$} & \textcolor{TrueC}{$+0.02^{\scriptscriptstyle 14}$} & \textcolor{TrueC}{$+0.01^{\scriptscriptstyle 9}$} & $\times1$ \\
\rowcolor{FalseC!8}
\texttt{L21.n4049} & mlp neuron & wrong & \textcolor{FalseC}{$-0.01$} & \textcolor{FalseC}{$-0.01$} & \textcolor{TrueC}{$+0.01^{\scriptscriptstyle 8}$} & \textcolor{TrueC}{$+0.01^{\scriptscriptstyle 10}$} & \textcolor{TrueC}{$+0.03^{\scriptscriptstyle 21}$} & \textcolor{TrueC}{$+0.03^{\scriptscriptstyle 28}$} & \textcolor{TrueC}{$+0.04^{\scriptscriptstyle 29}$} & \textcolor{TrueC}{$+0.05^{\scriptscriptstyle 37}$} & \textcolor{TrueC}{$+0.03^{\scriptscriptstyle 19}$} & \textcolor{TrueC}{$+0.04^{\scriptscriptstyle 30}$} & $\times1$ \\
\rowcolor{FalseC!8}
\texttt{L19.n2738} & mlp neuron & illusion & \textcolor{FalseC}{$-0.01$} & \textcolor{FalseC}{$-0.01$} & $+0.00$ & \textcolor{TrueC}{$+0.00^{\scriptscriptstyle 2}$} & \textcolor{TrueC}{$+0.01^{\scriptscriptstyle 5}$} & \textcolor{TrueC}{$+0.02^{\scriptscriptstyle 13}$} & \textcolor{TrueC}{$+0.01^{\scriptscriptstyle 6}$} & \textcolor{TrueC}{$+0.04^{\scriptscriptstyle 30}$} & \textcolor{TrueC}{$+0.01^{\scriptscriptstyle 4}$} & \textcolor{TrueC}{$+0.06^{\scriptscriptstyle 42}$} & $\times2$ \\
\rowcolor{FalseC!8}
\texttt{L22.h1.svd117} & ov svd & true & \textcolor{FalseC}{$-0.00$} & \textcolor{FalseC}{$-0.01$} & \textcolor{TrueC}{$+0.01^{\scriptscriptstyle 10}$} & \textcolor{TrueC}{$+0.01^{\scriptscriptstyle 9}$} & \textcolor{TrueC}{$+0.02^{\scriptscriptstyle 16}$} & \textcolor{TrueC}{$+0.01^{\scriptscriptstyle 12}$} & \textcolor{TrueC}{$+0.03^{\scriptscriptstyle 21}$} & \textcolor{TrueC}{$+0.02^{\scriptscriptstyle 18}$} & \textcolor{TrueC}{$+0.03^{\scriptscriptstyle 18}$} & \textcolor{TrueC}{$+0.02^{\scriptscriptstyle 18}$} & $\times1$ \\
\rowcolor{black!6}
\texttt{L21.n798} & mlp neuron & true & \textcolor{black!55}{$-0.00$} & \textcolor{black!55}{$-0.00$} & \textcolor{TrueC}{$+0.00^{\scriptscriptstyle 2}$} & \textcolor{TrueC}{$+0.00^{\scriptscriptstyle 2}$} & $+0.00$ & \textcolor{TrueC}{$+0.00^{\scriptscriptstyle 2}$} & \textcolor{TrueC}{$+0.00^{\scriptscriptstyle 3}$} & \textcolor{TrueC}{$+0.00^{\scriptscriptstyle 2}$} & \textcolor{FalseC}{$-0.01^{\scriptscriptstyle 10}$} & \textcolor{FalseC}{$-0.01^{\scriptscriptstyle 8}$} & $\times1$ \\
\rowcolor{black!6}
\texttt{L22.svd323} & mlp svd & -- & \textcolor{black!55}{$-0.00$} & \textcolor{black!55}{$-0.00$} & $+0.00$ & \textcolor{TrueC}{$+0.00^{\scriptscriptstyle 2}$} & $+0.00$ & $+0.00$ & \textcolor{TrueC}{$+0.00^{\scriptscriptstyle 4}$} & $+0.00$ & \textcolor{TrueC}{$+0.00^{\scriptscriptstyle 2}$} & \textcolor{TrueC}{$+0.00^{\scriptscriptstyle 3}$} & $\times4$ \\
\rowcolor{black!6}
\texttt{L21.n517} & mlp neuron & yes & \textcolor{black!55}{$-0.00$} & \textcolor{black!55}{$-0.00$} & $+0.00$ & \textcolor{TrueC}{$+0.00^{\scriptscriptstyle 3}$} & \textcolor{TrueC}{$+0.00^{\scriptscriptstyle 4}$} & \textcolor{TrueC}{$+0.01^{\scriptscriptstyle 4}$} & \textcolor{TrueC}{$+0.01^{\scriptscriptstyle 8}$} & \textcolor{TrueC}{$+0.01^{\scriptscriptstyle 10}$} & \textcolor{TrueC}{$+0.01^{\scriptscriptstyle 4}$} & \textcolor{TrueC}{$+0.02^{\scriptscriptstyle 13}$} & $\times2$ \\
\rowcolor{black!6}
\texttt{L28.h27.c120} & ov neuron & True & \textcolor{black!55}{$-0.00$} & \textcolor{black!55}{$-0.00$} & $+0.00$ & \textcolor{TrueC}{$+0.00^{\scriptscriptstyle 3}$} & \textcolor{TrueC}{$+0.00^{\scriptscriptstyle 2}$} & $+0.00$ & \textcolor{TrueC}{$+0.01^{\scriptscriptstyle 12}$} & $+0.00$ & \textcolor{TrueC}{$+0.02^{\scriptscriptstyle 16}$} & $+0.00$ & -- \\
\rowcolor{black!6}
\texttt{L20.h11.c95} & ov neuron & false & \textcolor{black!55}{$+0.00$} & \textcolor{black!55}{$-0.00$} & $-0.00$ & $+0.00$ & $+0.00$ & $+0.00$ & \textcolor{TrueC}{$+0.00^{\scriptscriptstyle 3}$} & \textcolor{TrueC}{$+0.01^{\scriptscriptstyle 10}$} & $+0.00$ & \textcolor{TrueC}{$+0.03^{\scriptscriptstyle 19}$} & $\times3$ \\
\rowcolor{black!6}
\texttt{L21.n14165} & mlp neuron & Genuine & \textcolor{black!55}{$-0.00$} & \textcolor{black!55}{$-0.00$} & \textcolor{TrueC}{$+0.00^{\scriptscriptstyle 4}$} & \textcolor{TrueC}{$+0.01^{\scriptscriptstyle 6}$} & \textcolor{TrueC}{$+0.01^{\scriptscriptstyle 8}$} & \textcolor{TrueC}{$+0.01^{\scriptscriptstyle 7}$} & \textcolor{TrueC}{$+0.02^{\scriptscriptstyle 13}$} & \textcolor{TrueC}{$+0.01^{\scriptscriptstyle 6}$} & \textcolor{TrueC}{$+0.01^{\scriptscriptstyle 11}$} & \textcolor{TrueC}{$+0.00^{\scriptscriptstyle 2}$} & $\times1$ \\
\rowcolor{black!6}
\texttt{L28.h27.c34} & ov neuron & True & \textcolor{black!55}{$+0.00$} & \textcolor{black!55}{$+0.00$} & \textcolor{TrueC}{$+0.00^{\scriptscriptstyle 2}$} & \textcolor{TrueC}{$+0.00^{\scriptscriptstyle 3}$} & \textcolor{TrueC}{$+0.00^{\scriptscriptstyle 5}$} & $+0.00$ & \textcolor{TrueC}{$+0.01^{\scriptscriptstyle 10}$} & \textcolor{TrueC}{$+0.00^{\scriptscriptstyle 5}$} & \textcolor{TrueC}{$+0.02^{\scriptscriptstyle 15}$} & \textcolor{TrueC}{$+0.01^{\scriptscriptstyle 5}$} & $\times1$ \\
\rowcolor{black!6}
\texttt{L22.h1.svd116} & ov svd & correct & \textcolor{black!55}{$-0.00$} & \textcolor{black!55}{$+0.00$} & \textcolor{TrueC}{$+0.00^{\scriptscriptstyle 3}$} & \textcolor{TrueC}{$+0.00^{\scriptscriptstyle 4}$} & \textcolor{TrueC}{$+0.01^{\scriptscriptstyle 6}$} & \textcolor{TrueC}{$+0.01^{\scriptscriptstyle 6}$} & \textcolor{TrueC}{$+0.01^{\scriptscriptstyle 7}$} & \textcolor{TrueC}{$+0.01^{\scriptscriptstyle 8}$} & \textcolor{TrueC}{$+0.01^{\scriptscriptstyle 5}$} & \textcolor{TrueC}{$+0.01^{\scriptscriptstyle 8}$} & $\times1$ \\
\rowcolor{black!6}
\texttt{L28.h27.c0} & ov neuron & real & \textcolor{black!55}{$+0.00$} & \textcolor{black!55}{$-0.00$} & $+0.00$ & $+0.00$ & $+0.00$ & $-0.00$ & \textcolor{TrueC}{$+0.01^{\scriptscriptstyle 5}$} & \textcolor{TrueC}{$+0.00^{\scriptscriptstyle 4}$} & \textcolor{TrueC}{$+0.00^{\scriptscriptstyle 3}$} & \textcolor{TrueC}{$+0.01^{\scriptscriptstyle 9}$} & $\times3$ \\
\rowcolor{black!6}
\texttt{L22.h1.c60} & ov neuron & valid & \textcolor{black!55}{$+0.00$} & \textcolor{black!55}{$-0.00$} & \textcolor{TrueC}{$+0.01^{\scriptscriptstyle 5}$} & \textcolor{TrueC}{$+0.00^{\scriptscriptstyle 3}$} & \textcolor{TrueC}{$+0.01^{\scriptscriptstyle 6}$} & \textcolor{TrueC}{$+0.00^{\scriptscriptstyle 3}$} & \textcolor{TrueC}{$+0.01^{\scriptscriptstyle 9}$} & \textcolor{TrueC}{$+0.01^{\scriptscriptstyle 8}$} & \textcolor{TrueC}{$+0.01^{\scriptscriptstyle 8}$} & \textcolor{TrueC}{$+0.01^{\scriptscriptstyle 9}$} & $\times1$ \\
\rowcolor{black!6}
\texttt{L28.h27.c43} & ov neuron & -- & \textcolor{black!55}{$+0.00$} & \textcolor{black!55}{$-0.00$} & $+0.00$ & \textcolor{TrueC}{$+0.00^{\scriptscriptstyle 3}$} & \textcolor{TrueC}{$+0.00^{\scriptscriptstyle 3}$} & \textcolor{TrueC}{$+0.00^{\scriptscriptstyle 2}$} & \textcolor{TrueC}{$+0.01^{\scriptscriptstyle 11}$} & \textcolor{TrueC}{$+0.00^{\scriptscriptstyle 3}$} & \textcolor{TrueC}{$+0.02^{\scriptscriptstyle 17}$} & \textcolor{TrueC}{$+0.01^{\scriptscriptstyle 5}$} & $\times2$ \\
\rowcolor{orange!8}
\texttt{L24.n4543} & mlp neuron & yes & \textcolor{orange!85!black}{$-0.00$} & \textcolor{orange!85!black}{$-0.01$} & $+0.00$ & \textcolor{TrueC}{$+0.01^{\scriptscriptstyle 8}$} & \textcolor{TrueC}{$+0.03^{\scriptscriptstyle 18}$} & \textcolor{TrueC}{$+0.02^{\scriptscriptstyle 13}$} & \textcolor{TrueC}{$+0.06^{\scriptscriptstyle 29}$} & \textcolor{FalseC}{$-0.01^{\scriptscriptstyle 9}$} & \textcolor{TrueC}{$+0.08^{\scriptscriptstyle 36}$} & \textcolor{FalseC}{$-0.05^{\scriptscriptstyle 23}$} & -- \\
\rowcolor{orange!8}
\texttt{L20.n14044} & mlp neuron & truly & \textcolor{orange!85!black}{$-0.00$} & \textcolor{orange!85!black}{$-0.00$} & $+0.00$ & \textcolor{TrueC}{$+0.00^{\scriptscriptstyle 4}$} & \textcolor{TrueC}{$+0.00^{\scriptscriptstyle 3}$} & \textcolor{TrueC}{$+0.00^{\scriptscriptstyle 4}$} & \textcolor{TrueC}{$+0.01^{\scriptscriptstyle 10}$} & \textcolor{TrueC}{$+0.01^{\scriptscriptstyle 6}$} & \textcolor{TrueC}{$+0.02^{\scriptscriptstyle 12}$} & \textcolor{TrueC}{$+0.01^{\scriptscriptstyle 5}$} & -- \\
\rowcolor{orange!8}
\texttt{L20.n1390} & mlp neuron & wrong & \textcolor{orange!85!black}{$+0.00$} & \textcolor{orange!85!black}{$+0.00$} & \textcolor{TrueC}{$+0.01^{\scriptscriptstyle 7}$} & \textcolor{TrueC}{$+0.01^{\scriptscriptstyle 11}$} & \textcolor{TrueC}{$+0.01^{\scriptscriptstyle 7}$} & \textcolor{TrueC}{$+0.02^{\scriptscriptstyle 11}$} & \textcolor{TrueC}{$+0.02^{\scriptscriptstyle 14}$} & \textcolor{TrueC}{$+0.03^{\scriptscriptstyle 19}$} & \textcolor{TrueC}{$+0.04^{\scriptscriptstyle 22}$} & \textcolor{TrueC}{$+0.05^{\scriptscriptstyle 27}$} & -- \\
\bottomrule\end{tabular}}\end{table*}

\begin{table*}[!htbp]\centering\footnotesize\setlength{\tabcolsep}{3pt}
\caption{\textbf{Backup recruitment across dose under JOINT ablation of both Qwen2.5-7B-Instruct verdict cores} (\texttt{L22.n13149}$+$\texttt{L24.n14758}, contrastive swap, \texttt{sec\_ref=regime}, $n{=}478$). Columns as in Table~\ref{tab:llama_regime_recruit}. Across the coloured entries the median $t\approx21$ (up to $t\approx201$). {$\kappa_{\mathrm{base}}=-18.8$, and the joint $\kappa_{\mathrm{primary}}$ at $\times1..\times4=-8.5,+4.7,+15.8,+23.4$ crosses $0$ at $\times2$, where the verdict flips.} Null-class rows stay $\sim\!0$ across all doses (specificity).}
\label{tab:qwen_regime_recruit}
\resizebox{\textwidth}{!}{\begin{tabular}{lll rr rr rr rr rr c}
\toprule
 & & & \multicolumn{2}{c}{$E^0$ (clean)} & \multicolumn{2}{c}{$E^{\pi}\times1$} & \multicolumn{2}{c}{$E^{\pi}\times2$} & \multicolumn{2}{c}{$E^{\pi}\times3$} & \multicolumn{2}{c}{$E^{\pi}\times4$} & \\
\cmidrule(lr){4-5}\cmidrule(lr){6-7}\cmidrule(lr){8-9}\cmidrule(lr){10-11}\cmidrule(lr){12-13}
secondary & family & detects & $T$ & $F$ & $T$ & $F$ & $T$ & $F$ & $T$ & $F$ & $T$ & $F$ & flip@ \\
\midrule
\rowcolor{FalseC!8}
\texttt{L27.n13033} & mlp neuron & false & \textcolor{FalseC}{$-2.86$} & \textcolor{FalseC}{$-2.88$} & \textcolor{FalseC}{$-0.50^{\scriptscriptstyle 15}$} & \textcolor{FalseC}{$-0.51^{\scriptscriptstyle 15}$} & \textcolor{TrueC}{$+2.93^{\scriptscriptstyle 116}$} & \textcolor{TrueC}{$+3.22^{\scriptscriptstyle 109}$} & \textcolor{TrueC}{$+3.95^{\scriptscriptstyle 189}$} & \textcolor{TrueC}{$+4.45^{\scriptscriptstyle 178}$} & \textcolor{TrueC}{$+4.08^{\scriptscriptstyle 199}$} & \textcolor{TrueC}{$+4.65^{\scriptscriptstyle 201}$} & $\times2$ \\
\rowcolor{FalseC!8}
\texttt{L27.n4228} & mlp neuron & truly & \textcolor{FalseC}{$-0.84$} & \textcolor{FalseC}{$-0.87$} & \textcolor{TrueC}{$+0.03^{\scriptscriptstyle 4}$} & \textcolor{TrueC}{$+0.03^{\scriptscriptstyle 4}$} & \textcolor{TrueC}{$+0.37^{\scriptscriptstyle 29}$} & \textcolor{TrueC}{$+0.41^{\scriptscriptstyle 29}$} & \textcolor{TrueC}{$+1.43^{\scriptscriptstyle 47}$} & \textcolor{TrueC}{$+1.58^{\scriptscriptstyle 47}$} & \textcolor{TrueC}{$+2.62^{\scriptscriptstyle 63}$} & \textcolor{TrueC}{$+2.96^{\scriptscriptstyle 63}$} & $\times1$ \\
\rowcolor{FalseC!8}
\texttt{L25.n4929} & mlp neuron & false & \textcolor{FalseC}{$-0.25$} & \textcolor{FalseC}{$-0.07$} & \textcolor{FalseC}{$-0.01^{\scriptscriptstyle 3}$} & \textcolor{FalseC}{$-0.00^{\scriptscriptstyle 2}$} & \textcolor{TrueC}{$+0.02^{\scriptscriptstyle 12}$} & \textcolor{TrueC}{$+0.05^{\scriptscriptstyle 17}$} & \textcolor{TrueC}{$+0.05^{\scriptscriptstyle 21}$} & \textcolor{TrueC}{$+0.21^{\scriptscriptstyle 35}$} & \textcolor{TrueC}{$+0.05^{\scriptscriptstyle 21}$} & \textcolor{TrueC}{$+0.41^{\scriptscriptstyle 45}$} & $\times2$ \\
\rowcolor{FalseC!8}
\texttt{L23.n8341} & mlp neuron & illusion & \textcolor{FalseC}{$-0.18$} & \textcolor{FalseC}{$-0.11$} & \textcolor{TrueC}{$+0.00^{\scriptscriptstyle 2}$} & $+0.00$ & \textcolor{TrueC}{$+0.04^{\scriptscriptstyle 18}$} & \textcolor{TrueC}{$+0.15^{\scriptscriptstyle 29}$} & \textcolor{TrueC}{$+0.08^{\scriptscriptstyle 28}$} & \textcolor{TrueC}{$+0.49^{\scriptscriptstyle 52}$} & \textcolor{TrueC}{$+0.12^{\scriptscriptstyle 37}$} & \textcolor{TrueC}{$+0.84^{\scriptscriptstyle 66}$} & $\times2$ \\
\rowcolor{FalseC!8}
\texttt{L24.n12848} & mlp neuron & false & \textcolor{FalseC}{$-0.21$} & \textcolor{FalseC}{$-0.04$} & $+0.01$ & $+0.00$ & \textcolor{TrueC}{$+0.11^{\scriptscriptstyle 25}$} & \textcolor{TrueC}{$+0.29^{\scriptscriptstyle 33}$} & \textcolor{TrueC}{$+0.04^{\scriptscriptstyle 15}$} & \textcolor{TrueC}{$+0.32^{\scriptscriptstyle 41}$} & \textcolor{FalseC}{$-0.01^{\scriptscriptstyle 3}$} & \textcolor{TrueC}{$+0.26^{\scriptscriptstyle 35}$} & $\times2$ \\
\rowcolor{FalseC!8}
\texttt{L23.n14476} & mlp neuron & real & \textcolor{FalseC}{$-0.05$} & \textcolor{FalseC}{$-0.09$} & \textcolor{FalseC}{$-0.00^{\scriptscriptstyle 3}$} & \textcolor{FalseC}{$-0.01^{\scriptscriptstyle 5}$} & \textcolor{TrueC}{$+0.02^{\scriptscriptstyle 10}$} & \textcolor{TrueC}{$+0.04^{\scriptscriptstyle 14}$} & \textcolor{TrueC}{$+0.13^{\scriptscriptstyle 34}$} & \textcolor{TrueC}{$+0.16^{\scriptscriptstyle 38}$} & \textcolor{TrueC}{$+0.29^{\scriptscriptstyle 44}$} & \textcolor{TrueC}{$+0.13^{\scriptscriptstyle 28}$} & $\times2$ \\
\rowcolor{FalseC!8}
\texttt{L25.n8444} & mlp neuron & fake & \textcolor{FalseC}{$-0.06$} & \textcolor{FalseC}{$-0.03$} & \textcolor{TrueC}{$+0.01^{\scriptscriptstyle 5}$} & \textcolor{TrueC}{$+0.02^{\scriptscriptstyle 7}$} & \textcolor{TrueC}{$+0.07^{\scriptscriptstyle 13}$} & \textcolor{TrueC}{$+0.14^{\scriptscriptstyle 14}$} & \textcolor{TrueC}{$+0.21^{\scriptscriptstyle 49}$} & \textcolor{TrueC}{$+0.64^{\scriptscriptstyle 44}$} & \textcolor{TrueC}{$+0.26^{\scriptscriptstyle 71}$} & \textcolor{TrueC}{$+1.09^{\scriptscriptstyle 70}$} & $\times1$ \\
\rowcolor{FalseC!8}
\texttt{L22.n13025} & mlp neuron & {\cjkjp{真正}{genuine}} & \textcolor{FalseC}{$-0.02$} & \textcolor{FalseC}{$-0.04$} & \textcolor{FalseC}{$-0.03^{\scriptscriptstyle 17}$} & \textcolor{FalseC}{$-0.06^{\scriptscriptstyle 29}$} & \textcolor{FalseC}{$-0.03^{\scriptscriptstyle 16}$} & \textcolor{FalseC}{$-0.04^{\scriptscriptstyle 22}$} & \textcolor{FalseC}{$-0.02^{\scriptscriptstyle 10}$} & \textcolor{FalseC}{$-0.01^{\scriptscriptstyle 6}$} & \textcolor{FalseC}{$-0.02^{\scriptscriptstyle 11}$} & $-0.00$ & -- \\
\rowcolor{FalseC!8}
\texttt{L22.n4538} & mlp neuron & false & \textcolor{FalseC}{$-0.03$} & \textcolor{FalseC}{$-0.03$} & \textcolor{FalseC}{$-0.03^{\scriptscriptstyle 18}$} & \textcolor{FalseC}{$-0.04^{\scriptscriptstyle 21}$} & \textcolor{FalseC}{$-0.02^{\scriptscriptstyle 13}$} & \textcolor{FalseC}{$-0.07^{\scriptscriptstyle 27}$} & \textcolor{FalseC}{$-0.01^{\scriptscriptstyle 7}$} & \textcolor{FalseC}{$-0.06^{\scriptscriptstyle 26}$} & \textcolor{FalseC}{$-0.01^{\scriptscriptstyle 4}$} & \textcolor{FalseC}{$-0.05^{\scriptscriptstyle 20}$} & -- \\
\rowcolor{FalseC!8}
\texttt{L26.h20.svd17} & ov svd & -- & \textcolor{FalseC}{$-0.01$} & \textcolor{FalseC}{$-0.01$} & \textcolor{FalseC}{$-0.00^{\scriptscriptstyle 4}$} & \textcolor{FalseC}{$-0.01^{\scriptscriptstyle 5}$} & $+0.00$ & $+0.00$ & \textcolor{TrueC}{$+0.00^{\scriptscriptstyle 2}$} & \textcolor{TrueC}{$+0.01^{\scriptscriptstyle 4}$} & $+0.00$ & $+0.00$ & $\times3$ \\
\rowcolor{black!6}
\texttt{L11.svd690} & mlp svd & true & \textcolor{black!55}{$-0.00$} & \textcolor{black!55}{$-0.00$} & $-0.00$ & $-0.00$ & \textcolor{FalseC}{$-0.01^{\scriptscriptstyle 2}$} & $-0.00$ & $-0.00$ & $+0.00$ & $-0.00$ & $+0.00$ & -- \\
\rowcolor{black!6}
\texttt{L24.n5510} & mlp neuron & none & \textcolor{black!55}{$-0.00$} & \textcolor{black!55}{$+0.00$} & $-0.00$ & $-0.00$ & \textcolor{FalseC}{$-0.01^{\scriptscriptstyle 3}$} & $-0.00$ & $-0.00$ & \textcolor{TrueC}{$+0.00^{\scriptscriptstyle 3}$} & $-0.00$ & $+0.00$ & -- \\
\rowcolor{orange!8}
\texttt{L21.svd2451} & mlp svd & -- & \textcolor{orange!85!black}{$+0.00$} & \textcolor{orange!85!black}{$+0.00$} & \textcolor{TrueC}{$+0.00^{\scriptscriptstyle 3}$} & $-0.00$ & \textcolor{FalseC}{$-0.00^{\scriptscriptstyle 2}$} & \textcolor{TrueC}{$+0.00^{\scriptscriptstyle 2}$} & \textcolor{FalseC}{$-0.00^{\scriptscriptstyle 2}$} & $+0.00$ & $+0.00$ & $-0.00$ & -- \\
\rowcolor{orange!8}
\texttt{L27.n6739} & mlp neuron & none & \textcolor{orange!85!black}{$+0.00$} & \textcolor{orange!85!black}{$+0.00$} & \textcolor{TrueC}{$+0.00^{\scriptscriptstyle 3}$} & \textcolor{TrueC}{$+0.00^{\scriptscriptstyle 3}$} & \textcolor{TrueC}{$+0.00^{\scriptscriptstyle 3}$} & \textcolor{TrueC}{$+0.00^{\scriptscriptstyle 4}$} & \textcolor{TrueC}{$+0.00^{\scriptscriptstyle 3}$} & \textcolor{TrueC}{$+0.00^{\scriptscriptstyle 5}$} & \textcolor{TrueC}{$+0.00^{\scriptscriptstyle 2}$} & \textcolor{TrueC}{$+0.00^{\scriptscriptstyle 3}$} & -- \\
\rowcolor{TrueC!8}
\texttt{L24.n8308} & mlp neuron & yes & \textcolor{TrueC}{$+0.16$} & \textcolor{TrueC}{$+0.49$} & \textcolor{TrueC}{$+0.06^{\scriptscriptstyle 27}$} & \textcolor{TrueC}{$+0.10^{\scriptscriptstyle 32}$} & \textcolor{FalseC}{$-0.25^{\scriptscriptstyle 44}$} & \textcolor{FalseC}{$-0.17^{\scriptscriptstyle 52}$} & \textcolor{FalseC}{$-0.46^{\scriptscriptstyle 68}$} & \textcolor{FalseC}{$-0.06^{\scriptscriptstyle 15}$} & \textcolor{FalseC}{$-0.58^{\scriptscriptstyle 85}$} & \textcolor{TrueC}{$+0.16^{\scriptscriptstyle 22}$} & -- \\
\rowcolor{TrueC!8}
\texttt{L22.n9609} & mlp neuron & False & \textcolor{TrueC}{$+0.29$} & \textcolor{TrueC}{$+0.28$} & \textcolor{TrueC}{$+0.39^{\scriptscriptstyle 46}$} & \textcolor{TrueC}{$+0.41^{\scriptscriptstyle 45}$} & \textcolor{TrueC}{$+0.40^{\scriptscriptstyle 43}$} & \textcolor{TrueC}{$+0.42^{\scriptscriptstyle 50}$} & \textcolor{TrueC}{$+0.24^{\scriptscriptstyle 44}$} & \textcolor{TrueC}{$+0.28^{\scriptscriptstyle 53}$} & \textcolor{TrueC}{$+0.17^{\scriptscriptstyle 42}$} & \textcolor{TrueC}{$+0.19^{\scriptscriptstyle 45}$} & -- \\
\rowcolor{TrueC!8}
\texttt{L27.n15192} & mlp neuron & actual & \textcolor{TrueC}{$+0.19$} & \textcolor{TrueC}{$+0.20$} & \textcolor{TrueC}{$+0.09^{\scriptscriptstyle 52}$} & \textcolor{TrueC}{$+0.10^{\scriptscriptstyle 55}$} & \textcolor{FalseC}{$-0.09^{\scriptscriptstyle 41}$} & \textcolor{FalseC}{$-0.10^{\scriptscriptstyle 41}$} & \textcolor{FalseC}{$-0.22^{\scriptscriptstyle 78}$} & \textcolor{FalseC}{$-0.24^{\scriptscriptstyle 77}$} & \textcolor{FalseC}{$-0.29^{\scriptscriptstyle 86}$} & \textcolor{FalseC}{$-0.33^{\scriptscriptstyle 85}$} & -- \\
\rowcolor{TrueC!8}
\texttt{L25.n11824} & mlp neuron & {\cjkjp{真正}{genuine}} & \textcolor{TrueC}{$+0.10$} & \textcolor{TrueC}{$+0.12$} & \textcolor{TrueC}{$+0.01^{\scriptscriptstyle 10}$} & \textcolor{TrueC}{$+0.04^{\scriptscriptstyle 15}$} & \textcolor{FalseC}{$-0.04^{\scriptscriptstyle 20}$} & \textcolor{FalseC}{$-0.11^{\scriptscriptstyle 52}$} & \textcolor{FalseC}{$-0.07^{\scriptscriptstyle 32}$} & \textcolor{FalseC}{$-0.09^{\scriptscriptstyle 43}$} & \textcolor{FalseC}{$-0.08^{\scriptscriptstyle 33}$} & \textcolor{FalseC}{$-0.03^{\scriptscriptstyle 11}$} & -- \\
\rowcolor{TrueC!8}
\texttt{L26.h17.svd117} & ov svd & {\cjkjp{真}{true}} & \textcolor{TrueC}{$+0.11$} & \textcolor{TrueC}{$+0.11$} & \textcolor{TrueC}{$+0.07^{\scriptscriptstyle 38}$} & \textcolor{TrueC}{$+0.13^{\scriptscriptstyle 57}$} & \textcolor{TrueC}{$+0.04^{\scriptscriptstyle 27}$} & \textcolor{TrueC}{$+0.13^{\scriptscriptstyle 61}$} & \textcolor{TrueC}{$+0.04^{\scriptscriptstyle 24}$} & \textcolor{TrueC}{$+0.08^{\scriptscriptstyle 52}$} & \textcolor{TrueC}{$+0.05^{\scriptscriptstyle 27}$} & \textcolor{TrueC}{$+0.03^{\scriptscriptstyle 18}$} & -- \\
\rowcolor{TrueC!8}
\texttt{L20.h1.svd6} & ov svd & YES & \textcolor{TrueC}{$+0.06$} & \textcolor{TrueC}{$+0.14$} & \textcolor{TrueC}{$+0.03^{\scriptscriptstyle 7}$} & \textcolor{TrueC}{$+0.17^{\scriptscriptstyle 48}$} & \textcolor{TrueC}{$+0.01^{\scriptscriptstyle 3}$} & \textcolor{TrueC}{$+0.10^{\scriptscriptstyle 33}$} & \textcolor{TrueC}{$+0.01^{\scriptscriptstyle 4}$} & \textcolor{TrueC}{$+0.03^{\scriptscriptstyle 8}$} & \textcolor{TrueC}{$+0.01^{\scriptscriptstyle 4}$} & $-0.01$ & -- \\
\rowcolor{TrueC!8}
\texttt{L23.h27.c82} & ov neuron & correct & \textcolor{TrueC}{$+0.01$} & \textcolor{TrueC}{$+0.02$} & \textcolor{TrueC}{$+0.02^{\scriptscriptstyle 12}$} & \textcolor{TrueC}{$+0.02^{\scriptscriptstyle 12}$} & \textcolor{TrueC}{$+0.03^{\scriptscriptstyle 14}$} & \textcolor{TrueC}{$+0.01^{\scriptscriptstyle 5}$} & \textcolor{TrueC}{$+0.02^{\scriptscriptstyle 11}$} & $+0.00$ & \textcolor{TrueC}{$+0.02^{\scriptscriptstyle 11}$} & $-0.00$ & -- \\
\bottomrule\end{tabular}}\end{table*}

\begin{table*}[!htbp]\centering\footnotesize\setlength{\tabcolsep}{3pt}
\caption{\textbf{Backup recruitment across dose under JOINT ablation of both Gemma-2-9b-it verdict cores} (\texttt{L28.n2046}$+$\texttt{L33.n4294}, contrastive swap, \texttt{sec\_ref=regime}, $n{=}583$). Columns as in Table~\ref{tab:llama_regime_recruit}. Across the coloured entries the median $t\approx20$ (up to $t\approx252$). {$\kappa_{\mathrm{base}}=-9.50$, and joint} $\kappa_{\mathrm{primary}}$ at $\times1..\times4=-4.38,+2.30,+5.59,+7.14$ (crosses $0$ at $\times2$). {The one null-class row, \texttt{L36.n1119}, turns reliably negative from $\times2$ ($E^{\pi}$ between $-0.10$ and $-0.22$) and never supports the verdict.}}
\label{tab:gemma_regime_recruit}
\resizebox{\textwidth}{!}{\begin{tabular}{lll rr rr rr rr rr c}
\toprule
 & & & \multicolumn{2}{c}{$E^0$ (clean)} & \multicolumn{2}{c}{$E^{\pi}\times1$} & \multicolumn{2}{c}{$E^{\pi}\times2$} & \multicolumn{2}{c}{$E^{\pi}\times3$} & \multicolumn{2}{c}{$E^{\pi}\times4$} & \\
\cmidrule(lr){4-5}\cmidrule(lr){6-7}\cmidrule(lr){8-9}\cmidrule(lr){10-11}\cmidrule(lr){12-13}
secondary & family & detects & $T$ & $F$ & $T$ & $F$ & $T$ & $F$ & $T$ & $F$ & $T$ & $F$ & flip@ \\
\midrule
\rowcolor{FalseC!8}
\texttt{L38.n7088} & mlp neuron & wrong & \textcolor{FalseC}{$-1.420$} & \textcolor{FalseC}{$-0.715$} & \textcolor{TrueC}{$+0.474^{\scriptscriptstyle 72}$} & \textcolor{TrueC}{$+0.613^{\scriptscriptstyle 54}$} & \textcolor{TrueC}{$+1.060^{\scriptscriptstyle 194}$} & \textcolor{TrueC}{$+2.202^{\scriptscriptstyle 122}$} & \textcolor{TrueC}{$+1.178^{\scriptscriptstyle 228}$} & \textcolor{TrueC}{$+2.587^{\scriptscriptstyle 163}$} & \textcolor{TrueC}{$+1.215^{\scriptscriptstyle 252}$} & \textcolor{TrueC}{$+2.727^{\scriptscriptstyle 197}$} & $\times1$ \\
\rowcolor{FalseC!8}
\texttt{L39.n10085} & mlp neuron & fake & \textcolor{FalseC}{$-0.504$} & \textcolor{FalseC}{$-0.418$} & \textcolor{FalseC}{$-0.364^{\scriptscriptstyle 88}$} & \textcolor{FalseC}{$-0.342^{\scriptscriptstyle 94}$} & \textcolor{FalseC}{$-0.179^{\scriptscriptstyle 44}$} & \textcolor{FalseC}{$-0.181^{\scriptscriptstyle 43}$} & \textcolor{FalseC}{$-0.017^{\scriptscriptstyle 4}$} & \textcolor{FalseC}{$-0.022^{\scriptscriptstyle 4}$} & \textcolor{TrueC}{$+0.071^{\scriptscriptstyle 15}$} & \textcolor{TrueC}{$+0.077^{\scriptscriptstyle 15}$} & $\times4$ \\
\rowcolor{FalseC!8}
\texttt{L38.n854} & mlp neuron & true & \textcolor{FalseC}{$-0.179$} & \textcolor{FalseC}{$-0.281$} & \textcolor{FalseC}{$-0.044^{\scriptscriptstyle 26}$} & \textcolor{FalseC}{$-0.055^{\scriptscriptstyle 26}$} & \textcolor{TrueC}{$+0.075^{\scriptscriptstyle 35}$} & \textcolor{TrueC}{$+0.072^{\scriptscriptstyle 40}$} & \textcolor{TrueC}{$+0.177^{\scriptscriptstyle 55}$} & \textcolor{TrueC}{$+0.146^{\scriptscriptstyle 66}$} & \textcolor{TrueC}{$+0.247^{\scriptscriptstyle 69}$} & \textcolor{TrueC}{$+0.192^{\scriptscriptstyle 78}$} & $\times2$ \\
\rowcolor{FalseC!8}
\texttt{L41.n8771} & mlp neuron & True & \textcolor{FalseC}{$-0.107$} & \textcolor{FalseC}{$-0.202$} & \textcolor{TrueC}{$+0.047^{\scriptscriptstyle 16}$} & \textcolor{TrueC}{$+0.057^{\scriptscriptstyle 16}$} & \textcolor{TrueC}{$+0.261^{\scriptscriptstyle 69}$} & \textcolor{TrueC}{$+0.221^{\scriptscriptstyle 66}$} & \textcolor{TrueC}{$+0.339^{\scriptscriptstyle 80}$} & \textcolor{TrueC}{$+0.236^{\scriptscriptstyle 80}$} & \textcolor{TrueC}{$+0.354^{\scriptscriptstyle 90}$} & \textcolor{TrueC}{$+0.232^{\scriptscriptstyle 86}$} & $\times1$ \\
\rowcolor{FalseC!8}
\texttt{L39.n10210} & mlp neuron & incorrect & \textcolor{FalseC}{$-0.167$} & \textcolor{FalseC}{$-0.127$} & \textcolor{FalseC}{$-0.019^{\scriptscriptstyle 13}$} & \textcolor{FalseC}{$-0.018^{\scriptscriptstyle 13}$} & \textcolor{TrueC}{$+0.046^{\scriptscriptstyle 30}$} & \textcolor{TrueC}{$+0.048^{\scriptscriptstyle 31}$} & \textcolor{TrueC}{$+0.019^{\scriptscriptstyle 15}$} & \textcolor{TrueC}{$+0.021^{\scriptscriptstyle 17}$} & \textcolor{TrueC}{$+0.007^{\scriptscriptstyle 5}$} & \textcolor{TrueC}{$+0.006^{\scriptscriptstyle 5}$} & $\times2$ \\
\rowcolor{FalseC!8}
\texttt{L41.n3789} & mlp neuron & real & \textcolor{FalseC}{$-0.100$} & \textcolor{FalseC}{$-0.182$} & \textcolor{FalseC}{$-0.051^{\scriptscriptstyle 37}$} & \textcolor{FalseC}{$-0.063^{\scriptscriptstyle 40}$} & \textcolor{TrueC}{$+0.064^{\scriptscriptstyle 40}$} & \textcolor{TrueC}{$+0.054^{\scriptscriptstyle 38}$} & \textcolor{TrueC}{$+0.141^{\scriptscriptstyle 69}$} & \textcolor{TrueC}{$+0.099^{\scriptscriptstyle 62}$} & \textcolor{TrueC}{$+0.177^{\scriptscriptstyle 88}$} & \textcolor{TrueC}{$+0.117^{\scriptscriptstyle 74}$} & $\times2$ \\
\rowcolor{FalseC!8}
\texttt{L32.n8200} & mlp neuron & False & \textcolor{FalseC}{$-0.067$} & \textcolor{FalseC}{$-0.175$} & \textcolor{TrueC}{$+0.028^{\scriptscriptstyle 12}$} & \textcolor{FalseC}{$-0.082^{\scriptscriptstyle 36}$} & $-0.002$ & \textcolor{FalseC}{$-0.003^{\scriptscriptstyle 2}$} & \textcolor{TrueC}{$+0.032^{\scriptscriptstyle 17}$} & \textcolor{FalseC}{$-0.048^{\scriptscriptstyle 22}$} & \textcolor{TrueC}{$+0.055^{\scriptscriptstyle 26}$} & \textcolor{FalseC}{$-0.080^{\scriptscriptstyle 32}$} & -- \\
\rowcolor{FalseC!8}
\texttt{L33.n1466} & mlp neuron & deception & \textcolor{FalseC}{$-0.147$} & \textcolor{FalseC}{$-0.039$} & \textcolor{FalseC}{$-0.035^{\scriptscriptstyle 22}$} & \textcolor{FalseC}{$-0.032^{\scriptscriptstyle 21}$} & \textcolor{TrueC}{$+0.011^{\scriptscriptstyle 9}$} & \textcolor{TrueC}{$+0.029^{\scriptscriptstyle 19}$} & \textcolor{TrueC}{$+0.016^{\scriptscriptstyle 13}$} & \textcolor{TrueC}{$+0.038^{\scriptscriptstyle 30}$} & \textcolor{TrueC}{$+0.013^{\scriptscriptstyle 12}$} & \textcolor{TrueC}{$+0.037^{\scriptscriptstyle 29}$} & $\times2$ \\
\rowcolor{FalseC!8}
\texttt{L34.n4784} & mlp neuron & truth & \textcolor{FalseC}{$-0.044$} & \textcolor{FalseC}{$-0.037$} & \textcolor{FalseC}{$-0.005^{\scriptscriptstyle 5}$} & \textcolor{FalseC}{$-0.004^{\scriptscriptstyle 4}$} & \textcolor{TrueC}{$+0.021^{\scriptscriptstyle 20}$} & \textcolor{TrueC}{$+0.039^{\scriptscriptstyle 37}$} & \textcolor{TrueC}{$+0.035^{\scriptscriptstyle 30}$} & \textcolor{TrueC}{$+0.057^{\scriptscriptstyle 48}$} & \textcolor{TrueC}{$+0.044^{\scriptscriptstyle 36}$} & \textcolor{TrueC}{$+0.065^{\scriptscriptstyle 51}$} & $\times2$ \\
\rowcolor{FalseC!8}
\texttt{L40.h11.svd18} & ov svd & truth & \textcolor{FalseC}{$-0.027$} & \textcolor{FalseC}{$-0.044$} & \textcolor{FalseC}{$-0.012^{\scriptscriptstyle 15}$} & \textcolor{FalseC}{$-0.015^{\scriptscriptstyle 18}$} & \textcolor{TrueC}{$+0.009^{\scriptscriptstyle 10}$} & \textcolor{TrueC}{$+0.008^{\scriptscriptstyle 10}$} & \textcolor{TrueC}{$+0.026^{\scriptscriptstyle 26}$} & \textcolor{TrueC}{$+0.020^{\scriptscriptstyle 22}$} & \textcolor{TrueC}{$+0.037^{\scriptscriptstyle 36}$} & \textcolor{TrueC}{$+0.027^{\scriptscriptstyle 29}$} & $\times2$ \\
\rowcolor{FalseC!8}
\texttt{L40.h7.c105} & ov neuron & real & \textcolor{FalseC}{$-0.015$} & \textcolor{FalseC}{$-0.014$} & \textcolor{FalseC}{$-0.003^{\scriptscriptstyle 4}$} & \textcolor{FalseC}{$-0.003^{\scriptscriptstyle 4}$} & \textcolor{TrueC}{$+0.006^{\scriptscriptstyle 8}$} & \textcolor{TrueC}{$+0.005^{\scriptscriptstyle 8}$} & \textcolor{TrueC}{$+0.012^{\scriptscriptstyle 15}$} & \textcolor{TrueC}{$+0.013^{\scriptscriptstyle 16}$} & \textcolor{TrueC}{$+0.017^{\scriptscriptstyle 19}$} & \textcolor{TrueC}{$+0.017^{\scriptscriptstyle 20}$} & $\times2$ \\
\rowcolor{FalseC!8}
\texttt{L40.h7.c177} & ov neuron & genuine & \textcolor{FalseC}{$-0.011$} & \textcolor{FalseC}{$-0.014$} & \textcolor{FalseC}{$-0.005^{\scriptscriptstyle 6}$} & \textcolor{FalseC}{$-0.004^{\scriptscriptstyle 6}$} & \textcolor{TrueC}{$+0.003^{\scriptscriptstyle 5}$} & \textcolor{TrueC}{$+0.002^{\scriptscriptstyle 2}$} & \textcolor{TrueC}{$+0.008^{\scriptscriptstyle 10}$} & \textcolor{TrueC}{$+0.007^{\scriptscriptstyle 9}$} & \textcolor{TrueC}{$+0.011^{\scriptscriptstyle 13}$} & \textcolor{TrueC}{$+0.010^{\scriptscriptstyle 12}$} & $\times2$ \\
\rowcolor{black!6}
\texttt{L36.n1119} & mlp neuron & truth & \textcolor{black!55}{$+0.041$} & \textcolor{black!55}{$+0.207$} & $+0.002$ & $+0.001$ & \textcolor{FalseC}{$-0.122^{\scriptscriptstyle 46}$} & \textcolor{FalseC}{$-0.222^{\scriptscriptstyle 62}$} & \textcolor{FalseC}{$-0.176^{\scriptscriptstyle 57}$} & \textcolor{FalseC}{$-0.166^{\scriptscriptstyle 42}$} & \textcolor{FalseC}{$-0.210^{\scriptscriptstyle 65}$} & \textcolor{FalseC}{$-0.104^{\scriptscriptstyle 31}$} & -- \\
\bottomrule\end{tabular}}\end{table*}

\begin{table*}[!htbp]\centering\footnotesize\setlength{\tabcolsep}{3pt}
\caption{\textbf{Backup recruitment across dose under JOINT ablation of both Mistral-7B-Instruct-v0.3 verdict cores} (\texttt{L20.n14286}$+$\texttt{L19.n8228}, contrastive swap, \texttt{sec\_ref=regime}, $n{=}598$). \textbf{Rows are the clean anti/null axes that become verdict contributors under the ablation} {(the recruitment set). Clean contributors} and never-flipping null controls are omitted (they are in the total-effect scan). Columns as in Table~\ref{tab:llama_regime_recruit}. Across the coloured entries the median $t\approx20$ (up to $t\approx129$). {$\kappa_{\mathrm{base}}=-11.6$, and joint} $\kappa_{\mathrm{primary}}$ at $\times1..\times4=-5.0,+3.9,+10.5,+15.9$ {(crosses $0$ at $\times2$, where the verdict flips).}}
\label{tab:mistral_regime_recruit}
\resizebox{\textwidth}{!}{\begin{tabular}{lll rr rr rr rr rr c}
\toprule
 & & & \multicolumn{2}{c}{$E^0$ (clean)} & \multicolumn{2}{c}{$E^{\pi}\times1$} & \multicolumn{2}{c}{$E^{\pi}\times2$} & \multicolumn{2}{c}{$E^{\pi}\times3$} & \multicolumn{2}{c}{$E^{\pi}\times4$} & \\
\cmidrule(lr){4-5}\cmidrule(lr){6-7}\cmidrule(lr){8-9}\cmidrule(lr){10-11}\cmidrule(lr){12-13}
secondary & family & detects & $T$ & $F$ & $T$ & $F$ & $T$ & $F$ & $T$ & $F$ & $T$ & $F$ & flip@ \\
\midrule
\rowcolor{FalseC!8}
\texttt{L31.n13669} & mlp neuron & true & \textcolor{FalseC}{$-1.929$} & \textcolor{FalseC}{$-2.083$} & \textcolor{FalseC}{$-1.470^{\scriptscriptstyle 88}$} & \textcolor{FalseC}{$-1.624^{\scriptscriptstyle 79}$} & \textcolor{TrueC}{$+0.227^{\scriptscriptstyle 7}$} & \textcolor{TrueC}{$+0.209^{\scriptscriptstyle 6}$} & \textcolor{TrueC}{$+0.736^{\scriptscriptstyle 43}$} & \textcolor{TrueC}{$+0.793^{\scriptscriptstyle 44}$} & \textcolor{TrueC}{$+0.522^{\scriptscriptstyle 40}$} & \textcolor{TrueC}{$+0.583^{\scriptscriptstyle 40}$} & $\times2$ \\
\rowcolor{FalseC!8}
\texttt{L31.n8773} & mlp neuron & false & \textcolor{FalseC}{$-0.771$} & \textcolor{FalseC}{$-0.748$} & $+0.024$ & \textcolor{TrueC}{$+0.050^{\scriptscriptstyle 2}$} & \textcolor{TrueC}{$+1.088^{\scriptscriptstyle 61}$} & \textcolor{TrueC}{$+1.240^{\scriptscriptstyle 56}$} & \textcolor{TrueC}{$+1.489^{\scriptscriptstyle 78}$} & \textcolor{TrueC}{$+1.754^{\scriptscriptstyle 72}$} & \textcolor{TrueC}{$+1.438^{\scriptscriptstyle 58}$} & \textcolor{TrueC}{$+1.732^{\scriptscriptstyle 56}$} & $\times2$ \\
\rowcolor{FalseC!8}
\texttt{L28.n4079} & mlp neuron & true & \textcolor{FalseC}{$-0.254$} & \textcolor{FalseC}{$-0.757$} & \textcolor{FalseC}{$-0.062^{\scriptscriptstyle 33}$} & \textcolor{FalseC}{$-0.149^{\scriptscriptstyle 38}$} & \textcolor{TrueC}{$+0.415^{\scriptscriptstyle 37}$} & \textcolor{TrueC}{$+0.325^{\scriptscriptstyle 56}$} & \textcolor{TrueC}{$+0.958^{\scriptscriptstyle 55}$} & \textcolor{TrueC}{$+0.511^{\scriptscriptstyle 95}$} & \textcolor{TrueC}{$+1.153^{\scriptscriptstyle 88}$} & \textcolor{TrueC}{$+0.569^{\scriptscriptstyle 129}$} & $\times2$ \\
\rowcolor{FalseC!8}
\texttt{L22.h23.svd2} & ov svd & wrong & \textcolor{FalseC}{$-0.092$} & \textcolor{FalseC}{$-0.132$} & \textcolor{FalseC}{$-0.049^{\scriptscriptstyle 29}$} & \textcolor{FalseC}{$-0.092^{\scriptscriptstyle 37}$} & \textcolor{FalseC}{$-0.018^{\scriptscriptstyle 12}$} & \textcolor{FalseC}{$-0.017^{\scriptscriptstyle 10}$} & \textcolor{TrueC}{$+0.010^{\scriptscriptstyle 8}$} & \textcolor{TrueC}{$+0.018^{\scriptscriptstyle 11}$} & \textcolor{TrueC}{$+0.032^{\scriptscriptstyle 25}$} & \textcolor{TrueC}{$+0.053^{\scriptscriptstyle 36}$} & $\times3$ \\
\rowcolor{FalseC!8}
\texttt{L22.n8735} & mlp neuron & false & \textcolor{FalseC}{$-0.123$} & \textcolor{FalseC}{$-0.083$} & \textcolor{FalseC}{$-0.045^{\scriptscriptstyle 16}$} & \textcolor{FalseC}{$-0.096^{\scriptscriptstyle 23}$} & \textcolor{TrueC}{$+0.018^{\scriptscriptstyle 5}$} & \textcolor{FalseC}{$-0.034^{\scriptscriptstyle 6}$} & \textcolor{TrueC}{$+0.025^{\scriptscriptstyle 11}$} & \textcolor{TrueC}{$+0.058^{\scriptscriptstyle 8}$} & \textcolor{TrueC}{$+0.008^{\scriptscriptstyle 5}$} & \textcolor{TrueC}{$+0.112^{\scriptscriptstyle 18}$} & $\times3$ \\
\rowcolor{FalseC!8}
\texttt{L30.n11434} & mlp neuron & false & \textcolor{FalseC}{$-0.111$} & \textcolor{FalseC}{$-0.068$} & \textcolor{FalseC}{$-0.013^{\scriptscriptstyle 11}$} & \textcolor{FalseC}{$-0.017^{\scriptscriptstyle 14}$} & \textcolor{TrueC}{$+0.010^{\scriptscriptstyle 11}$} & \textcolor{TrueC}{$+0.024^{\scriptscriptstyle 19}$} & \textcolor{TrueC}{$+0.030^{\scriptscriptstyle 24}$} & \textcolor{TrueC}{$+0.103^{\scriptscriptstyle 51}$} & \textcolor{TrueC}{$+0.044^{\scriptscriptstyle 34}$} & \textcolor{TrueC}{$+0.151^{\scriptscriptstyle 68}$} & $\times2$ \\
\rowcolor{FalseC!8}
\texttt{L22.n1070} & mlp neuron & accurate & \textcolor{FalseC}{$-0.024$} & \textcolor{FalseC}{$-0.040$} & \textcolor{FalseC}{$-0.017^{\scriptscriptstyle 13}$} & \textcolor{FalseC}{$-0.032^{\scriptscriptstyle 21}$} & $+0.001$ & $-0.002$ & \textcolor{TrueC}{$+0.008^{\scriptscriptstyle 6}$} & \textcolor{TrueC}{$+0.008^{\scriptscriptstyle 7}$} & \textcolor{TrueC}{$+0.013^{\scriptscriptstyle 11}$} & \textcolor{TrueC}{$+0.014^{\scriptscriptstyle 12}$} & $\times3$ \\
\rowcolor{FalseC!8}
\texttt{L20.n4786} & mlp neuron & right & \textcolor{FalseC}{$-0.013$} & \textcolor{FalseC}{$-0.003$} & \textcolor{FalseC}{$-0.008^{\scriptscriptstyle 6}$} & \textcolor{FalseC}{$-0.009^{\scriptscriptstyle 7}$} & $+0.002$ & \textcolor{TrueC}{$+0.005^{\scriptscriptstyle 5}$} & \textcolor{TrueC}{$+0.006^{\scriptscriptstyle 5}$} & \textcolor{TrueC}{$+0.016^{\scriptscriptstyle 12}$} & \textcolor{TrueC}{$+0.005^{\scriptscriptstyle 4}$} & \textcolor{TrueC}{$+0.025^{\scriptscriptstyle 18}$} & $\times3$ \\
\bottomrule\end{tabular}}\end{table*}

\subsection{Dose-resolved recruitment}
\label{app:recruitment-figures}
\begin{figure*}[!htbp]
  \centering
  \includegraphics[width=\linewidth]{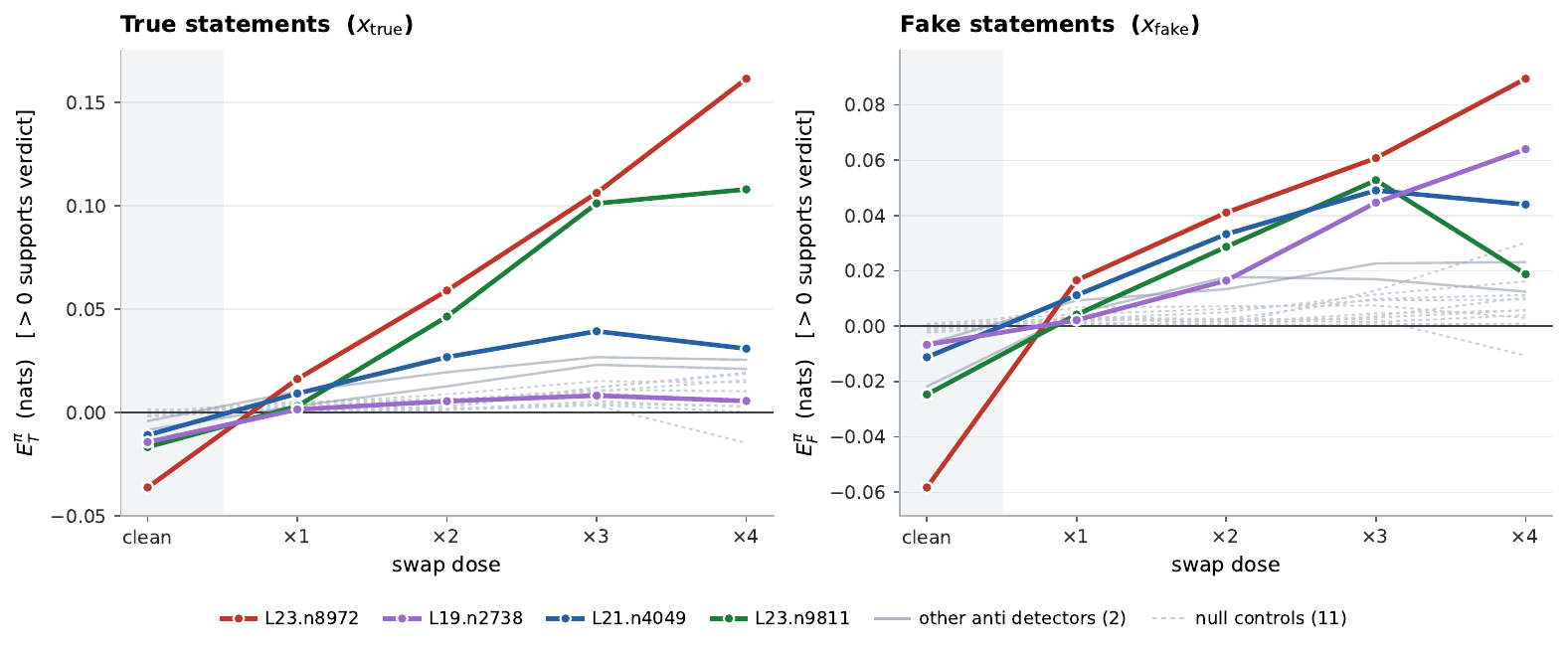}
  \caption[SCM-faithful recruitment of dormant truth detectors under ablation]{    \textbf{Interpretable truth detectors that are dormant or opposing in the
    clean model become a verdict contributor when the {core neuron} is ablated, in proportion to dose and on both prompt classes.}
    For each named secondary $r$ we exchange its activation with the value it
    takes on the contrastive partner \emph{within the ablated regime}
{($\mathrm{sec}_r$ in Eq.~\ref{eq:reference})}, the SCM-faithful
    counterfactual, and read off $E^{\pi}_r$ (Eq.~{\ref{eq:contribution}}), the
    faithful True$-$False margin the swap erases in the ablated world $\pi$
    ($E>0$ \emph{supports} the verdict, $E<0$ \emph{opposes} it). The
    horizontal axis sweeps the dose from the clean baseline $E^{0}$
    (leftmost) through swap scales $k\in\{1,2,3,4\}$, driving the primary
    neuron \texttt{L18.n11065} progressively past its contrastive value.
{The \textbf{left} panel shows true statements, the \textbf{right} fake statements.} Each
    detector starts in the dormant/opposing band ($E\le 0$) and rises into the
    contributor band ($E>0$) {as the ablation deepens, monotonically} for the
    core detectors, led by the lie-writer \texttt{L23.n8972}. {A} contributor must be positive on {\emph{both} panels, and the both-classes gate}
    also exposes the lone one-sided case, \texttt{L24.n4543}
    (\emph{yes}/\emph{true}), which rises on true statements but reverses on
    fake statements at high dose. Junk-poled directions (grey dashed) stay
    pinned at zero. \textsc{Llama-3-8B-Instruct}.}
  \label{fig:dose-llama}
\end{figure*}

\begin{figure}[!htbp]\centering
  \includegraphics[width=\linewidth]{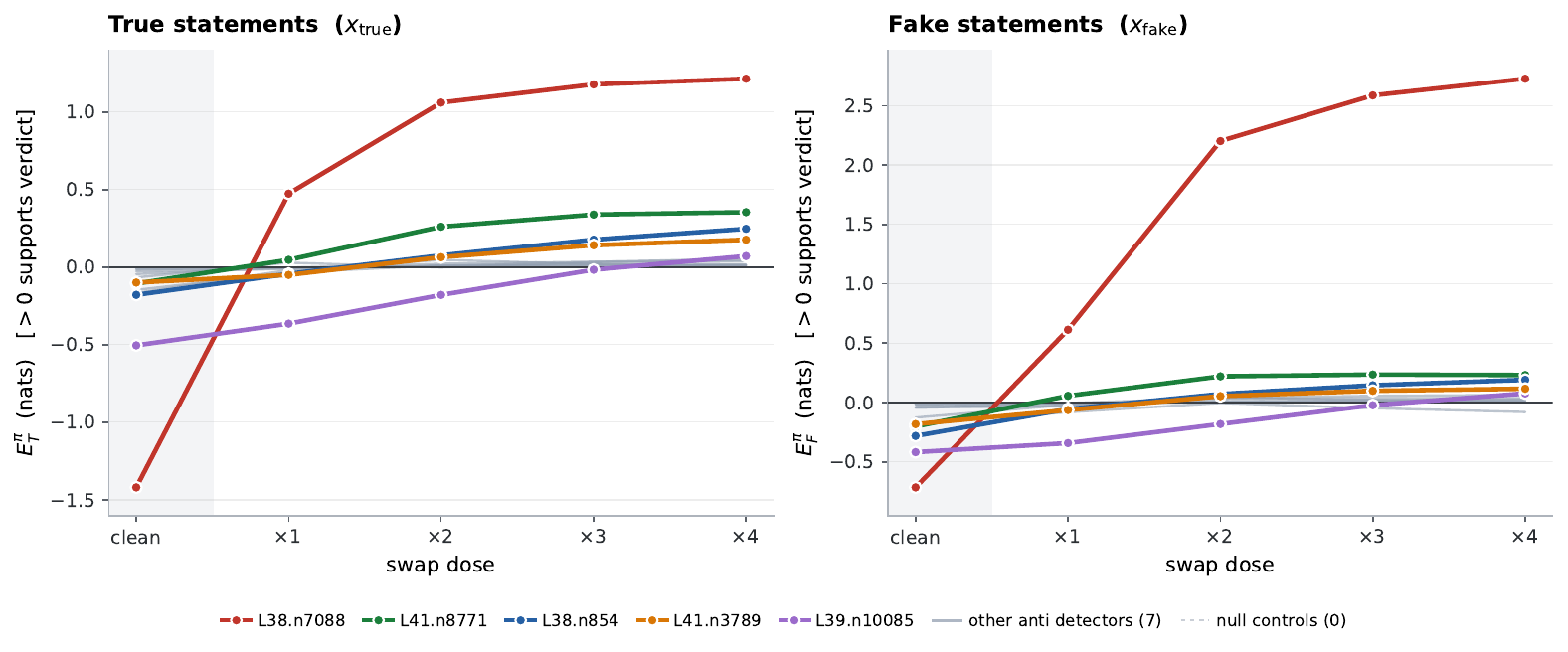}
  \caption{\textbf{Dose-resolved recruitment under joint ablation of both Gemma-2-9b-it verdict cores}
  (\texttt{L28.n2046}$+$\texttt{L33.n4294}). Per-side erased margin $E^{\pi}$ vs.\ swap dose
  (\emph{clean}$=E^0$, then {$\times1..\times4$), with true statements on the left and fake statements on the right. Each coloured curve is one recruited anti-detector, and the dotted grey curves are the null controls.} As the joint swap
  strengthens, the opposing detectors are pulled from $E^{\pi}<0$ up through zero into positive
  (verdict-supporting) territory, while the null controls stay {flat, so recruitment is specific to truth axes, including opposing ones.}}
  \label{fig:gemma_dose}
\end{figure}

\begin{figure}[!htbp]
  \centering
  \includegraphics[width=\linewidth]{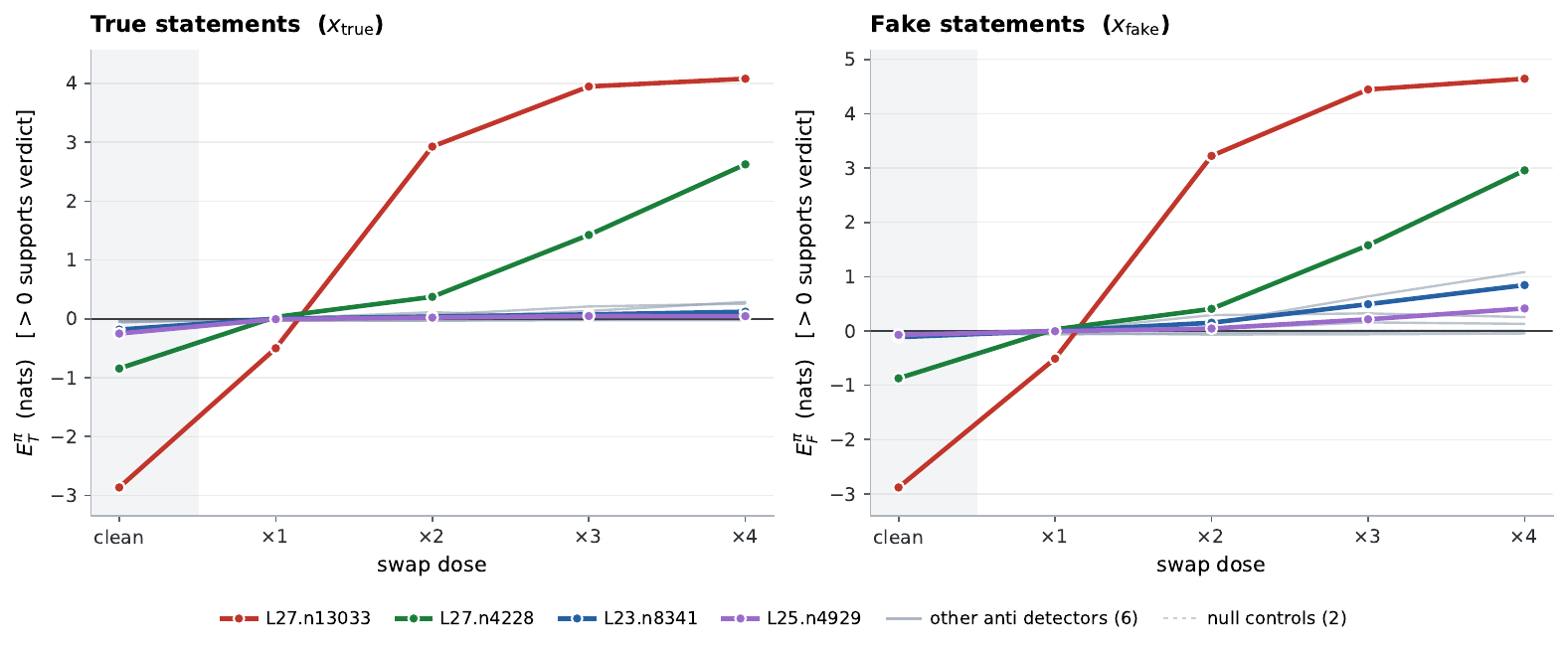}
  \caption{\textbf{Dose-resolved recruitment under joint ablation of both
  Qwen2.5-7B-Instruct verdict cores} (\texttt{L22.n13149}$+$\texttt{L24.n14758},
  contrastive swap, \texttt{sec\_ref=regime}, $n{=}100$). Per-detector erased
  margin $E^{\pi}$ on true statements (left) and fake statements (right) as the
  joint swap dose increases {from clean to $\times4$, where $E^{\pi}>0$ means} the
  detector now supports the verdict. The four identified detectors (colour)
  start as clean opponents ($E^0<0$, shaded clean band) and are {progressively
  recruited. \texttt{L27.n13033}} (\emph{false}) rockets from $-2$ to $+3.7/+4.2$
  nats, flipping at $\times2$, and \texttt{L27.n4228} (suppresses \emph{true})
{flips at $\times1$. The remaining anti detectors} (slate) follow the same
  upward drift at smaller magnitude, while the null controls (dotted) stay
  pinned at zero across all doses. The joint ablation flips
  the verdict at $\times2$ ($\kappa_{\mathrm{primary}}$ crosses $0$), so beyond
  that the panels report reconstitution of the erased fact/fake signal by the
  backup detectors.}
  \label{fig:qwen_dose}
\end{figure}

\begin{figure*}[!htbp]
  \centering
  \includegraphics[width=\linewidth]{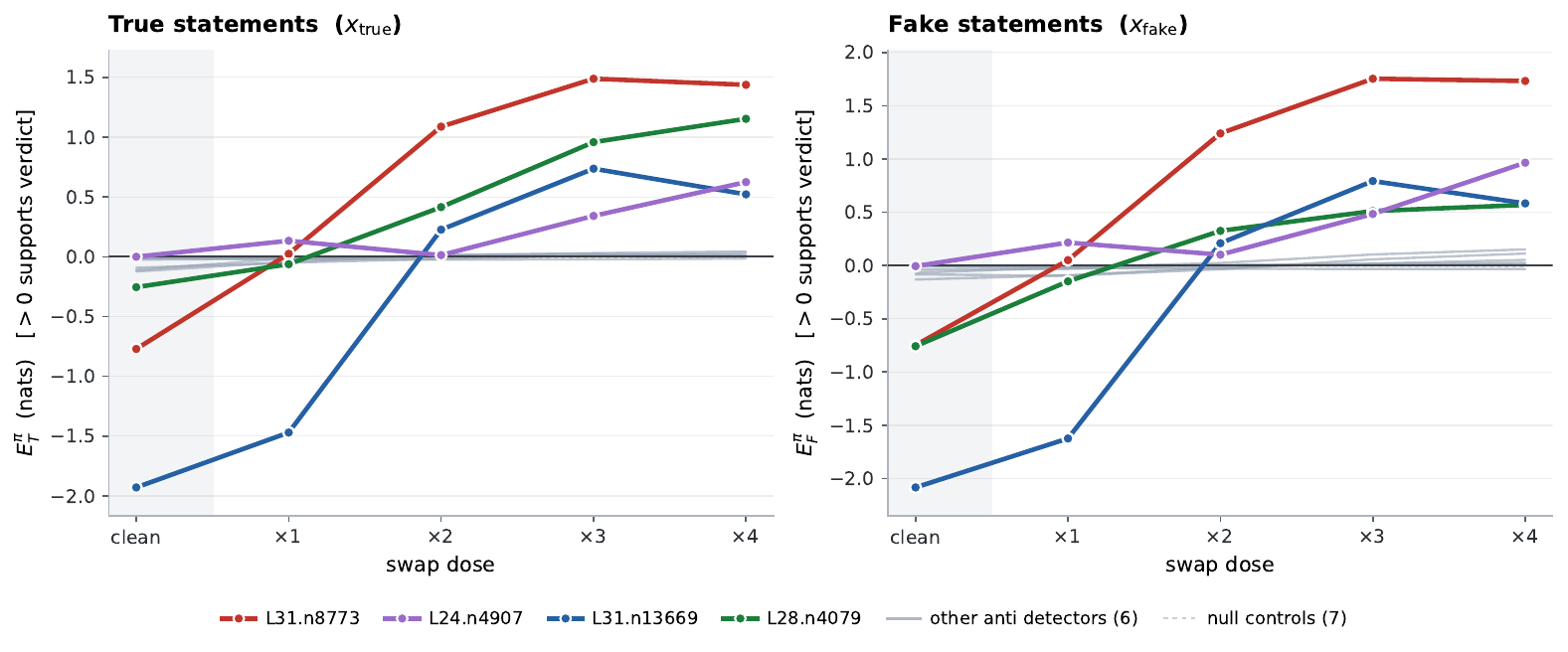}
  \caption{\textbf{Dose-resolved recruitment under joint ablation of both
  Mistral-7B-Instruct-v0.3 verdict cores} (\texttt{L20.n14286}$+$\texttt{L19.n8228},
  contrastive swap, \texttt{sec\_ref=regime}, $n{=}598$). Per-side erased
  margin $E^{\pi}$ vs.\ swap dose (\emph{clean}$=E^0$, then {$\times1..\times4$), with true statements on the left and fake statements on the right, following the layout of Figs.~\ref{fig:gemma_dose} to \ref{fig:qwen_dose}.} The eight recruited
  anti-detectors of Table~\ref{tab:mistral_regime_recruit} are pulled from
  $E^{\pi}<0$ through zero as the joint swap {strengthens. \texttt{L31.n8773} and \texttt{L31.n13669} show the largest swings (clean $E^0\approx-0.8$ and $-1.9$ to $-2.1$, rising to $E^{\pi}\approx+1.4$ to $1.8$ and $+0.5$ to $0.8$ by $\times3$ to $\times4$), and \texttt{L28.n4079} reaches $+1.15$ on the true side by $\times4$}, while
  \texttt{L22.h23.svd2}, \texttt{L22.n8735}, \texttt{L22.n1070} and
  \texttt{L20.n4786} recruit later and more weakly, consistent with the
  \texttt{flip@} column of Table~\ref{tab:mistral_regime_recruit}.}
  \label{fig:mistral_dose}
\end{figure*}

{{Figures~\ref{fig:dose-llama} to \ref{fig:mistral_dose}} plot the same contributions against dose.}

\subsection{Single-core versus joint-core recruitment}
\label{app:recruitment-svj}

\begin{figure}[!htbp]\centering
\includegraphics[width=\linewidth]{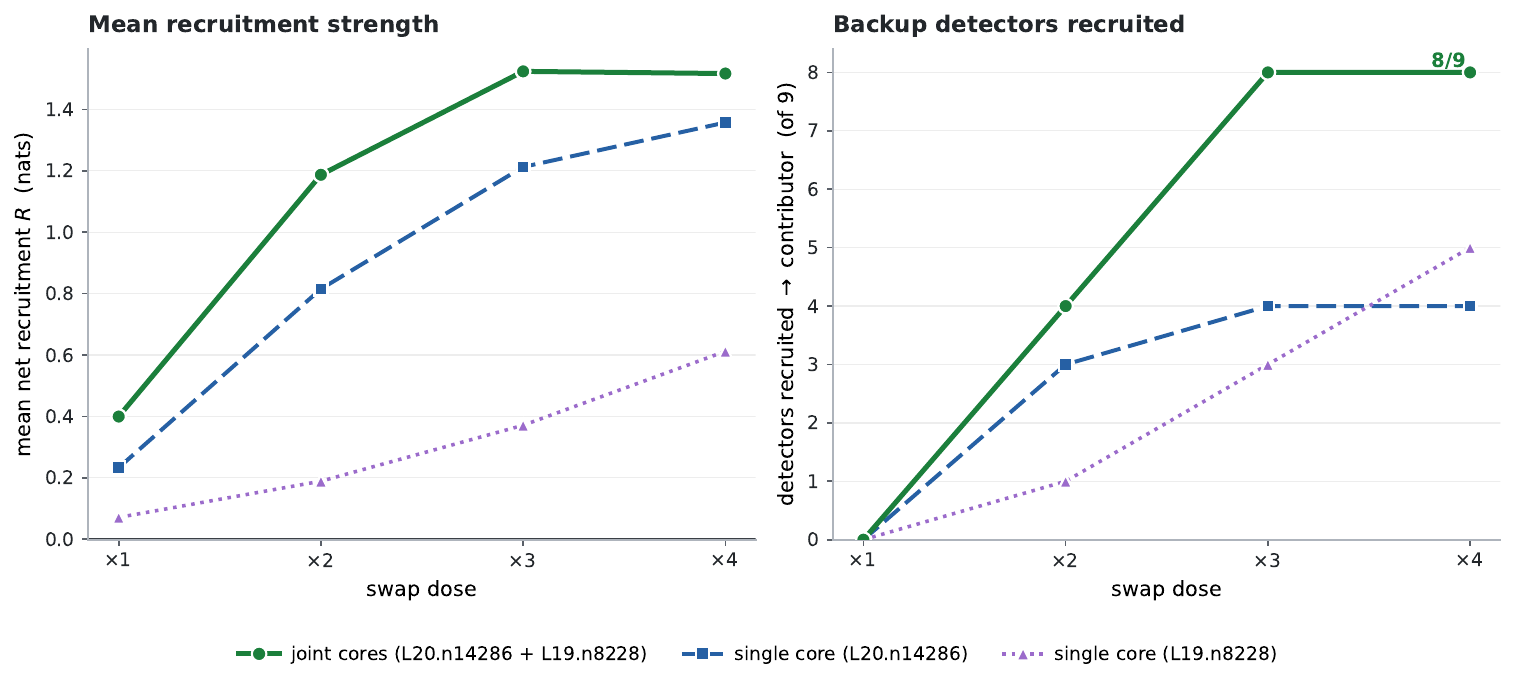}
\caption{\textbf{Removing the verdict as a whole recruits more backup than
removing either core separately} (Mistral-7B-Instruct-v0.3, $n{=}598$). {Over the $9$ axes that oppose the verdict in the clean model, the panels show mean net recruitment $R$}
(left) and the number crossing into net contribution (right), with the cores
\texttt{L20.n14286} and \texttt{L19.n8228} ablated jointly (solid) or each alone
(dashed, dotted). Joint ablation dominates at every dose ($R{=}1.52$, $8/9$
recruited at $\times4$, vs.\ $1.36$/$4$ and $0.61$/$5$ for the single cores) and
is super-additive at low dose, exceeding the sum of the two single-core effects at
$\times1$ and $\times2$ before saturating.}
\label{fig:mistral_pop_svj_short}\end{figure}

\begin{figure}[!htbp]
  \centering
  \includegraphics[width=\linewidth]{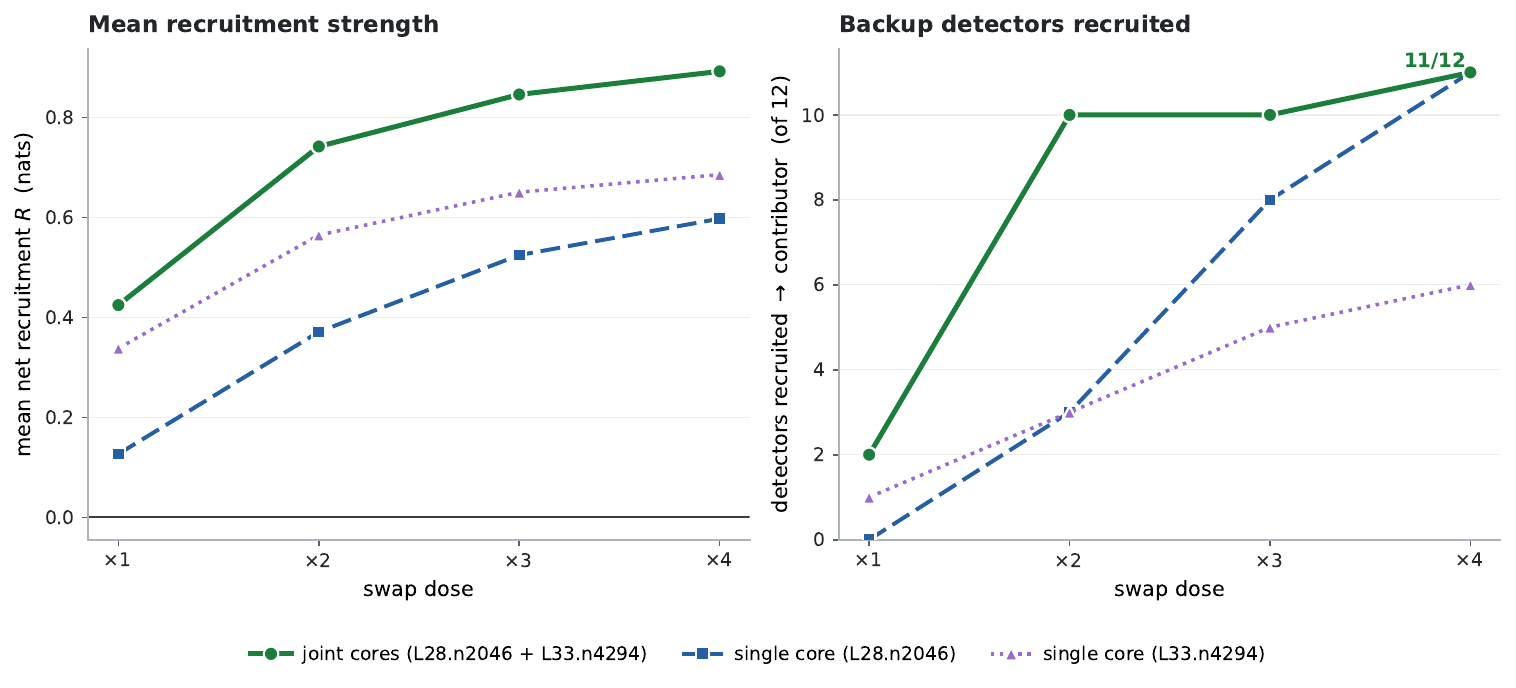}
  \caption{\textbf{Single-core vs.\ joint-core recruitment across dose}
  (Gemma-2-9b-it, contrastive swap, \texttt{sec\_ref=regime}, $n{=}100$).
  Recruitment of the $12$ clean \emph{anti} detectors when we ablate each
  verdict core \emph{alone} {(\texttt{L28.n2046} dashed, \texttt{L33.n4294} dotted)} versus \emph{jointly} (both cores, solid), as a function of the swap
  dose $\times1..\times4$. {The \textbf{left} panel shows mean net recruitment $R=R_T+R_F$ averaged over the $12$ detectors. The joint ablation dominates}
  either core alone at every dose {($0.43/0.74/0.85/0.89$ vs.\ \texttt{L33} $0.34/0.57/0.65/0.69$ vs.\ \texttt{L28} $0.13/0.37/0.53/0.60$)}. {The \textbf{right} panel shows the number of the $12$ that cross} into net contribution ($E^{\pi}>0$ on both sides). {The joint ablation recruits $10/12$ by $\times2$ and $11/12$ at $\times4$. \texttt{L33} leads at $\times1$ ($1$ vs.\ $0$), the two single cores tie at $\times2$ ($3$ each), and \texttt{L28} then overtakes, $8$ vs.\ $5$ at $\times3$ and $11$ vs.\ $6$ at $\times4$, where it matches the joint count. No single core reproduces the joint mean recruitment.}}
  \label{fig:gemma_svj_agg}
\end{figure}

\begin{figure}[!htbp]
  \centering
  \includegraphics[width=\linewidth]{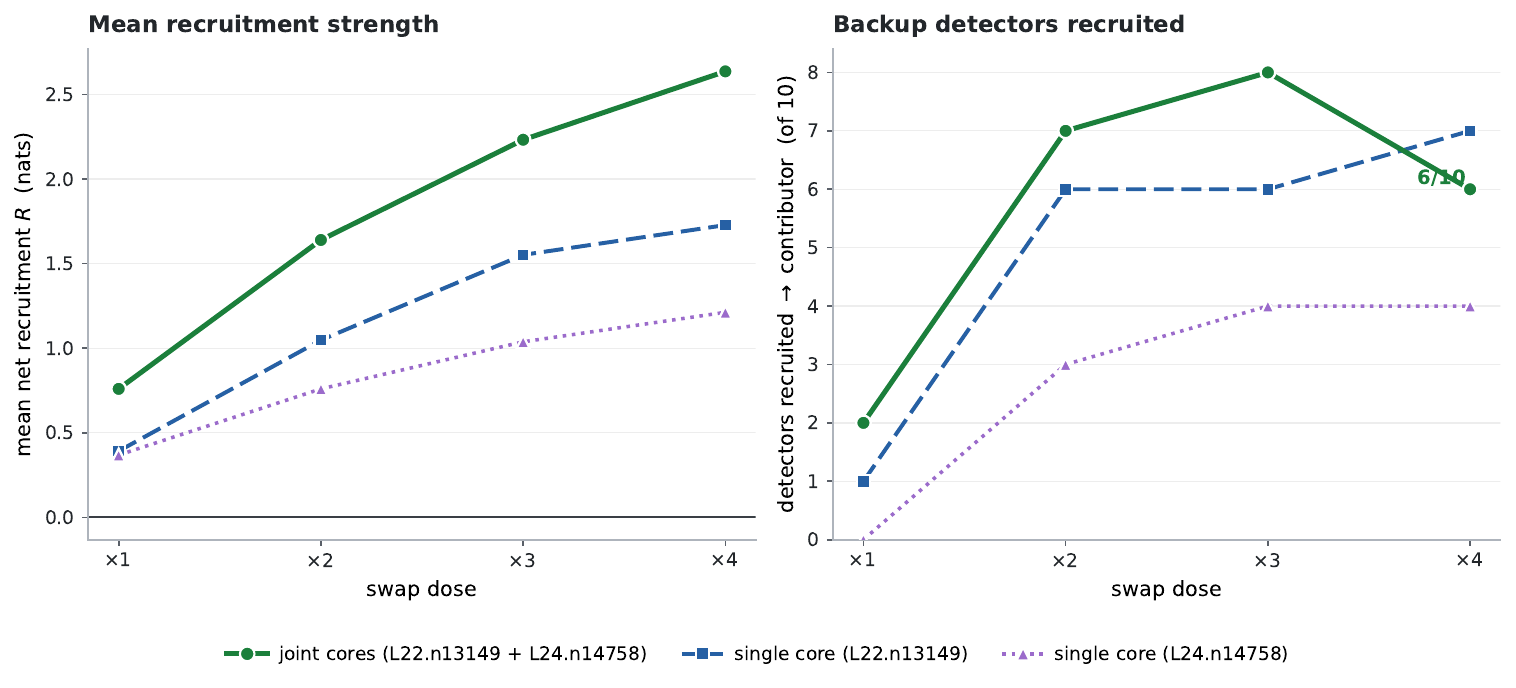}
  \caption{\textbf{Single-core vs.\ joint-core recruitment across dose}
  (Qwen2.5-7B-Instruct, contrastive swap, \texttt{sec\_ref=regime}, $n{=}100$).
  Recruitment of the {$10$} clean \emph{anti} detectors when we ablate the dominant
  truth core \texttt{L22.n13149} alone (dashed), the correctness co-core
  \texttt{L24.n14758} alone (dotted), or both \emph{jointly} (solid), vs.\ swap
  dose $\times1..\times4$. {The \textbf{left} panel shows mean net recruitment $R=R_T+R_F$ over the anti detectors. The joint ablation dominates} either core alone at every
  dose, but \texttt{L22.n13149} alone{, which carries $\sim\!35\%$ of the clean fact/fake margin,} recovers most of it, while the correctness co-core
  \texttt{L24.n14758} recruits far less. {The \textbf{right} panel shows the number crossing into net contribution ($E^{\pi}>0$ both sides). The joint ablation peaks at $8/10$ at $\times3$ and falls to $6/10$ at $\times4$, while \texttt{L22.n13149} alone reaches $7/10$ and \texttt{L24.n14758} alone $4/10$.} Unlike
  Gemma-2-9b, whose two co-equal cores each recruit only partially, Qwen's
  backup recruitment is carried predominantly by the single dominant core.}
  \label{fig:qwen_svj_agg}
\end{figure}

\begin{figure}[!htbp]\centering
\includegraphics[width=\linewidth]{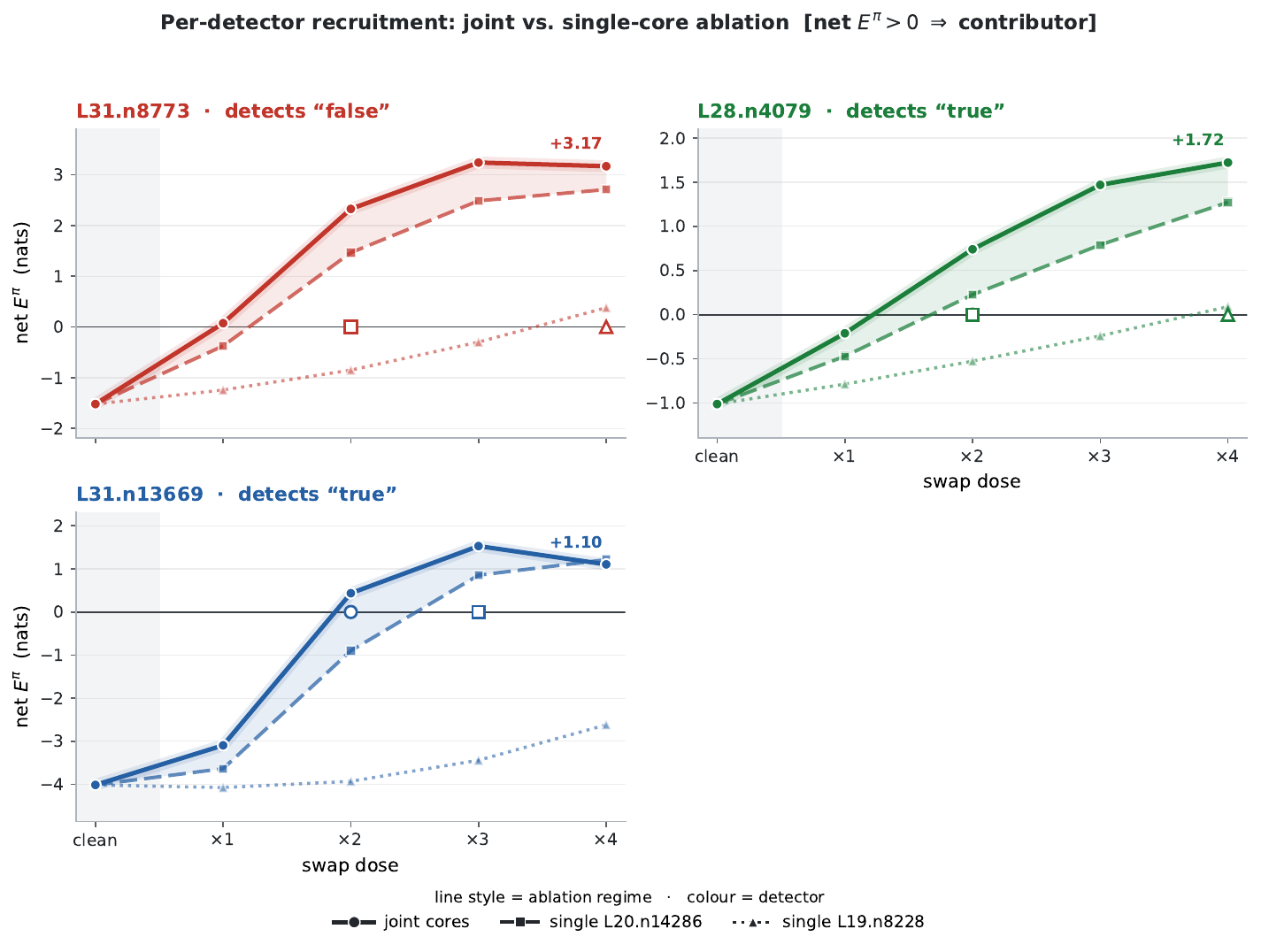}
\caption{\textbf{Backup recruitment is {driven by the cores jointly rather than by either core alone}} (Mistral-7B-Instruct-v0.3, contrastive swap, \texttt{sec\_ref=regime},
perfect-on-pairs partition, $n{=}598$). One panel per recruited detector, showing
its net erased margin $E^{\pi}=E^{\pi}_{T}+E^{\pi}_{F}$ against swap dose when the
two verdict cores \texttt{L20.n14286} and \texttt{L19.n8228} are ablated
\emph{jointly} (solid) or each \emph{alone} (dashed, dotted). $E^{\pi}>0$ means the
detector now {supports the verdict, and the shaded band} is the gain from joint ablation
beyond whichever single core does better. Every panel is recruited by the joint
{ablation at $\times2$. \texttt{L31.n8773}} moves $-1.52\!\rightarrow\!+3.17$,
\texttt{L28.n4079} $-1.01\!\rightarrow\!+1.72$, \texttt{L31.n13669}
$-4.01\!\rightarrow\!+1.10$ (clean $\rightarrow$ joint at $\times4$).
Single-core ablation is {strictly weaker and asymmetric. \texttt{L20.n14286}}
alone reaches $+2.71$, $+1.27$ and $+1.21$ and recruits $4$ of the $9$
clean-anti axes in the bench, \texttt{L19.n8228} alone reaches only $+0.38$,
$+0.09$ and $-2.62$ {(it never recruits \texttt{L31.n13669} at any dose),} and both together as a joint ablation recruit $8$ of $9$.}
\label{fig:mistral_components_svj}\end{figure}

\begin{figure}[!htbp]
  \centering
  \includegraphics[width=\linewidth]{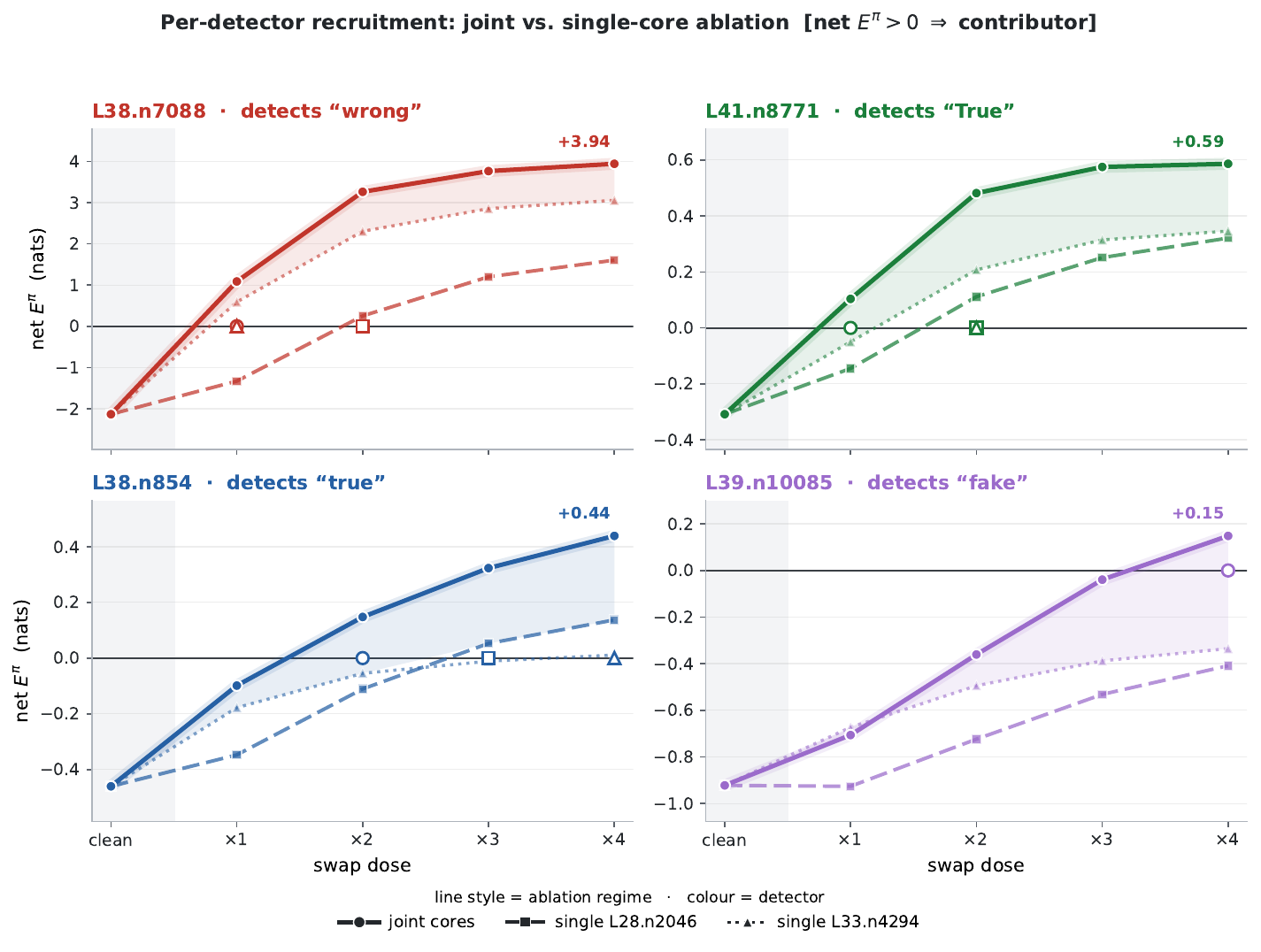}
  \caption{\textbf{Backup detectors are recruited into verdict contributors, and
  more strongly under joint than single-core ablation (Gemma-2-9b-it).} Each
  panel follows one of the four identified {secondary detectors, and the net erased margin $E^{\pi}=E^{\pi}_T+E^{\pi}_F$ is plotted at the clean state ($E^{0}$) and at contrastive-swap doses $\times1$ to $\times4$, under} \emph{joint}
  ablation of both verdict cores (\texttt{L28.n2046}$+$\texttt{L33.n4294},
  solid) versus each core alone (\texttt{L28.n2046} dashed, \texttt{L33.n4294}
  dotted). All four are \emph{anti} {directions at rest ($E^{0}<0$, with values}
  $-2.14,-0.31,-0.46,-0.92$ for \texttt{L38.n7088}, \texttt{L41.n8771},
  \texttt{L38.n854}, \texttt{L39.n10085}) yet cross zero under ablation. Joint
  ablation flips \texttt{L38.n7088} and \texttt{L41.n8771} already at
  $\times1$, \texttt{L38.n854} at $\times2$, and even the initially strongly
  opposing \texttt{L39.n10085} by $\times4$. $n{=}583$ perfect-on-pairs
  prompts.}
  \label{fig:gemma_svj}
\end{figure}

\begin{figure}[!htbp]
  \centering
  \includegraphics[width=\linewidth]{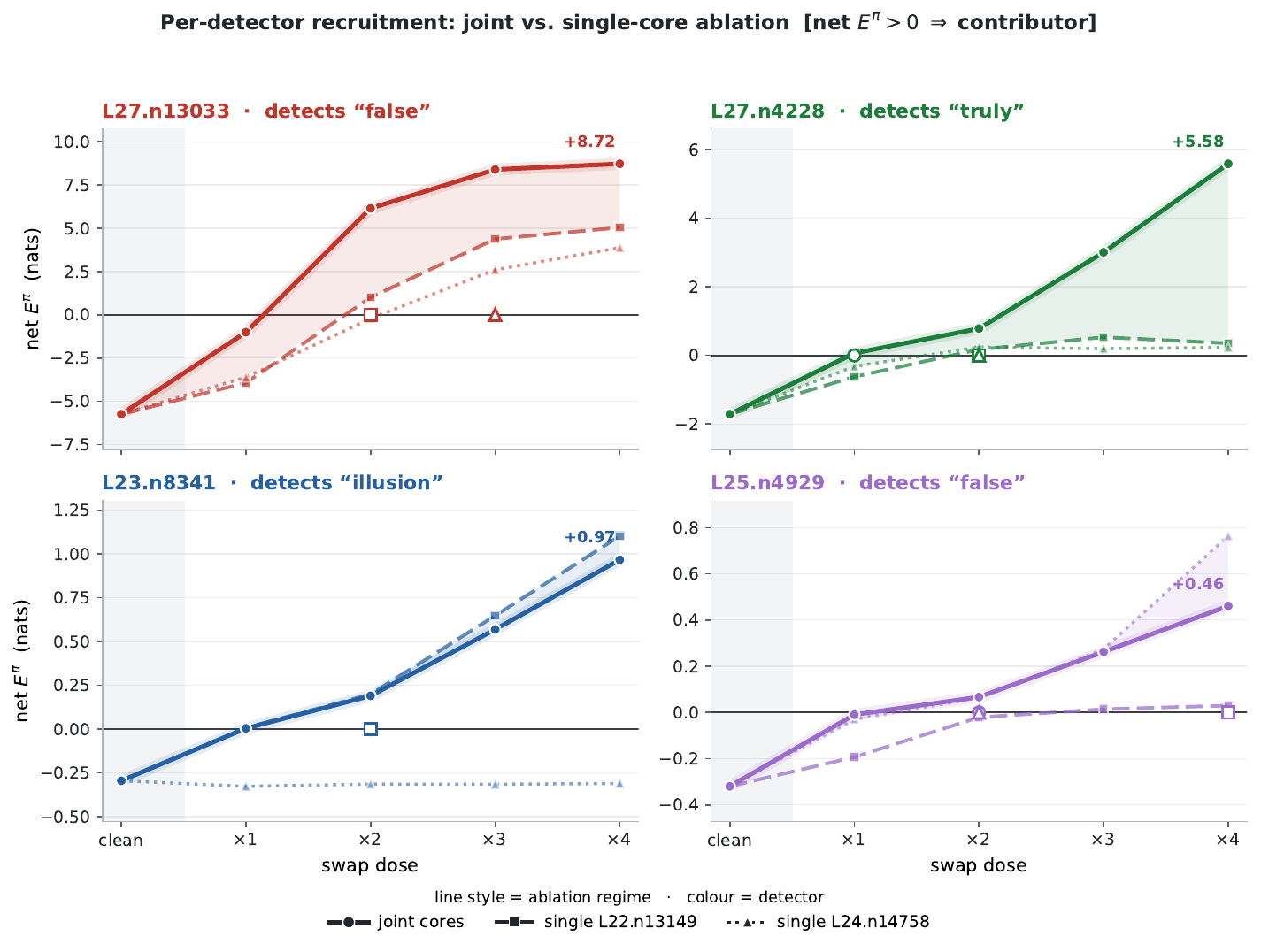}
  \caption{\textbf{Backup detectors are recruited into verdict contributors, and
  more strongly under joint than single-core ablation (Qwen2.5-7B-Instruct).}
  Each panel follows one of the four identified {secondary detectors and plots the net erased margin $E^{\pi}=E^{\pi}_T+E^{\pi}_F$ at the clean state ($E^{0}$) and at contrastive-swap doses $\times1$ to $\times4$, under} \emph{joint} ablation
  of both verdict cores (\texttt{L22.n13149}$+$\texttt{L24.n14758}, solid)
  versus each core alone (truth core \texttt{L22.n13149} dashed, correctness
  co-core \texttt{L24.n14758} dotted). All four are \emph{anti} at rest
{($E^{0}<0$, with values $-5.74,-1.71,-0.29,-0.32$} for \texttt{L27.n13033},
  \texttt{L27.n4228}, \texttt{L23.n8341}, \texttt{L25.n4929}). Joint ablation
  flips \texttt{L27.n4228} at $\times1$ and the {remaining three by $\times2$.
  Several} directions are recruited by only one of the two cores unless both
  are removed together. $n{=}478$ perfect-on-pairs prompts.}
  \label{fig:qwen_svj}
\end{figure}

Gemma, Qwen, and Mistral each carry two {cores, while Llama carries one,}
so no single-vs-joint comparison exists for it. Figures~\ref{fig:mistral_pop_svj_short},
\ref{fig:gemma_svj_agg}, and~\ref{fig:qwen_svj_agg} give the aggregate
recruitment (mean $R$ and count crossing into contribution) under each core
alone versus both jointly. The per-detector breakdowns
(Figs.~\ref{fig:mistral_components_svj}, \ref{fig:gemma_svj}, \ref{fig:qwen_svj})
are kept as a distinct, lower-priority {subsection rather than cut, because they show which specific detector each single core recruits, information the aggregates lack} (e.g.\ Mistral's \texttt{L19.n8228} alone never recruits
\texttt{L31.n13669} at any dose, a complementarity invisible in the aggregate
count).

\FloatBarrier
\section{{The Learned Perturbation as an Independent Primary}}
\label{app:perturbation}

\subsection{Objective and training}
\label{app:perturbation-objective}

\paragraph{Overview.}
We localise the verdict mechanism by learning a sparse perturbation $\delta$ of
the receptor coefficients $\alpha_{ck}$ {(\S\ref{sec:setup})} that, applied with
opposite signs, simultaneously flips and preserves the model's binary verdict
across a contrastive dataset. We describe the construction for a generic binary
{classification. In the main text} the classes are True/False, and the same design
instantiates other contrastive verdicts (e.g.\ harmful/harmless).

\section*{Contrastive Units (Bundles)}

The dataset is structured into distinct groups called \textbf{bundles}. Each bundle uses a $2 \times 2$ factorial design to test how the model {responds to two variables, the underlying class of the test query (e.g., true vs. fake) and the arbitrary labels} assigned to those classes in the prompt.

Every bundle starts with a matched pair of queries, $(q^+, q^-)$, which are identical except for their class. Instead of using standard labels like ``True'' and ``False,'' each bundle randomly draws a pair of neutral tokens {$(x, y)$, such as random letters or digits with similar baseline probabilities, to serve as the target answers.}

The bundle tests all four possible combinations of {these queries and label assignments, listed in the table below.}

\begin{table}[h]
\centering
\begin{tabular}{|l|l|l|}
\hline
\textbf{Label Assignment Rule ($\pi$)} & \textbf{Class $+$ Query ($q^+$)} & \textbf{Class $-$ Query ($q^-$)} \\
\hline
\textbf{$xy$} (Class $+$ maps to $x$) & Prompt: $\mathrm{ICL}^i_{xy}(q^+)$ & Prompt: $\mathrm{ICL}^i_{xy}(q^-)$ \\
& Expected Output: $x$ & Expected Output: $y$ \\
\hline
\textbf{$yx$} (Class $+$ maps to $y$) & Prompt: $\mathrm{ICL}^i_{yx}(q^+)$ & Prompt: $\mathrm{ICL}^i_{yx}(q^-)$ \\
& Expected Output: $y$ & Expected Output: $x$ \\
\hline
\end{tabular}
\end{table}

Structuring the data this way {guarantees two experimental properties.}
\begin{itemize}
\item \textbf{Label Invariance.} Because the answer tokens $(x, y)$ change from bundle to bundle, the learned perturbation $(\delta)$ cannot succeed by memorizing a fixed target word.
\item \textbf{Factor Decoupling.} Within a single bundle, every class yields both possible answer tokens depending on the context. This breaks any statistical correlation between the query's class, the assignment rule, and the final output token.
\end{itemize}

Consequently, a successful intervention must act on the combination of the query's actual class and the mapping rules established in the prompt. This forces the model into a ``semantically-unrelated-label regime,'' meaning it must actively read and apply the rules demonstrated in the context rather than relying on its pre-trained biases for standard labels.

\paragraph{Perturbation and asymmetric magnitudes.}

During the forward pass for a given prompt $p$, the intervention targets the receptor coefficients $\alpha_{ck}$, which correspond to the singular directions of the weight matrices as established by \cite{ahmad2025componentssingularvectorbasedinterpretability}. Specifically, these coefficients are perturbed according to the mapping $\alpha_{ck} \mapsto \alpha_{ck} + \eta(p,s)\delta_{ck}$. The scalar multiplier $\eta(p,s)$ is determined exclusively by the steering direction $s$ and the prompt's underlying class $q(p) \in \{+, -\}$.

The steering direction $s$ dictates the sign of the perturbation, whereas its magnitude is drawn from two learnable, class-specific parameters, $\eta_{+}$ and $\eta_{-}$ (where $\eta_{+}, \eta_{-} > 0$). These magnitudes are intentionally kept independent to account for the model's {asymmetric commitment, since a network may} not require a symmetric application of force to be pushed in opposing directions.

To structure the optimization, cells where the steering direction $s$ aligns with the ground truth are designated as the \emph{preserve} group, while the remaining cells constitute the \emph{flip} group. {Within any single class}, the preserve and flip cells share the identical magnitude parameter $\eta_{q(p)}$. Coupling the magnitudes in this way prevents the preserve magnitude from trivially collapsing to zero, which would otherwise happen given that the unperturbed model already predicts the correct answer on those instances.

\paragraph{Label-Space Scoring and Objective}

To evaluate the effect of the perturbation, we compute the probability of a given label string $\ell$. Letting $\tau(\ell)$ denote the set of tokenizer-variant IDs for $\ell$, this probability is the sum of the softmax probabilities {over the perturbed logits $\widetilde{L}(p)$, giving $P(\ell \mid p) = \sum_{v \in \tau(\ell)} \mathrm{softmax}(\widetilde{L}(p))_v$}.

We then isolate the binary classification probability over the two available label strings, $\ell_a$ and $\ell_b$, calculating $\widetilde{P}(t^\star \mid p, s) = \frac{P(t^\star \mid p)}{P(\ell_a \mid p) + P(\ell_b \mid p)}$. The target $t^\star$ represents the ground-truth label for cells in the \emph{preserve} group, and the complementary (flipped) label for cells in the \emph{flip} group.

The loss for an individual cell is the negative log-likelihood, $-\log \widetilde{P}(t^\star \mid p, s)$. Because each bundle $d_i$ generates four preserve and four flip prompts, the averaged loss terms {across the $N$ bundles are defined as}
$$ \mathcal{L}_{\text{flip}} = \frac{1}{N} \sum_i \frac{1}{4} \sum_{p \in d_i} -\log \widetilde{P}(t^\star \mid p, \text{flip}) $$
$$ \mathcal{L}_{\text{pres}} = \frac{1}{N} \sum_i \frac{1}{4} \sum_{p \in d_i} -\log \widetilde{P}(t^\star \mid p, \text{pres}) $$
.

{These two terms fulfill distinct theoretical roles.}
\begin{itemize}
    \item $\mathcal{L}_{\text{flip}}$ certifies \emph{sufficiency}, ensuring the intervention is strong enough to invert the model's verdict.
    \item $\mathcal{L}_{\text{pres}}$ certifies \emph{necessity}, ensuring that applying the steering vector with the opposite sign correctly maintains the ground-truth verdict.
\end{itemize}
Both terms will vanish only if the perturbation $\delta$ {functions as a directed control.}

{To enforce sparsity and localize} the intervention, the objective incorporates a Hard-Concrete $\ell_0$ relaxation \citep{louizos2018learningsparseneuralnetworks}. This approach gates each coordinate using a stochastic variable on $[0,1]$ with endpoint mass, yielding $\mathcal{L}_0(\delta)$, a differentiable expectation of the number of open gates.

With the language model's weights completely frozen, the perturbation $\delta$ and the magnitudes $\eta = (\eta_+, \eta_-)$ are optimized end-to-end using AdamW {to minimize the total objective}
$$ \mathcal{L}(\delta, \eta) = \mathcal{L}_{\text{flip}} + \mathcal{L}_{\text{pres}} + \lambda_0 \mathcal{L}_0(\delta) $$
.

Here, the hyperparameter $\lambda_0$ governs the trade-off between behavioral fidelity (achieving the desired flip/preserve outcomes) and localization (maintaining a sparse perturbation). \chg{It follows the fixed schedule given under the training recipe below.}

\paragraph{Training recipe.}
\chg{For every layer we take the SVD of each attention head's OV map and of the MLP output map and keep the directions with $\sigma_k>10^{-5}\sigma_1$ as receptors. The perturbation $\alpha=c\,(z\odot v)$ is added to every attention and MLP output at the final token only. $v$ is initialized at $0$, and $c$ is $\pm s_{\mathrm{fl}}$ or $\pm s_{\mathrm{lf}}$ by verdict side, each a learned softplus scale initialized at $1$. The gate $z$ is Hard-Concrete with $\log\alpha_0=2.5$, fixed temperature $2/3$, $\gamma=-0.1$ and $\zeta=1.1$, sampled during training and deterministic at evaluation. Training uses 12-shot in-context true/false judgements on known facts, with random single-letter labels from a pool of $13$ and no chat template, restricted to bundles the model answers correctly on all four quad prompts and split 70/10/20 with seed 42. The objective sums eight restricted two-way negative log-likelihoods per bundle (four prompts, each preserved and flipped), averages them over a batch of four bundles, and adds $\lambda(t)\,\mathbb{E}\lVert z\rVert_0$. The weight $\lambda$ is $0$ in epoch 1 and then ramps linearly from $0$ to $9\times10^{-4}$ over epochs 2 to 20, to $10^{-2}$ over epochs 21 to 25 and to $1$ over epochs 26 to 35, resetting to $0$ at each phase. We optimize with AdamW at learning rate $5\times10^{-3}$ for all intervention parameters, with no weight decay and a gradient-norm clip of $1.0$, for 35 epochs with seed $0$ and one run per model. The base model is frozen.}

\paragraph{Deployment.}
\chg{We use the epoch-31 checkpoint, where $\lambda$ is between $0.5$ and $0.6$, and apply $\delta=\pm2\,s_{\mathrm{side}}(z\odot v)$, twice the trained magnitude, a scale at which the perturbation flips the verdict in all four models. Evaluation is zero-shot on the True/False template over all statements ($n=532$, $583$, $478$ and $598$), which includes statements whose in-context bundles were in the perturbation's training split.}

\subsection{Results}
\label{app:perturbation-results}
{Tables~\ref{tab:perturb_flip_summary} and~\ref{tab:perturb_recruit_axes} report the verdict flips and the recruited detectors under the learned perturbation.}

\begin{table}[!htbp]\centering\small\setlength{\tabcolsep}{5pt}
\caption{\textbf{The trained receptor-space perturbation as the primary intervention ($\delta=\mathrm{scale}\cdot\eta_{\text{learned}}\cdot v$, scale$=2$, no dose sweep, epoch~31).} Per-class verdict $D_{TF}=\mathrm{mean}(\mathrm{logit}[\text{True}])-\mathrm{mean}(\mathrm{logit}[\text{False}])$ at the answer token, on \texttt{perfect\_on\_pairs}. A \emph{strict} verdict flip {requires BOTH sides to cross zero, so the true statement} must read False (suppress, $D_{TF}\!<\!0$) \emph{and} the fake statement must read True (fabricate, $D_{TF}\!>\!0$). {Only Qwen2.5-7B inverts both sides, Llama-3-8B and Gemma-2-9B flip the fabricate side only, and Mistral-7B-v0.3 crosses neither side.} ``recruited'' counts interpretable null/anti axes turned into positive contributors (Table~\ref{tab:perturb_recruit_axes}).}
\label{tab:perturb_flip_summary}
\begin{tabular}{l r r c r}
\toprule
 & true $D_{TF}$ & fake $D_{TF}$ & strict & interp.\ backups \\
model & clean$\to$sup & clean$\to$fab & flip? & recruited \\
\midrule
\texttt{Llama-3-8B} & $+3.60\!\to\!$\textcolor{FalseC}{$+2.40$} & $-4.17\!\to\!$\textcolor{TrueC}{$+0.08$} & $\times$ & 10 \\
\texttt{Gemma-2-9B} & $+4.28\!\to\!$\textcolor{FalseC}{$+0.41$} & $-5.22\!\to\!$\textcolor{TrueC}{$+1.67$} & $\times$ & 5 \\
\texttt{Qwen2.5-7B} & $+6.69\!\to\!$\textcolor{TrueC}{$-0.99$} & $-12.07\!\to\!$\textcolor{TrueC}{$+2.28$} & \checkmark & 6 \\
\texttt{Mistral-7B-v0.3} & $+4.60\!\to\!$\textcolor{FalseC}{$+4.23$} & $-6.99\!\to\!$\textcolor{FalseC}{$-0.65$} & $\times$ & 3 \\
\bottomrule\end{tabular}\end{table}

\begin{table*}[!htbp]\centering\small\setlength{\tabcolsep}{5.5pt}
\caption{\textbf{Interpretable backup detectors recruited from opposing/inert to contributing by the trained receptor-space perturbation} (scale$=2$, epoch~31, {\texttt{perfect\_on\_pairs}, with 10 / 5 / 6 / 3 axes} for Llama-3-8B/Gemma-2-9B/Qwen2.5-7B/Mistral-7B-v0.3, ranked by recruitment $R$). Each axis is \emph{null/anti} in the clean model and {turns a positive contributor to $D_{TF}$ under the primary intervention, shown separately on the two sides as $E_T$ (suppress side) and $E_F$ (fabricate side). $E^0$ is the same quantity with no primary, and $R=E^{\pi}-E^0$ is the recruitment.} Row shade $=$ clean class (\colorbox{FalseC!8}{anti}, \colorbox{black!6}{null}). {Superscripts are paired $t$ values, which are inflated because a fixed intervention makes each per-prompt shift near-deterministic, so recruitment is read from the $E^0\!\to\!E^{\pi}$ sign flip and $R$ rather than from $t$.}}
\label{tab:perturb_recruit_axes}
\resizebox{\textwidth}{!}{\begin{tabular}{l l l r r r r r r}
\toprule
 & & & \multicolumn{2}{c}{$E^0$ (clean)} & \multicolumn{2}{c}{$E^{\pi}$ (perturb)} & \multicolumn{2}{c}{$R=E^{\pi}\!-\!E^0$} \\
\cmidrule(lr){4-5}\cmidrule(lr){6-7}\cmidrule(lr){8-9}
axis & family & detects & $E^0_T$ & $E^0_F$ & $E^{\pi}_T$ & $E^{\pi}_F$ & $R_T$ & $R_F$ \\
\midrule
\rowcolor{white}\multicolumn{9}{l}{\textbf{Llama-3-8B}\ \small(clean $D_{TF}$: T$+3.60$/F$-4.17$ $\to$ prim T$+2.40$/F$+0.08$)} \\
\rowcolor{FalseC!8}
\texttt{L22.h1.svd117} & ov svd & true & \textcolor{FalseC}{$-0.004$} & \textcolor{FalseC}{$-0.006$} & \textcolor{TrueC}{$+0.111^{\scriptscriptstyle 88}$} & \textcolor{TrueC}{$+0.177^{\scriptscriptstyle 117}$} & $+0.115$ & $+0.183$ \\
\rowcolor{FalseC!8}
\texttt{L19.n3754} & mlp neuron & sure & \textcolor{FalseC}{$-0.008$} & \textcolor{FalseC}{$-0.022$} & \textcolor{TrueC}{$+0.018^{\scriptscriptstyle 16}$} & \textcolor{TrueC}{$+0.034^{\scriptscriptstyle 29}$} & $+0.027$ & $+0.056$ \\
\rowcolor{black!6}
\texttt{L22.h1.c60} & ov neuron & valid & \textcolor{NullC}{$+0.000$} & \textcolor{NullC}{$-0.001$} & \textcolor{TrueC}{$+0.035^{\scriptscriptstyle 30}$} & \textcolor{TrueC}{$+0.030^{\scriptscriptstyle 27}$} & $+0.035$ & $+0.031$ \\
\rowcolor{FalseC!8}
\texttt{L21.n4049} & mlp neuron & wrong & \textcolor{FalseC}{$-0.011$} & \textcolor{FalseC}{$-0.011$} & \textcolor{TrueC}{$+0.018^{\scriptscriptstyle 16}$} & \textcolor{TrueC}{$+0.018^{\scriptscriptstyle 16}$} & $+0.029$ & $+0.029$ \\
\rowcolor{black!6}
\texttt{L21.n798} & mlp neuron & true & \textcolor{NullC}{$-0.002$} & \textcolor{NullC}{$-0.002$} & \textcolor{TrueC}{$+0.021^{\scriptscriptstyle 18}$} & \textcolor{TrueC}{$+0.028^{\scriptscriptstyle 24}$} & $+0.023$ & $+0.030$ \\
\rowcolor{black!6}
\texttt{L28.h27.c120} & ov neuron & True & \textcolor{NullC}{$-0.002$} & \textcolor{NullC}{$-0.001$} & \textcolor{TrueC}{$+0.017^{\scriptscriptstyle 20}$} & \textcolor{TrueC}{$+0.030^{\scriptscriptstyle 33}$} & $+0.019$ & $+0.031$ \\
\rowcolor{black!6}
\texttt{L28.h27.c34} & ov neuron & True & \textcolor{NullC}{$+0.000$} & \textcolor{NullC}{$+0.001$} & \textcolor{TrueC}{$+0.015^{\scriptscriptstyle 16}$} & \textcolor{TrueC}{$+0.023^{\scriptscriptstyle 29}$} & $+0.015$ & $+0.023$ \\
\rowcolor{black!6}
\texttt{L28.h27.c0} & ov neuron & real & \textcolor{NullC}{$+0.000$} & \textcolor{NullC}{$-0.001$} & \textcolor{TrueC}{$+0.009^{\scriptscriptstyle 10}$} & \textcolor{TrueC}{$+0.015^{\scriptscriptstyle 18}$} & $+0.009$ & $+0.016$ \\
\rowcolor{black!6}
\texttt{L20.h11.c95} & ov neuron & false & \textcolor{NullC}{$+0.000$} & \textcolor{NullC}{$-0.002$} & \textcolor{TrueC}{$+0.009^{\scriptscriptstyle 8}$} & \textcolor{TrueC}{$+0.006^{\scriptscriptstyle 5}$} & $+0.009$ & $+0.008$ \\
\rowcolor{black!6}
\texttt{L22.h1.svd116} & ov svd & correct & \textcolor{NullC}{$-0.000$} & \textcolor{NullC}{$+0.001$} & \textcolor{TrueC}{$+0.006^{\scriptscriptstyle 6}$} & \textcolor{TrueC}{$+0.010^{\scriptscriptstyle 9}$} & $+0.006$ & $+0.009$ \\
\midrule
\rowcolor{white}\multicolumn{9}{l}{\textbf{Gemma-2-9B}\ \small(clean $D_{TF}$: T$+4.28$/F$-5.22$ $\to$ prim T$+0.41$/F$+1.67$)} \\
\rowcolor{FalseC!8}
\texttt{L38.n7088} & mlp neuron & wrong & \textcolor{FalseC}{$-1.420$} & \textcolor{FalseC}{$-0.715$} & \textcolor{TrueC}{$+0.014^{\scriptscriptstyle 11}$} & \textcolor{TrueC}{$+0.024^{\scriptscriptstyle 16}$} & $+1.434$ & $+0.738$ \\
\rowcolor{FalseC!8}
\texttt{L32.n8200} & mlp neuron & False & \textcolor{FalseC}{$-0.067$} & \textcolor{FalseC}{$-0.175$} & \textcolor{TrueC}{$+0.865^{\scriptscriptstyle 63}$} & \textcolor{TrueC}{$+0.105^{\scriptscriptstyle 50}$} & $+0.932$ & $+0.280$ \\
\rowcolor{FalseC!8}
\texttt{L39.n10210} & mlp neuron & incorrect & \textcolor{FalseC}{$-0.167$} & \textcolor{FalseC}{$-0.127$} & \textcolor{TrueC}{$+0.251^{\scriptscriptstyle 23}$} & \textcolor{TrueC}{$+0.216^{\scriptscriptstyle 23}$} & $+0.418$ & $+0.343$ \\
\rowcolor{FalseC!8}
\texttt{L34.n4784} & mlp neuron & truth & \textcolor{FalseC}{$-0.044$} & \textcolor{FalseC}{$-0.037$} & \textcolor{TrueC}{$+0.027^{\scriptscriptstyle 20}$} & \textcolor{TrueC}{$+0.017^{\scriptscriptstyle 16}$} & $+0.072$ & $+0.053$ \\
\rowcolor{FalseC!8}
\texttt{L40.h11.svd18} & ov svd & truth & \textcolor{FalseC}{$-0.027$} & \textcolor{FalseC}{$-0.044$} & \textcolor{TrueC}{$+0.030^{\scriptscriptstyle 25}$} & \textcolor{TrueC}{$+0.007^{\scriptscriptstyle 8}$} & $+0.057$ & $+0.051$ \\
\midrule
\rowcolor{white}\multicolumn{9}{l}{\textbf{Qwen2.5-7B}\ \small(clean $D_{TF}$: T$+6.69$/F$-12.07$ $\to$ prim T$-0.99$/F$+2.28$)} \\
\rowcolor{FalseC!8}
\texttt{L27.n13033} & mlp neuron & false & \textcolor{FalseC}{$-2.863$} & \textcolor{FalseC}{$-2.881$} & \textcolor{TrueC}{$+0.249^{\scriptscriptstyle 39}$} & \textcolor{TrueC}{$+0.265^{\scriptscriptstyle 39}$} & $+3.112$ & $+3.147$ \\
\rowcolor{FalseC!8}
\texttt{L22.n4538} & mlp neuron & false & \textcolor{FalseC}{$-0.035$} & \textcolor{FalseC}{$-0.026$} & \textcolor{TrueC}{$+0.201^{\scriptscriptstyle 75}$} & \textcolor{TrueC}{$+0.389^{\scriptscriptstyle 74}$} & $+0.236$ & $+0.415$ \\
\rowcolor{FalseC!8}
\texttt{L23.n8341} & mlp neuron & illusion & \textcolor{FalseC}{$-0.182$} & \textcolor{FalseC}{$-0.113$} & \textcolor{TrueC}{$+0.042^{\scriptscriptstyle 39}$} & \textcolor{TrueC}{$+0.046^{\scriptscriptstyle 27}$} & $+0.224$ & $+0.159$ \\
\rowcolor{FalseC!8}
\texttt{L25.n4929} & mlp neuron & false & \textcolor{FalseC}{$-0.249$} & \textcolor{FalseC}{$-0.071$} & \textcolor{TrueC}{$+0.009^{\scriptscriptstyle 7}$} & \textcolor{TrueC}{$+0.012^{\scriptscriptstyle 6}$} & $+0.258$ & $+0.083$ \\
\rowcolor{FalseC!8}
\texttt{L25.n8444} & mlp neuron & \cjkjp{假}{fake} & \textcolor{FalseC}{$-0.063$} & \textcolor{FalseC}{$-0.031$} & \textcolor{TrueC}{$+0.027^{\scriptscriptstyle 14}$} & \textcolor{TrueC}{$+0.040^{\scriptscriptstyle 14}$} & $+0.091$ & $+0.071$ \\
\rowcolor{FalseC!8}
\texttt{L22.n13025} & mlp neuron & \cjkjp{真正}{genuine} & \textcolor{FalseC}{$-0.017$} & \textcolor{FalseC}{$-0.045$} & \textcolor{TrueC}{$+0.040^{\scriptscriptstyle 36}$} & \textcolor{TrueC}{$+0.044^{\scriptscriptstyle 23}$} & $+0.056$ & $+0.089$ \\
\midrule
\rowcolor{white}\multicolumn{9}{l}{\textbf{Mistral-7B-v0.3}\ \small(clean $D_{TF}$: T$+4.60$/F$-6.99$ $\to$ prim T$+4.23$/F$-0.65$)} \\
\rowcolor{FalseC!8}
\texttt{L31.n13669} & mlp neuron & true & \textcolor{FalseC}{$-1.929$} & \textcolor{FalseC}{$-2.083$} & \textcolor{TrueC}{$+1.045^{\scriptscriptstyle 128}$} & \textcolor{TrueC}{$+1.238^{\scriptscriptstyle 126}$} & $+2.974$ & $+3.322$ \\
\rowcolor{FalseC!8}
\texttt{L20.n4786} & mlp neuron & right & \textcolor{FalseC}{$-0.013$} & \textcolor{FalseC}{$-0.003$} & \textcolor{TrueC}{$+0.008^{\scriptscriptstyle 8}$} & \textcolor{TrueC}{$+0.011^{\scriptscriptstyle 10}$} & $+0.021$ & $+0.014$ \\
\rowcolor{black!6}
\texttt{L16.n9991} & mlp neuron & false & \textcolor{NullC}{$+0.002$} & \textcolor{NullC}{$+0.001$} & \textcolor{TrueC}{$+0.005^{\scriptscriptstyle 5}$} & \textcolor{TrueC}{$+0.006^{\scriptscriptstyle 5}$} & $+0.004$ & $+0.005$ \\
\bottomrule
\end{tabular}}\end{table*}

\subsection{Three-route convergence}
\label{app:convergence}

Table~\ref{tab:cross_intervention_consistency} gives the per-axis, per-side
detail behind the convergence claim of {main text \S\ref{sec:census}. It shows which}
interpretable truth/falsity axes, null or opposing in the clean model, are
recruited into two-sided verdict defence under each of the three
mechanistically distinct primaries (the dosed swap of \S\ref{app:recruitment},
the learned perturbation of \S\ref{app:perturbation-objective}, and their
conjunction). Restricting to the three-primary comparison (excluding rows only
tested under the swap), $15$ of $31$ axes are recruited on both sides under
\emph{all three} primaries, and a further $12$ axes are recruited under the
swap alone (not separately tested under the other two primaries in this
table).

\begin{table*}[!htbp]\centering\small\setlength{\tabcolsep}{5pt}
\caption{\textbf{Recruitment is consistent across three distinct
interventions.} Each row is an interpretable truth/falsity axis that is null/anti in the clean
model ($E^0\!\le\!0$). The verdict sides are scored separately{,} \textbf{T} (true/suppress, $E_T$) and \textbf{F} (false/fabricate, $E_F$){,} and \textcolor{TrueC}{\checkmark} marks that
side becoming a contributor ($E^{\pi}\!>\!0$, $|t|\!\ge\!2$). \textbf{Swap} is an
in-distribution contrastive fact$\leftrightarrow$fake exchange, dose-swept
$\times1\!-\!\times4$ (\textcolor{TrueC}{\checkmark} $=$ rose monotonically and reliable by
{$\times4$). \textbf{Perturb} is an out-of-distribution trained steering vector applied at every
layer (scale $2$), and \textbf{Swap$+$Perturb} stacks both in one pass. The clean class is
\textcolor{FalseC}{anti} ($E^0\!<\!0$) or \textcolor{NullC}{null} ($E^0\!\approx\!0$), and
$\bigstar$ marks recruitment on both sides under all three.}}
\label{tab:cross_intervention_consistency}
\begin{tabular}{l l c @{\hskip 8pt} cc @{\hskip 8pt} cc @{\hskip 8pt} cc @{\hskip 6pt} c}
\toprule
 & & clean & \multicolumn{2}{c}{Swap} & \multicolumn{2}{c}{Perturb} & \multicolumn{2}{c}{Swap$+$Perturb} & \\
\cmidrule(lr){4-5}\cmidrule(lr){6-7}\cmidrule(lr){8-9}
axis & detects & $E^0$ & T & F & T & F & T & F & all \\
\midrule
\multicolumn{10}{l}{\textbf{Llama-3-8B}} \\
\rowcolor{TrueC!7}
\textbf{\texttt{L19.n3754}} & sure & \textcolor{FalseC}{\footnotesize anti} & \textcolor{TrueC}{\checkmark} & \textcolor{TrueC}{\checkmark} & \textcolor{TrueC}{\checkmark} & \textcolor{TrueC}{\checkmark} & \textcolor{TrueC}{\checkmark} & \textcolor{TrueC}{\checkmark} & \textcolor{TrueC}{$\bigstar$} \\
\rowcolor{TrueC!7}
\textbf{\texttt{L21.n4049}} & wrong & \textcolor{FalseC}{\footnotesize anti} & \textcolor{TrueC}{\checkmark} & \textcolor{TrueC}{\checkmark} & \textcolor{TrueC}{\checkmark} & \textcolor{TrueC}{\checkmark} & \textcolor{TrueC}{\checkmark} & \textcolor{TrueC}{\checkmark} & \textcolor{TrueC}{$\bigstar$} \\
\rowcolor{TrueC!7}
\textbf{\texttt{L22.h1.c60}} & valid & \textcolor{NullC}{\footnotesize null} & \textcolor{TrueC}{\checkmark} & \textcolor{TrueC}{\checkmark} & \textcolor{TrueC}{\checkmark} & \textcolor{TrueC}{\checkmark} & \textcolor{TrueC}{\checkmark} & \textcolor{TrueC}{\checkmark} & \textcolor{TrueC}{$\bigstar$} \\
\rowcolor{TrueC!7}
\textbf{\texttt{L22.h1.svd116}} & correct & \textcolor{NullC}{\footnotesize null} & \textcolor{TrueC}{\checkmark} & \textcolor{TrueC}{\checkmark} & \textcolor{TrueC}{\checkmark} & \textcolor{TrueC}{\checkmark} & \textcolor{TrueC}{\checkmark} & \textcolor{TrueC}{\checkmark} & \textcolor{TrueC}{$\bigstar$} \\
\rowcolor{TrueC!7}
\textbf{\texttt{L22.h1.svd117}} & true & \textcolor{FalseC}{\footnotesize anti} & \textcolor{TrueC}{\checkmark} & \textcolor{TrueC}{\checkmark} & \textcolor{TrueC}{\checkmark} & \textcolor{TrueC}{\checkmark} & \textcolor{TrueC}{\checkmark} & \textcolor{TrueC}{\checkmark} & \textcolor{TrueC}{$\bigstar$} \\
\rowcolor{TrueC!7}
\textbf{\texttt{L28.h27.c0}} & real & \textcolor{NullC}{\footnotesize null} & \textcolor{TrueC}{\checkmark} & \textcolor{TrueC}{\checkmark} & \textcolor{TrueC}{\checkmark} & \textcolor{TrueC}{\checkmark} & \textcolor{TrueC}{\checkmark} & \textcolor{TrueC}{\checkmark} & \textcolor{TrueC}{$\bigstar$} \\
\rowcolor{TrueC!7}
\textbf{\texttt{L28.h27.c34}} & True & \textcolor{NullC}{\footnotesize null} & \textcolor{TrueC}{\checkmark} & \textcolor{TrueC}{\checkmark} & \textcolor{TrueC}{\checkmark} & \textcolor{TrueC}{\checkmark} & \textcolor{TrueC}{\checkmark} & \textcolor{TrueC}{\checkmark} & \textcolor{TrueC}{$\bigstar$} \\
\texttt{L20.h11.c95} & false & \textcolor{NullC}{\footnotesize null} & \textcolor{black!22}{--} & \textcolor{TrueC}{\checkmark} & \textcolor{TrueC}{\checkmark} & \textcolor{TrueC}{\checkmark} & \textcolor{TrueC}{\checkmark} & \textcolor{TrueC}{\checkmark} &  \\
\texttt{L28.h27.c120} & True & \textcolor{NullC}{\footnotesize null} & \textcolor{TrueC}{\checkmark} & \textcolor{black!22}{--} & \textcolor{TrueC}{\checkmark} & \textcolor{TrueC}{\checkmark} & \textcolor{TrueC}{\checkmark} & \textcolor{TrueC}{\checkmark} &  \\
\texttt{L19.n2738} & illusion & \textcolor{FalseC}{\footnotesize anti} & \textcolor{TrueC}{\checkmark} & \textcolor{TrueC}{\checkmark} & \textcolor{black!22}{--} & \textcolor{black!22}{--} & \textcolor{TrueC}{\checkmark} & \textcolor{TrueC}{\checkmark} &  \\
\texttt{L21.n798} & true & \textcolor{NullC}{\footnotesize null} & \textcolor{black!22}{--} & \textcolor{black!22}{--} & \textcolor{TrueC}{\checkmark} & \textcolor{TrueC}{\checkmark} & \textcolor{TrueC}{\checkmark} & \textcolor{TrueC}{\checkmark} &  \\
\texttt{L21.n14165} & Genuine & \textcolor{NullC}{\footnotesize null} & \textcolor{TrueC}{\checkmark} & \textcolor{TrueC}{\checkmark} & \textcolor{black!22}{--} & \textcolor{black!22}{--} & \textcolor{black!22}{--} & \textcolor{black!22}{--} &  \\
\texttt{L23.n8972} & lie & \textcolor{FalseC}{\footnotesize anti} & \textcolor{TrueC}{\checkmark} & \textcolor{TrueC}{\checkmark} & \textcolor{black!22}{--} & \textcolor{black!22}{--} & \textcolor{black!22}{--} & \textcolor{black!22}{--} &  \\
\texttt{L23.n9811} & truth & \textcolor{FalseC}{\footnotesize anti} & \textcolor{TrueC}{\checkmark} & \textcolor{TrueC}{\checkmark} & \textcolor{black!22}{--} & \textcolor{black!22}{--} & \textcolor{black!22}{--} & \textcolor{black!22}{--} &  \\
\midrule
\multicolumn{10}{l}{\textbf{Gemma-2-9B}} \\
\rowcolor{TrueC!7}
\textbf{\texttt{L34.n4784}} & truth & \textcolor{FalseC}{\footnotesize anti} & \textcolor{TrueC}{\checkmark} & \textcolor{TrueC}{\checkmark} & \textcolor{TrueC}{\checkmark} & \textcolor{TrueC}{\checkmark} & \textcolor{TrueC}{\checkmark} & \textcolor{TrueC}{\checkmark} & \textcolor{TrueC}{$\bigstar$} \\
\rowcolor{TrueC!7}
\textbf{\texttt{L38.n7088}} & wrong & \textcolor{FalseC}{\footnotesize anti} & \textcolor{TrueC}{\checkmark} & \textcolor{TrueC}{\checkmark} & \textcolor{TrueC}{\checkmark} & \textcolor{TrueC}{\checkmark} & \textcolor{TrueC}{\checkmark} & \textcolor{TrueC}{\checkmark} & \textcolor{TrueC}{$\bigstar$} \\
\rowcolor{TrueC!7}
\textbf{\texttt{L39.n10210}} & incorrect & \textcolor{FalseC}{\footnotesize anti} & \textcolor{TrueC}{\checkmark} & \textcolor{TrueC}{\checkmark} & \textcolor{TrueC}{\checkmark} & \textcolor{TrueC}{\checkmark} & \textcolor{TrueC}{\checkmark} & \textcolor{TrueC}{\checkmark} & \textcolor{TrueC}{$\bigstar$} \\
\rowcolor{TrueC!7}
\textbf{\texttt{L40.h11.svd18}} & truth & \textcolor{FalseC}{\footnotesize anti} & \textcolor{TrueC}{\checkmark} & \textcolor{TrueC}{\checkmark} & \textcolor{TrueC}{\checkmark} & \textcolor{TrueC}{\checkmark} & \textcolor{TrueC}{\checkmark} & \textcolor{TrueC}{\checkmark} & \textcolor{TrueC}{$\bigstar$} \\
\texttt{L32.n8200} & False & \textcolor{FalseC}{\footnotesize anti} & \textcolor{TrueC}{\checkmark} & \textcolor{black!22}{--} & \textcolor{TrueC}{\checkmark} & \textcolor{TrueC}{\checkmark} & \textcolor{TrueC}{\checkmark} & \textcolor{TrueC}{\checkmark} &  \\
\texttt{L38.n854} & true & \textcolor{FalseC}{\footnotesize anti} & \textcolor{TrueC}{\checkmark} & \textcolor{TrueC}{\checkmark} & \textcolor{black!22}{--} & \textcolor{black!22}{--} & \textcolor{TrueC}{\checkmark} & \textcolor{TrueC}{\checkmark} &  \\
\texttt{L40.h7.c177} & genuine & \textcolor{FalseC}{\footnotesize anti} & \textcolor{TrueC}{\checkmark} & \textcolor{TrueC}{\checkmark} & \textcolor{black!22}{--} & \textcolor{black!22}{--} & \textcolor{TrueC}{\checkmark} & \textcolor{TrueC}{\checkmark} &  \\
\texttt{L41.n3789} & real & \textcolor{FalseC}{\footnotesize anti} & \textcolor{TrueC}{\checkmark} & \textcolor{TrueC}{\checkmark} & \textcolor{black!22}{--} & \textcolor{black!22}{--} & \textcolor{TrueC}{\checkmark} & \textcolor{TrueC}{\checkmark} &  \\
\texttt{L41.n8771} & True & \textcolor{FalseC}{\footnotesize anti} & \textcolor{TrueC}{\checkmark} & \textcolor{TrueC}{\checkmark} & \textcolor{black!22}{--} & \textcolor{black!22}{--} & \textcolor{TrueC}{\checkmark} & \textcolor{TrueC}{\checkmark} &  \\
\texttt{L33.n1466} & deception & \textcolor{FalseC}{\footnotesize anti} & \textcolor{TrueC}{\checkmark} & \textcolor{TrueC}{\checkmark} & \textcolor{black!22}{--} & \textcolor{black!22}{--} & \textcolor{black!22}{--} & \textcolor{black!22}{--} &  \\
\texttt{L39.n10085} & fake & \textcolor{FalseC}{\footnotesize anti} & \textcolor{TrueC}{\checkmark} & \textcolor{TrueC}{\checkmark} & \textcolor{black!22}{--} & \textcolor{black!22}{--} & \textcolor{black!22}{--} & \textcolor{black!22}{--} &  \\
\texttt{L40.h7.c105} & real & \textcolor{FalseC}{\footnotesize anti} & \textcolor{TrueC}{\checkmark} & \textcolor{TrueC}{\checkmark} & \textcolor{black!22}{--} & \textcolor{black!22}{--} & \textcolor{black!22}{--} & \textcolor{black!22}{--} &  \\
\midrule
\multicolumn{10}{l}{\textbf{Qwen2.5-7B}} \\
\rowcolor{TrueC!7}
\textbf{\texttt{L23.n8341}} & illusion & \textcolor{FalseC}{\footnotesize anti} & \textcolor{TrueC}{\checkmark} & \textcolor{TrueC}{\checkmark} & \textcolor{TrueC}{\checkmark} & \textcolor{TrueC}{\checkmark} & \textcolor{TrueC}{\checkmark} & \textcolor{TrueC}{\checkmark} & \textcolor{TrueC}{$\bigstar$} \\
\rowcolor{TrueC!7}
\textbf{\texttt{L27.n13033}} & false & \textcolor{FalseC}{\footnotesize anti} & \textcolor{TrueC}{\checkmark} & \textcolor{TrueC}{\checkmark} & \textcolor{TrueC}{\checkmark} & \textcolor{TrueC}{\checkmark} & \textcolor{TrueC}{\checkmark} & \textcolor{TrueC}{\checkmark} & \textcolor{TrueC}{$\bigstar$} \\
\texttt{L22.n13025} & \cjkjp{真正}{genuine} & \textcolor{FalseC}{\footnotesize anti} & \textcolor{black!22}{--} & \textcolor{black!22}{--} & \textcolor{TrueC}{\checkmark} & \textcolor{TrueC}{\checkmark} & \textcolor{TrueC}{\checkmark} & \textcolor{TrueC}{\checkmark} &  \\
\texttt{L22.n4538} & false & \textcolor{FalseC}{\footnotesize anti} & \textcolor{black!22}{--} & \textcolor{black!22}{--} & \textcolor{TrueC}{\checkmark} & \textcolor{TrueC}{\checkmark} & \textcolor{TrueC}{\checkmark} & \textcolor{TrueC}{\checkmark} &  \\
\texttt{L25.n4929} & false & \textcolor{FalseC}{\footnotesize anti} & \textcolor{TrueC}{\checkmark} & \textcolor{TrueC}{\checkmark} & \textcolor{TrueC}{\checkmark} & \textcolor{TrueC}{\checkmark} & \textcolor{black!22}{--} & \textcolor{black!22}{--} &  \\
\texttt{L25.n8444} & \cjkjp{假}{fake} & \textcolor{FalseC}{\footnotesize anti} & \textcolor{TrueC}{\checkmark} & \textcolor{TrueC}{\checkmark} & \textcolor{TrueC}{\checkmark} & \textcolor{TrueC}{\checkmark} & \textcolor{black!22}{--} & \textcolor{black!22}{--} &  \\
\texttt{L23.n14476} & \cjkjp{真正的}{genuine} & \textcolor{FalseC}{\footnotesize anti} & \textcolor{TrueC}{\checkmark} & \textcolor{TrueC}{\checkmark} & \textcolor{black!22}{--} & \textcolor{black!22}{--} & \textcolor{black!22}{--} & \textcolor{black!22}{--} &  \\
\texttt{L27.n4228} & truly & \textcolor{FalseC}{\footnotesize anti} & \textcolor{TrueC}{\checkmark} & \textcolor{TrueC}{\checkmark} & \textcolor{black!22}{--} & \textcolor{black!22}{--} & \textcolor{black!22}{--} & \textcolor{black!22}{--} &  \\
\midrule
\multicolumn{10}{l}{\textbf{Mistral-7B-v0.3}} \\
\rowcolor{TrueC!7}
\textbf{\texttt{L20.n4786}} & right & \textcolor{FalseC}{\footnotesize anti} & \textcolor{TrueC}{\checkmark} & \textcolor{TrueC}{\checkmark} & \textcolor{TrueC}{\checkmark} & \textcolor{TrueC}{\checkmark} & \textcolor{TrueC}{\checkmark} & \textcolor{TrueC}{\checkmark} & \textcolor{TrueC}{$\bigstar$} \\
\rowcolor{TrueC!7}
\textbf{\texttt{L31.n13669}} & true & \textcolor{FalseC}{\footnotesize anti} & \textcolor{TrueC}{\checkmark} & \textcolor{TrueC}{\checkmark} & \textcolor{TrueC}{\checkmark} & \textcolor{TrueC}{\checkmark} & \textcolor{TrueC}{\checkmark} & \textcolor{TrueC}{\checkmark} & \textcolor{TrueC}{$\bigstar$} \\
\texttt{L22.n1070} & accurate & \textcolor{FalseC}{\footnotesize anti} & \textcolor{TrueC}{\checkmark} & \textcolor{TrueC}{\checkmark} & \textcolor{black!22}{--} & \textcolor{TrueC}{\checkmark} & \textcolor{TrueC}{\checkmark} & \textcolor{TrueC}{\checkmark} &  \\
\texttt{L16.n9991} & false & \textcolor{NullC}{\footnotesize null} & \textcolor{black!22}{--} & \textcolor{black!22}{--} & \textcolor{TrueC}{\checkmark} & \textcolor{TrueC}{\checkmark} & \textcolor{TrueC}{\checkmark} & \textcolor{TrueC}{\checkmark} &  \\
\texttt{L22.h23.svd2} & wrong & \textcolor{FalseC}{\footnotesize anti} & \textcolor{TrueC}{\checkmark} & \textcolor{TrueC}{\checkmark} & \textcolor{black!22}{--} & \textcolor{black!22}{--} & \textcolor{TrueC}{\checkmark} & \textcolor{TrueC}{\checkmark} & \\
\texttt{L22.n8735} & false & \textcolor{FalseC}{\footnotesize anti} & \textcolor{TrueC}{\checkmark} & \textcolor{TrueC}{\checkmark} & \textcolor{black!22}{--} & \textcolor{black!22}{--} & \textcolor{black!22}{--} & \textcolor{black!22}{--} &  \\
\texttt{L28.n4079} & true & \textcolor{FalseC}{\footnotesize anti} & \textcolor{TrueC}{\checkmark} & \textcolor{TrueC}{\checkmark} & \textcolor{black!22}{--} & \textcolor{black!22}{--} & \textcolor{black!22}{--} & \textcolor{black!22}{--} &  \\
\texttt{L30.n11434} & false & \textcolor{FalseC}{\footnotesize anti} & \textcolor{TrueC}{\checkmark} & \textcolor{TrueC}{\checkmark} & \textcolor{black!22}{--} & \textcolor{black!22}{--} & \textcolor{black!22}{--} & \textcolor{black!22}{--} &  \\
\texttt{L31.n8773} & false & \textcolor{FalseC}{\footnotesize anti} & \textcolor{TrueC}{\checkmark} & \textcolor{TrueC}{\checkmark} & \textcolor{black!22}{--} & \textcolor{black!22}{--} & \textcolor{black!22}{--} & \textcolor{black!22}{--} &  \\
\bottomrule
\end{tabular}\end{table*}

\FloatBarrier
\section{{Pretraining-Native Recruitment in Base versus Instruct}}
\label{app:pretraining}

The identical dose-graded crossing of the bench is present in the base
(pretrained) checkpoint for each family, on the \emph{same} pairs each
checkpoint answers correctly on both sides zero-shot, so prompt, metric and
items are identical and the only difference is post-training.

\begin{figure}[!htbp]\centering
\includegraphics[width=\linewidth]{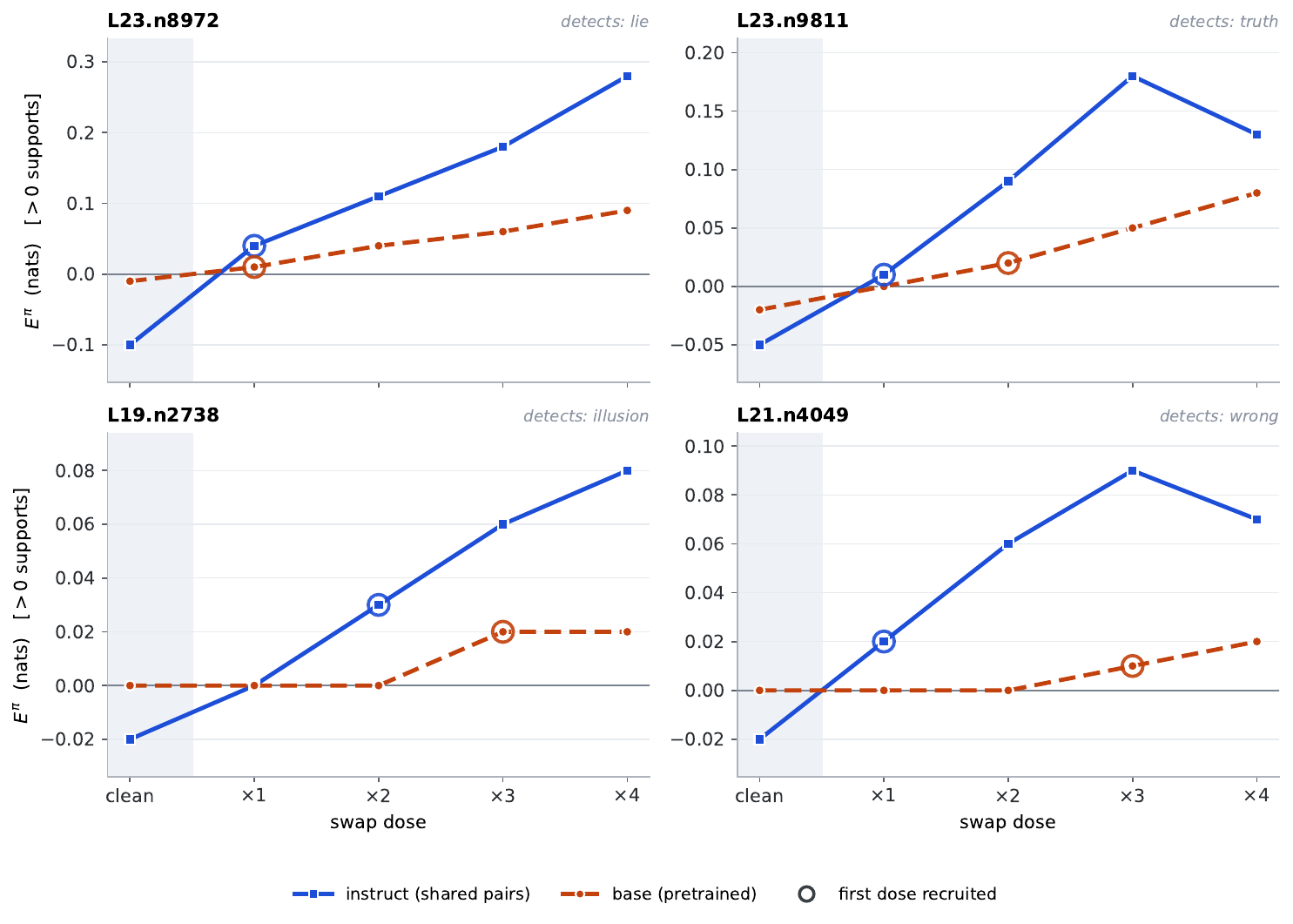}
\caption{\textbf{The same flagship backups are recruited in the pretrained
checkpoint} (Llama-3-8B base vs.\ Llama-3-8B-Instruct, contrastive swap of the
single verdict core \texttt{L18.n11065}, \texttt{sec\_ref=regime}, $n{=}240$).
{Each panel shows one flagship detector identified as a recruit in the instruct model,
plotting net erased margin $E^{\pi}=E^{\pi}_{T}+E^{\pi}_{F}$ against swap dose for
instruct (solid) and base (dashed). The ring marks} the first dose at which
that checkpoint's mean margin turns positive. \textbf{Every one of the four is
recruited in the base checkpoint as well as in instruct}, monotonically in
{dose and with the same ordering. \texttt{L23.n8972}} (`lie') reaches $+0.09$ in
base against $+0.28$ in instruct, \texttt{L23.n9811} (`truth') $+0.08$ against
$+0.13$, and the two weaker axes \texttt{L19.n2738} and \texttt{L21.n4049}
$+0.02$ against $+0.08$ and $+0.07$. The paired $t$ reaches $44$ (instruct)
and $24$ (base) {on \texttt{L23.n8972}, while base's whole verdict} is smaller to
begin with ($\kappa_{\mathrm{base}}{=}-2.0$ against $-8.4$). The clean-model
{classification does differ for \texttt{L19.n2738} and
\texttt{L21.n4049}, which oppose the verdict} outright in instruct but are only
weakly polarised in base, so instruction tuning sharpened them into opponents
without changing whether the ablation recruits them.}
\label{fig:llama_base_vs_instruct}\end{figure}

\begin{figure}[!htbp]\centering
\includegraphics[width=\linewidth]{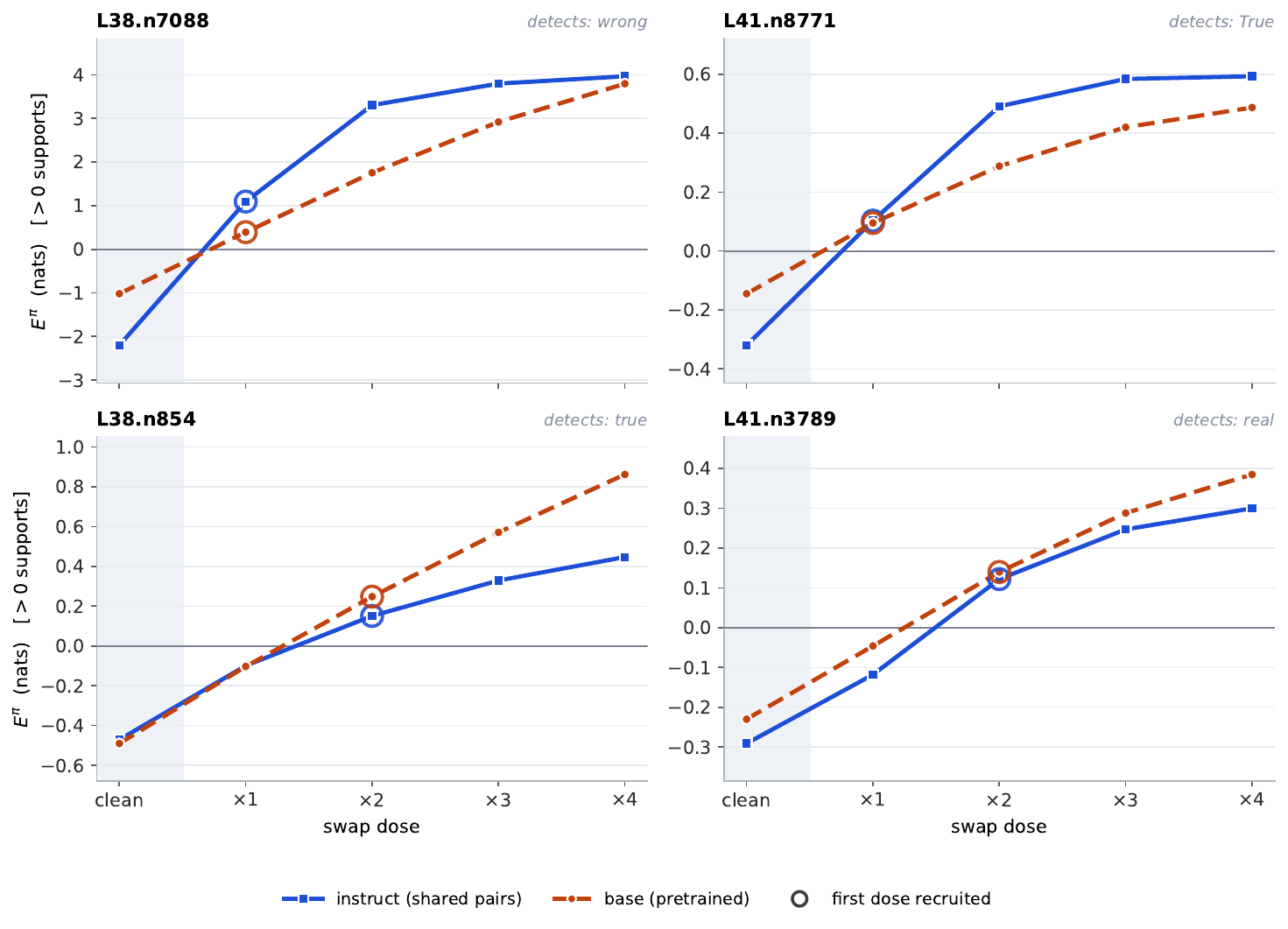}
\caption{\textbf{Backup recruitment is already present in the pretrained
checkpoint, in the same components} (Gemma-2-9B base vs.\ Gemma-2-9b-it, joint
contrastive swap of the verdict cores \texttt{L28.n2046}$+$\texttt{L33.n4294},
\texttt{sec\_ref=regime}, $n{=}535$). {Each panel shows one flagship recruited
detector, plotting net erased margin} $E^{\pi}=E^{\pi}_{T}+E^{\pi}_{F}$ against swap
dose for instruct (solid) and base (dashed). All four oppose the verdict in
\emph{both} clean models and are recruited in \emph{both}, at the same dose
in three of four panels. \texttt{L38.n7088} (`wrong') moves
$-1.02\!\rightarrow\!+3.79$ in base against $-2.20\!\rightarrow\!+3.97$ in instruct {(the curves nearly coincide at $\times4$),} and on
\texttt{L38.n854} and \texttt{L41.n3789} the base curve finishes \emph{above}
instruct. The paired $t$ reaches $120$ in base and $262$ in instruct. Across
the full bench, \textbf{all $11$ detectors that oppose the verdict in both
clean models are recruited in both checkpoints} by $\times4$. Base
magnitudes are comparable despite its verdict being far smaller
{($\kappa_{\mathrm{base}}{=}-2.83$ against $-9.81$). Normalised by the margin
the ablation destroys, the base backups restore $11$ to $13\%$ of the erased
verdict against instruct's $6$ to $9\%$.}}
\label{fig:gemma_base_vs_instruct}\end{figure}

\begin{figure}[!htbp]\centering
\includegraphics[width=\linewidth]{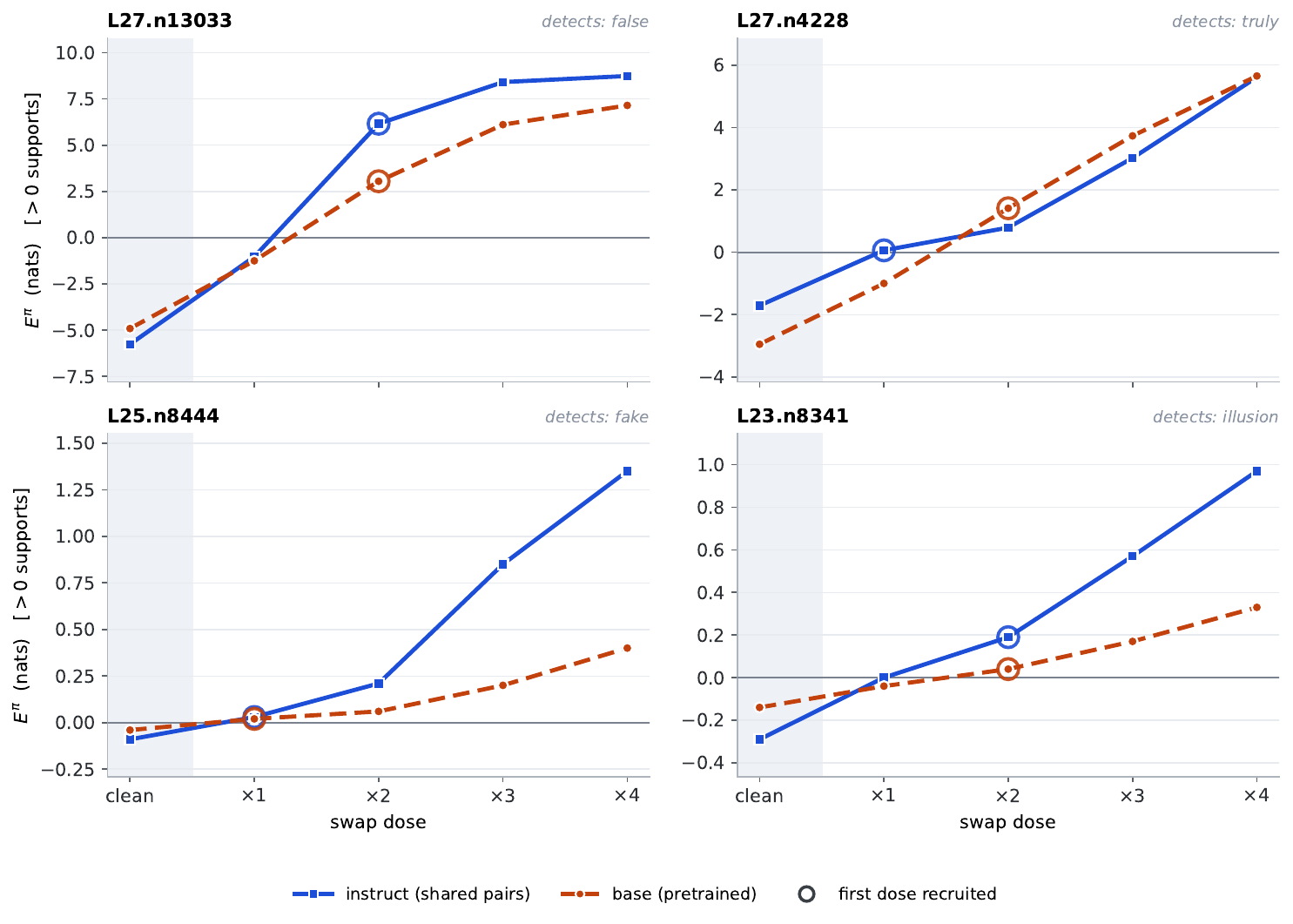}
\caption{\textbf{Backup recruitment is already present in the pretrained
checkpoint, in the same components} (Qwen2.5-7B base vs.\ Qwen2.5-7B-Instruct,
joint contrastive swap of the verdict cores
\texttt{L22.n13149}$+$\texttt{L24.n14758}, \texttt{sec\_ref=regime}, $n{=}470$).
One panel per flagship recruited detector, showing its net erased margin
$E^{\pi}=E^{\pi}_{T}+E^{\pi}_{F}$ against swap dose for the instruct (solid)
and base (dashed) checkpoint. Every detector opposes the verdict in \emph{both}
clean models and is recruited in \emph{both} under ablation, at the same dose
in three of four panels. \texttt{L27.n13033} moves
$-4.91\!\rightarrow\!+7.15$ in base against $-5.76\!\rightarrow\!+8.74$ in
instruct, and \texttt{L27.n4228} $-2.95\!\rightarrow\!+5.65$ against
{$-1.71\!\rightarrow\!+5.61$. In that panel} the base curve finishes
\emph{above} instruct. The recruitment ordering is identical in the two
checkpoints (\texttt{L27.n13033} $>$ \texttt{L27.n4228} $>$ \texttt{L25.n8444}
$>$ \texttt{L23.n8341}), including the multilingual \texttt{L25.n8444} (top
suppressed token \cjkjp{假}{fake}). Absolute magnitudes are smaller in base
{only because its whole verdict is smaller ($\kappa_{\mathrm{base}}{=}-6.16$
against $-18.87$).} Normalised by the margin the ablation destroys, the base
{backups restore $22$ to $27\%$ of the erased verdict against instruct's
$9$ to $10\%$.}}
\label{fig:qwen_base_vs_instruct}\end{figure}

\begin{figure}[!htbp]\centering
\includegraphics[width=\linewidth]{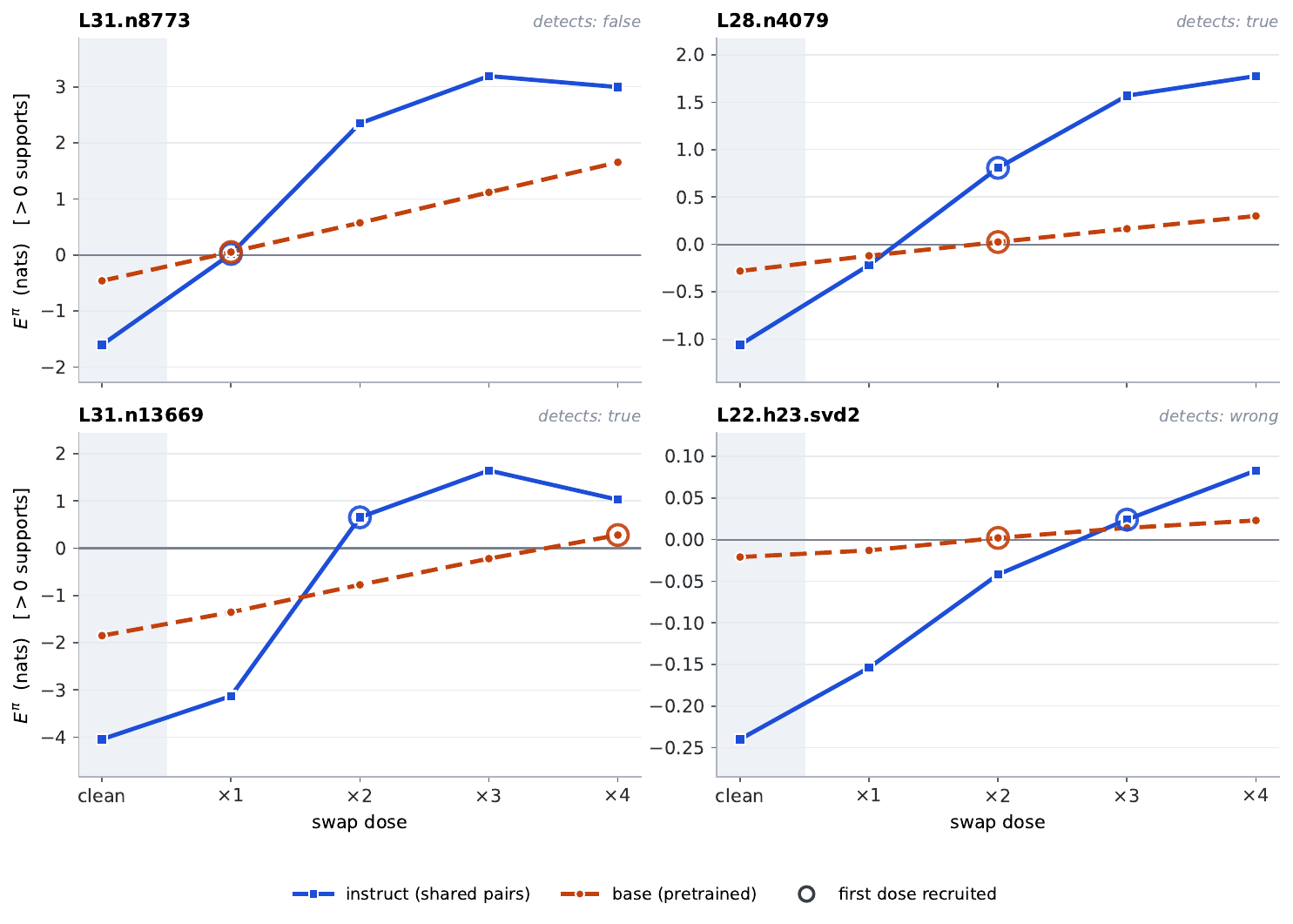}
\caption{\textbf{Backup recruitment is already present in the pretrained
checkpoint, in the same components} (Mistral-7B-v0.3 base vs.\
Mistral-7B-Instruct-v0.3, joint contrastive swap of the verdict cores
\texttt{L20.n14286}$+$\texttt{L19.n8228}, \texttt{sec\_ref=regime}, $n{=}423$).
All four detectors oppose the verdict in \emph{both} clean models and are
{recruited in \emph{both}. \texttt{L31.n8773}} reaches $+1.65$ in base against
$+3.00$ in instruct, \texttt{L28.n4079} $+0.30$ against $+1.78$,
\texttt{L31.n13669} $+0.28$ against $+1.03$, and the attention-head axis
\texttt{L22.h23.svd2} {$+0.02$ against $+0.08$, which extends the same recruitment one
family beyond MLP neurons.} Base magnitudes are smaller only because its whole
{verdict is smaller ($\kappa_{\mathrm{base}}{=}-1.80$ against $-12.40$, a
$7\times$ scale gap).} Normalised by the margin the ablation destroys, the base
backups restore {a near-constant $24$ to $25\%$ of the erased verdict at every
dose, against instruct's $8$ to $11\%$. By $\times4$} all six shared clean-anti
detectors are recruited in both checkpoints.}
\label{fig:mistral_base_vs_instruct}\end{figure}

Together, the four figures make backup recruitment a property of the
pretrained network that instruction tuning inherits (and, in absolute terms, amplifies for Qwen and Mistral) rather than one instruction tuning creates {(Figures~\ref{fig:llama_base_vs_instruct} to \ref{fig:mistral_base_vs_instruct})}.

\FloatBarrier
\section{Recruitment Is Not Renormalization}
\label{app:renorm}

\subsection{The objection}
\label{app:renorm-objection}

Recruitment compares a secondary's contribution {across two causal worlds.
The score $R_r = E^{\pi}_r - E^{0}_r$} (main text Eq.~\ref{eq:recruitment}) reads the
same direction once in the clean model and once in the {dosed} one. The two
worlds differ in one respect that affects every direction at once. A
component's write reaches the logits through the final normalization, which
the readout of main text \S\ref{sec:setup} folds into $W_U$ as a per-input
gain proportional to $1/\rho_b(x)$, where
\begin{equation}
\rho_b(x) \;=\; \operatorname{RMS}\!\big(h^{L}(x;b)\big)
\;=\; \Big(\tfrac{1}{d}\textstyle\sum_{i} h^{L}_i(x;b)^2 + \varepsilon\Big)^{1/2}
\label{eq:rho}
\end{equation}
is the RMS of the answer-position stream in world $b$.\footnote{Absolute
$\rho$ is family-idiosyncratic (Gemma-2's late-stream norms are an order of
magnitude larger than Qwen's). Every inference below uses within-family
ratios, which are invariant to any fixed rescaling of the measurement hook.}
If the primary shrinks this denominator, as one would expect of an
intervention that removes a large write, then every surviving direction's
path to the logits is multiplied by
\begin{equation}
g_k \;=\; \rho_{\varnothing} \,/\, \rho_{\pi_k},
\label{eq:gain}
\end{equation}
and if $g_k$ grows with dose, so does every direction's measured effect, with
no change in how the model uses any of them. On this reading,
\S\ref{app:recruitment} documents a change of measurement frame rather than a
change of computation. {The concern has precedent, since a share} of the head-level
self-repair observed under ablation has been attributed to exactly this
mediator \citep{rushing2024explorationsselfrepairlanguagemodels}, and $\rho$
is the only multiplicative factor between a weight-fixed write and the margin
on the direct path.

\subsection{What the objection predicts}
\label{app:renorm-predicts}

In its strict form, the null says that the two worlds differ only in the
denominator. Every logit then carries the same factor, and because $D_{XY}$ is
a difference of logits (main text Eq.~\ref{eq:margin}), {so does the margin,
$D_{XY}(x;\pi_k) = g_k D_{XY}(x;\varnothing)$}. For any secondary, therefore,
\begin{equation}
E^{\pi}_r = g_k E^{0}_r,
\qquad
R_r = (g_k-1)\, E^{0}_r .
\label{eq:null_prediction}
\end{equation}
Three testable consequences follow.

\textbf{(P1) No direction can change sign.} Since $g_k>0$ for any nonzero
stream, $E^{\pi}_r$ and $E^{0}_r$ share a sign. A dismissed direction
($E^{0}_r\le 0$) may shrink toward zero or become more negative, but it cannot
become a two-sided contributor. Certification
(\S\ref{app:census}) requires $E>0$ on both classes, each
reliably, and is therefore unreachable under the null.

\textbf{(P2) Amplification predicts the wrong sign of $R$.} On the dismissed
bench $E^{0}_r\le 0$, so a gain above unity {gives $R_r=(g_k-1)E^{0}_r\le 0$, and
amplification} \emph{deepens} opposition. Only attenuation ($g_k<1$) can
produce $R_r>0$ on these directions, and then only by shrinking opposition
toward zero, never by crossing it.

\textbf{(P3) The gain-attributable share of $R$ is bounded by
$|1-g_k|\,|E^{0}_r|$.} This bound holds under the {mixed account as well.
Whatever genuine} recruitment occurs, Eq.~\ref{eq:null_prediction} fixes the
most that renormalization alone can contribute, per direction, class, and
dose, from measured quantities alone.

\subsection{Measurement}
\label{app:renorm-measurement}

We record $\rho$ at the answer position, in the clean model and inside each
$\pi_k$, per statement class, on $n=50$ evaluation pairs per family, using the
primaries of \S\ref{app:recruitment} {(the single core
\texttt{L18.n11065} for Llama, the joint \texttt{L28.n2046}+\texttt{L33.n4294} for Gemma, and the
joint \texttt{L22.n13149}+\texttt{L24.n14758} for Qwen)}. Since $\rho(x;\pi_k)$ is a
property of the world rather than of any secondary, the bounds below {apply to
the entire bench, including units not tabulated.} Per-condition dispersion is
small (s.e.m.\ $\le 0.9\%$ of the mean throughout), and the subsample is
{faithful to the full runs. The class-resolved} margins of
Table~\ref{tab:margin_worlds} cross zero at the same doses as at full $n$, and
the emblematic contributions they encode reproduce the grids of
\S\ref{app:recruitment-tables} to within $0.03$ nats.\footnote{Gemma
{\texttt{L38.n7088} at $\times1$ gives $E=(+0.473,+0.615)$ here against
$(+0.474,+0.613)$ at $n{=}583$, Qwen \texttt{L27.n13033} at $\times2$ gives
$(+2.95,+3.25)$ against $(+2.93,+3.22)$, and Llama \texttt{L23.n8972} at
$\times1$ gives $(+0.024,+0.015)$} against $(+0.02,+0.02)$.}

\begin{table}[!htbp]
\centering
\caption{\textbf{The final-norm gain across dose.} $\rho_{\varnothing}$ is the clean answer-position RMS of the {pre-unembedding stream, and $g_k=\rho_{\varnothing}/\rho_{\pi_k}$} is the factor the final normalization applies to every fixed write's direct path in the dosed world. Values above unity amplify, values below unity attenuate. The artifact account requires $g_k>1$ \emph{and increasing in $k$}. {In the data, the gain \emph{decreases} with dose on true statements in both joint-core families, is elevated but dose-flat on Gemma's fake side, and at the on-manifold dose $\times1$ exceeds $1.03$ in no other family-class cell.}}
\label{tab:gain}
\vspace{2pt}
\small
\begin{tabular}{llcccccc}
\toprule
Model & Class & $\rho_{\varnothing}$ & $g_{\times1}$ & $g_{\times2}$ & $g_{\times3}$ & $g_{\times4}$ & profile in $k$ \\
\midrule
Llama-3-8B-Instruct & T & 0.990 & 1.023 & 1.060 & 1.105 & 1.157 & increasing \\
                    & F & 0.930 & 0.992 & 1.000 & 1.010 & 1.032 & flat/slight \\
\addlinespace
Gemma-2-9b-it       & T & 28.99 & 0.997 & 0.957 & 0.934 & 0.926 & decreasing \\
                    & F & 31.07 & 1.080 & 1.103 & 1.089 & 1.086 & flat \\
\addlinespace
Qwen2.5-7B-Instruct & T & 6.09  & 0.996 & 0.978 & 0.941 & 0.890 & decreasing \\
                    & F & 5.92  & 1.024 & 1.033 & 1.017 & 0.985 & flat \\
\bottomrule
\end{tabular}
\end{table}

\subsection{Class-resolved margins in the audited worlds}
\label{app:renorm-worlds}

Table~\ref{tab:margin_worlds} reports the class-resolved verdict margins
across dose in the audit's subsample. Three features of it matter for what
follows.

{First, the margins fix the pre-/post-failure partition.} At $\times1$ every
family still answers correctly on both classes, and {every family-class pair
crosses zero between $\times1$ and $\times2$, so the on-manifold} dose is
pre-failure everywhere.

{Second, the primary's dose-response} is smooth and family-shaped. On true
statements, Llama's single-core margin falls by a near-constant {$2.1$ to $2.7$
nats per dose, whereas both joint-core families decelerate sharply at depth.
The per-dose drop} shrinks from $2.85$ to $0.54$ nats in Gemma and from $5.26$
to $2.17$ in Qwen, and the fake sides mirror the pattern after an initial
rise. {Compensatory recruitment predicts a marginal effect that flattens as the
intervention deepens, and Llama, whose reserve is by far the
smallest of the three (\S\ref{app:pretraining}), is the family whose}
steps do not flatten. The deceleration coincides with the true-side norm
growth in Table~\ref{tab:gain}, for the reason given in
\S\ref{app:renorm-expansion}. Gemma's final logit soft-capping cannot account
for it on its own, since Qwen, which has none, decelerates as strongly.

Third, the verdict's redundancy is class-asymmetric. Qwen's fake-side margin
($-11.86$ clean) withstands $\times1$ at $-7.16$ {and barely crosses at}
$\times2$ ($+0.38$), while its true side falls five nats in a single dose.

\begin{table}[!htbp]
\centering
\caption{\textbf{Class-resolved verdict margins across dose} {($n{=}50$ pairs per family, same runs as Table~\ref{tab:gain}). $D$ is the class-fixed margin $D_{TF}$ of main text \S\ref{sec:setup}, with correct sign $+$ on true statements and $-$ on fakes. $D_{\pi}$ is the margin in the dosed world $\pi_k$, and $D_{\pi,\mathrm{sec}}$ additionally} overrides the emblematic anti-detector with its regime reference (Llama \texttt{L23.n8972}, Gemma \texttt{L38.n7088}, Qwen \texttt{L27.n13033}), so each column pair regenerates $E^{\pi}$ via $E_T = D_{\pi} - D_{\pi,\mathrm{sec}}$ and $E_F = D_{\pi,\mathrm{sec}} - D_{\pi}$.}
\label{tab:margin_worlds}
\vspace{2pt}
\small
\setlength{\tabcolsep}{4pt}
\begin{tabular}{ll r rr rr rr rr}
\toprule
& & & \multicolumn{2}{c}{$\times1$} & \multicolumn{2}{c}{$\times2$} & \multicolumn{2}{c}{$\times3$} & \multicolumn{2}{c}{$\times4$}\\
\cmidrule(lr){4-5}\cmidrule(lr){6-7}\cmidrule(lr){8-9}\cmidrule(lr){10-11}
Model & Cls & $D_{\varnothing}$ & $D_{\pi}$ & $D_{\pi,\mathrm{sec}}$ & $D_{\pi}$ & $D_{\pi,\mathrm{sec}}$ & $D_{\pi}$ & $D_{\pi,\mathrm{sec}}$ & $D_{\pi}$ & $D_{\pi,\mathrm{sec}}$\\
\midrule
Llama-3-8B & T & $+3.51$ & $+1.44$ & $+1.42$ & $-0.83$ & $-0.89$ & $-3.31$ & $-3.41$ & $-6.00$ & $-6.17$\\
           & F & $-4.13$ & $-2.08$ & $-2.07$ & $+0.70$ & $+0.74$ & $+3.71$ & $+3.77$ & $+6.60$ & $+6.69$\\
\addlinespace
Gemma-2-9b & T & $+4.40$ & $+1.55$ & $+1.08$ & $-0.83$ & $-1.90$ & $-1.92$ & $-3.12$ & $-2.46$ & $-3.68$\\
           & F & $-5.21$ & $-2.80$ & $-2.19$ & $+1.73$ & $+3.95$ & $+3.93$ & $+6.53$ & $+4.90$ & $+7.65$\\
\addlinespace
Qwen2.5-7B & T & $+6.39$ & $+1.12$ & $+1.50$ & $-3.95$ & $-6.89$ & $-7.51$ & $-11.50$ & $-9.68$ & $-13.82$\\
           & F & $-11.86$ & $-7.16$ & $-7.54$ & $+0.38$ & $+3.62$ & $+7.65$ & $+12.14$ & $+13.11$ & $+17.80$\\
\bottomrule
\end{tabular}
\end{table}

\subsection{The refutation}
\label{app:renorm-refutation}

{The artifact account requires a gain above unity that also strengthens with
dose.} Table~\ref{tab:gain} shows that this conjunction is
absent in every joint-core cell. The one cell in which the gain does rise
with dose, Llama's true side, is addressed under (P2) and (P3) below.

\paragraph{Sign crossings are outside the null (P1).}
The central phenomenon of \S\ref{app:recruitment} {is that dismissed
directions \emph{change sign}, over and above any growth in magnitude.} Llama's
\texttt{L23.n8972} moves from $(-0.04,-0.06)$ to $(+0.17,+0.10)$, Gemma's
\texttt{L38.n7088} from $(-1.42,-0.72)$ to $(+1.22,+2.73)$, and Qwen's
\texttt{L27.n13033} from $(-2.86,-2.88)$ to $(+4.08,+4.65)$. No positive
scalar produces a sign change.

\paragraph{Where the effects are largest, the gain moves the wrong way (P2).}
On true statements in both joint-core families the stream norm \emph{grows}
with dose, driving $g$ down to $0.926$ (Gemma) and $0.890$ (Qwen). Every
fixed write's direct path is attenuated as recruitment proceeds, so the
reported $E^{\pi}_T$ \emph{understate} their fixed-gain counterparts, by
roughly $11\%$ for Qwen's strongest recruit at $\times4$. On the fake sides of
Gemma and Qwen, and on Llama's true side, the gain is instead mildly
{amplifying, and there Eq.~\ref{eq:null_prediction}} predicts $R_r\le 0$ for every
dismissed direction, the opposite of what is observed in every such cell.

\paragraph{Where the gain is elevated, it is flat in dose.}
Gemma's fake side carries the largest amplification in the study,
$g_F\in[1.080,1.103]$, and it is constant across the sweep, while
$E^{\pi}_F$ for \texttt{L38.n7088} rises from $+0.62$ to $+2.73$ nats, a
factor of $4.5$. A dose-independent multiplier cannot generate a dose-graded
interaction. Dividing the $\times1$ value by the full gain leaves
$+0.613/1.080\approx+0.57$ nats, still the largest pre-failure effect on the
bench.

\paragraph{The gain-attributable share of $R$ is at most a few percent (P3).}
Table~\ref{tab:bound} applies the bound of Eq.~\ref{eq:null_prediction} to
the three emblematic recruits. In half the cells the predicted contribution
{has the wrong sign, and where it has the right sign it accounts for at most}
$4.5\%$ of the observed recruitment. At the on-manifold dose $\times1$ the
largest correction anywhere is $0.6\%$ of $R$, and in Llama the implied
inflation of $E^{\pi}$ is about $0.0005$ nats against a per-cell standard
error of about $0.0016$. Applying the correction unit by unit changes no
{certification, no sign, and no ordering. Llama's} dose trajectory, divided
through by $g$, remains strictly monotone ($0.021\to0.060\to0.104\to0.143$).

\begin{table}[!htbp]
\centering
\caption{\textbf{The maximum recruitment attributable to renormalization.} For each emblematic recruit, $E^{0}$ is the clean contribution, $R=E^{\pi}-E^{0}$ the observed recruitment, and $R_{\mathrm{gain}}=(g_k-1)E^{0}$ the recruitment predicted by the strict gain null of Eq.~\ref{eq:null_prediction}. ``Wrong sign'' marks cells in which the gain predicts recruitment opposite in direction to what is observed.}
\label{tab:bound}
\vspace{2pt}
\small
\setlength{\tabcolsep}{5pt}
\begin{tabular}{llrrrrl}
\toprule
Unit (detects) & Cls & $E^{0}$ & dose & $R$ & $R_{\mathrm{gain}}$ & share of $R$ \\
\midrule
\texttt{L23.n8972} (\emph{lie}) & T & $-0.04$ & $\times1$ & $+0.06$ & $-0.001$ & wrong sign \\
Llama-3-8B                      & T & $-0.04$ & $\times4$ & $+0.21$ & $-0.006$ & wrong sign \\
                                & F & $-0.06$ & $\times1$ & $+0.08$ & $+0.000$ & $0.6\%$ \\
                                & F & $-0.06$ & $\times4$ & $+0.16$ & $-0.002$ & wrong sign \\
\addlinespace
\texttt{L38.n7088} (\emph{wrong}) & T & $-1.42$ & $\times1$ & $+1.89$ & $+0.004$ & $0.2\%$ \\
Gemma-2-9b                        & T & $-1.42$ & $\times4$ & $+2.64$ & $+0.105$ & $4.0\%$ \\
                                  & F & $-0.72$ & $\times1$ & $+1.33$ & $-0.057$ & wrong sign \\
                                  & F & $-0.72$ & $\times4$ & $+3.44$ & $-0.061$ & wrong sign \\
\addlinespace
\texttt{L27.n13033} (\emph{false}) & T & $-2.86$ & $\times2$ & $+5.79$ & $+0.063$ & $1.1\%$ \\
Qwen2.5-7B                         & T & $-2.86$ & $\times4$ & $+6.94$ & $+0.315$ & $4.5\%$ \\
                                   & F & $-2.88$ & $\times2$ & $+6.10$ & $-0.095$ & wrong sign \\
                                   & F & $-2.88$ & $\times4$ & $+7.53$ & $+0.043$ & $0.6\%$ \\
\bottomrule
\end{tabular}
\end{table}

The mediator also leaves the estimand's internal comparison untouched. The
secondary's own effect on the denominator,
$\rho(x;\pi_k,\mathrm{sec}_r)/\rho(x;\pi_k)$, departs from unity by at most
$0.28\%$, $0.94\%$, and $0.50\%$ (Llama, Gemma, Qwen) at any dose, so the two
forward passes differenced inside every $E^{b}_r$ (main text
Eq.~\ref{eq:contribution}) are gain-matched to under $1\%$. {Only the
cross-world comparison $R_r$ was exposed to the gain, and
Tables~\ref{tab:gain} and~\ref{tab:bound} bound that comparison.}

\paragraph{{Information in the margin levels.}}
The paired columns of Table~\ref{tab:margin_worlds} regenerate $E^{\pi}$, but
the margin levels carry information of their own. At $\times1$, where the
verdict still stands in every family, overriding the emblematic recruit costs
Gemma $31\%$ of its surviving true-side margin ($1.55\to1.08$) and $22\%$ of
its fake-side margin {($-2.80\to-2.19$). Measured} against what the primary
erased, that single unit is restoring $17\%$ and $25\%$ of it. The same
operation costs Llama $1.7\%$ ($1.44\to1.42$), the family asymmetry in
reserve size seen again at the level of margins. In Qwen at $\times1$ the
override moves the margin {the \emph{other} way ($1.12\to1.50$).
\texttt{L27.n13033} is} still depressing the verdict there, and it crosses at
$\times2$, so the opponent-to-recruit transition is legible without
differencing. Past the crossing, $D_{\pi,\mathrm{sec}}$ sits farther from
zero than $D_{\pi}$ on the inverted side {(for Gemma at $\times2$, $-0.83\to-1.90$ on
$T$ and $+1.73\to+3.95$ on $F$, or $20\%$ and $32\%$ of the
erased-and-inverted signal), so the recruit} is partially reconstituting the
erased class signal.

\subsection{Why an inversion-type primary need not shrink the stream}
\label{app:renorm-expansion}

The expectation that suppressing an important component shrinks the residual
norm is an expectation about deletion. The dosed swap of main text
{Eq.~\ref{eq:swap} deletes nothing. It replaces} the core's coefficient with a
value drawn from the opposite class and, at $k>1$, injects a write
\emph{larger} in magnitude than the one it displaces. Writing
$\Delta a \propto k$ for the induced coefficient shift,
\begin{equation}
\big\lVert h + \Delta a\,\omega_{\star}\big\rVert^{2}
= \lVert h\rVert^{2}
+ 2\,\Delta a\,\langle h,\omega_{\star}\rangle
+ \Delta a^{2}\lVert \omega_{\star}\rVert^{2}.
\label{eq:normexpansion}
\end{equation}
The linear term is negative, because the injected signal opposes the stream's
prevailing class direction, while the quadratic term {grows as $k^{2}$. The
recruits'} own writes then feed back into $\rho$ in the same direction. Beyond
a crossover dose the norm must therefore rise. The joint-core primaries, which
inject two enlarged writes at once, {cross it by $\times1$ to $\times2$, while Llama's}
single, more modest core remains in the linear regime through $\times4$ on
the true side. This locates the LayerNorm-mediated component of self-repair
reported for ablation studies
\citep{rushing2024explorationsselfrepairlanguagemodels} rather than
{contradicting it. Deletion-type} interventions sit in the shrinking regime,
where the mediator is real and must be controlled. We therefore retain the
frozen-denominator control for any deletion-type variant of the primary, to
which the audit above does not transfer.\footnote{The RMSNorm Jacobian is
$\rho^{-1}\!\left(I - \hat h \hat h^{\top}\right)$, so the scalar $g$
summarizes the isotropic part of the rescaling. The projection term and all
finite-swap curvature are included exactly in every reported effect, none of
{which is linearized, and $g$ serves} here only to bound the passive channel.}

\subsection{What this establishes, and what it leaves open}
\label{app:renorm-scope}

The final normalization is the only multiplicative mediator between a
weight-fixed write and the logits {on the direct path, because above the write's layer}
the residual stream is the identity. We have measured that mediator and
bounded its contribution to a few percent of $R$, with the wrong sign in half
the cells, the wrong dose profile in the rest, and no capacity in principle to
produce the sign crossings that define certification. We regard the
renormalization account of recruitment as excluded.

The causal claim itself never depended on this mediator. Every effect
reported in this paper is a total effect computed exactly in a deterministic,
fully observed structural causal model. Recruitment is the statement that a
direction's contribution to the verdict is larger in the {dosed} world than
in the clean one and, for the certified recruits, positive where it was
negative. {The interventions themselves establish that the verdict comes to depend on
these directions once the core is suppressed, without appeal to a model of
how the dependence arises.}

{It remains open whether recruitment is}
carried by re-expression, the detectors computing different coefficients
because their inputs have changed, or by re-routing through downstream
components. This concerns the internal route of an effect whose existence and
magnitude are already fixed. Both are the model computing, neither is an
artifact of measurement, and the choice between them does not bear on whether
the reserve is real. We leave the decomposition to future work, noting that
emergent self-repair was likewise established causally before any instance of
it received an end-to-end account.

\FloatBarrier

\FloatBarrier

\section{{Fits and Diagnostics of the Coupling Law}}
\label{app:coupling-fits}

\subsection{Estimation procedure}
\label{app:fits}

\begin{table}[!htbp]
\centering
\small
\setlength{\tabcolsep}{4pt}
\resizebox{\textwidth}{!}{\begin{tabular}{@{}lllll@{}}
\toprule
Intervention & Clamp $v$ & Dose $\lambda(v)$ & In expectation & Used by \\
\midrule
Counterfactual patch & $a_{x'}$ (partner)           & $-1$   & $-1$ & \citet{vig2020causalmediationanalysisinterpreting,geiger2025causalabstractiontheoreticalfoundation} \\
Neutral clamp        & $m$                          & $0$    & $0$  & this paper \\
Dosed swap           & $m-\tfrac{\lambda}{2}\Delta$ & chosen & {$\lambda$}   & this paper \\
\addlinespace
Zero ablation     & $0$                      & $2m/\Delta$                & unit-set; $0$ iff $m = 0$ & \citet{michel2019sixteen}; \citet{gong2026conditional} \\
Mean ablation     & $\bar a$                 & $2(m-\bar a)/\Delta$       & $2q-1$; $0$ if balanced   & \citet{wang2022interpretability} \\
Resample ablation & $a_c(x')$, $x'\!\sim\!D$ & $2(m-a_c(x'))/\Delta$      & $\lambda(\mu_D)$, sd $\rho_D$ & \citet{chan2022causal,mcgrath2023hydraeffectemergentselfrepair,rushing2024explorationsselfrepairlanguagemodels} \\
\bottomrule
\end{tabular}}
\caption{Every clamp is a dose.  For any $v$, Equation~{\ref{eq:dose}} gives
its position on the axis and Equation~{\ref{eq:law}} its downstream effect.  $q$ is the fraction of the presented class
{in the reference set. $\mu_D$} and $\sigma_D$ are the donor mean and standard
deviation, and $\rho_D = 2\sigma_D/|\Delta|$.  Measured placements are in
{Figure~\ref{fig:deletion_dose_a}, and derivations and per-core detail are in}
Appendix~\ref{app:deletion}.}
\label{tab:clamp}
\end{table}

For every bench direction we regress the class-summed contribution $E_r$ on
the signed dose $\lambda = 1-2k$ of Eq.~{\ref{eq:dose}}, using
the per-pair values rather than the tabulated means so that standard errors
propagate. The reported fit is on the on-manifold segment, $\lambda \in
\{+1, 0, -1\}$ (clean, class-neutral clamp, {full inversion), and the stress}
doses $\lambda \in \{-3,-5,-7\}$ are held out and used only to locate the
boundary below. Slopes are tested against $\gamma_r = 0$ with
{Benjamini-Hochberg control within each family at $q = 0.05$, and certification
uses the slope rather than $R^2$.} We fit each class separately and pooled, and
flag directions whose per-class slopes disagree in sign
(Table~\ref{tab:coupling}). Directions at or above the core's layer cannot
read its write and are excluded from all fits (\S\ref{app:wiring},
Table~\ref{tab:upstream}).

\paragraph{Where the fit stops holding.}
{Figures~\ref{fig:fitdiag-llama} to \ref{fig:fitdiag-mistral} plot, per}
direction, the observed $E_r$ against $\lambda$ with {each side's on-manifold fit extrapolated across the full range}. The \emph{lin.} column of Table~\ref{tab:coupling}
records the largest dose at which the residual from the on-manifold line
stays within twice its {combined standard error. For most} certified directions
this is $\times 1$, and the residuals turn systematically negative {beyond it, because the response saturates once the injected} signal is amplified past any
value an input attains. Ten of the sixty-eight certified directions remain
on the line past full inversion, six of them in Llama.

\paragraph{Crossing-dose comparison.}
$k^\star$ is obtained by setting Eq.~{\ref{eq:law}} to zero,
$\lambda^\star = -\mathrm{own}_r/\gamma_r$ and $k^\star = (1-\lambda^\star)/2$,
using only the fitted parameters. The reference $k_{\mathrm{obs}}$ is not a
{measurement. The ladder} is coarse, and we locate the crossing by linear
interpolation between the two rungs that straddle zero. Where those rungs lie
inside the on-manifold segment the interpolation is over a stretch the fit
{certifies as linear. Beyond full} inversion it is not, and because the true
response saturates there, the interpolated reference is biased toward
earlier crossings. The discrepancies in the \emph{beyond} column of
Table~\ref{tab:coupling-summary} are therefore upper bounds on the prediction
error rather than estimates of it.

\subsection{Family-level summary}
\label{app:coupling-family}
{Table~\ref{tab:coupling-summary} summarises the certified couplings by family.}

\begin{table}[!htbp]\centering\small\setlength{\tabcolsep}{5pt}
\caption{\textbf{The coupling law across families.} \emph{ds.} counts bench directions downstream of the {certified core(s), and \emph{cert.} those} whose coupling is reliably non-zero (BH-FDR within family, $q{=}0.05$). $R^2$ medians are given separately for MLP-neuron and attention-derived (OV channel, singular mode) directions. \emph{inh.} is the median share of the clean contribution inherited from the core, over the certified directions whose clean value is resolved ($n$ in parentheses). \emph{flip err.} is the median discrepancy between the predicted and the interpolated crossing dose, split by whether the crossing falls inside the on-manifold segment or beyond full inversion.}
\label{tab:coupling-summary}
\resizebox{\textwidth}{!}{\begin{tabular}{lccccccccc}
\toprule
 & & & \multicolumn{2}{c}{coupling sign} & \multicolumn{2}{c}{median $R^2$} & & \multicolumn{2}{c}{flip err.\ (doses)} \\
\cmidrule(lr){4-5}\cmidrule(lr){6-7}\cmidrule(lr){9-10}
model & ds. & cert. & $\gamma<0$ & $\gamma>0$ & mlp & ov/svd & inh. & in-window & beyond \\
\midrule
Llama-3-8B-Instruct     & 20 & 18 & 18 & 0 & 0.84 & 0.74 & 0.67 \small(6)  & 0.02 \small(13) & --- \\
Gemma-2-9b-it           & 22 & 17 & 14 & 3 & 0.78 & 0.72 & 0.38 \small(15) & 0.05 \small(2)  & 0.12 \small(12) \\
Qwen2.5-7B-Instruct     & 16 & 14 &  9 & 5 & 0.79 & 0.24 & 0.40 \small(13) & 0.02 \small(3)  & 0.22 \small(9) \\
Mistral-7B-Instruct-v0.3& 23 & 19 & 11 & 8 & 0.58 & 0.45 & 0.23 \small(13) & 0.03 \small(3)  & 0.22 \small(11) \\
\midrule
pooled                  & 81 & 68 & 52 & 16 & 0.76 & 0.68 & 0.39 \small(47) & 0.02 \small(21) & 0.19 \small(32) \\
\bottomrule
\end{tabular}}
\end{table}

\subsection{Per-direction fits}
\label{app:coupling-perdirection}
{Table~\ref{tab:coupling} lists every certified direction's fit.}

\begin{table*}[!htbp]\centering\footnotesize\renewcommand{\arraystretch}{0.9}\setlength{\tabcolsep}{4pt}
\caption{\textbf{The coupling law, per direction.} For every downstream bench direction with a coupling reliably different from zero (BH-FDR within family, {$q{=}0.05$), the table gives the on-manifold fit of Eq.~\ref{eq:law} (pooled intercept, as certified in App.~\ref{app:law-certification}, while the exemplar tables split it), with intercept} $\mathrm{own}_r$, slope $\gamma_r$ (negative $=$ \emph{counterweight}, positive $=$ \emph{relay}), and fit quality $R^2$. \emph{inh.}~$=\gamma_r/(\mathrm{own}_r{+}\gamma_r)$. $k^\star$ is the crossing dose predicted by the fit, $k_{\mathrm{obs}}$ the crossing located in the ladder {by interpolation, and $^{\dagger}$ marks} crossings beyond full inversion. \emph{lin.} is the largest dose at which the residual stays on the on-manifold line. $z$ is the weight-only wiring alignment of \S\ref{app:wiring}. Rows shaded \colorbox{black!8}{grey} are hybrids, whose clean class and coupling sign disagree. Directions upstream of or level with the cores {are omitted because they cannot read the core write. They are listed in} Table~\ref{tab:upstream}.}
\label{tab:coupling}
\begin{tabular}{lrrrcrrrcr}
\toprule
direction & $\mathrm{own}_r$ & $\gamma_r$ & $R^2$ & inh. & $k^\star$ & $k_{\mathrm{obs}}$ & lin. & $z$ \\
\midrule
\multicolumn{9}{l}{\textsc{Llama-3-8B-Instruct}}\\[1pt]
\texttt{L23.n8972} & -0.015 & -0.032 & 0.95 & 0.68 & 0.74 & 0.73 & $\times1$ & +4.7 \\
\texttt{L23.n9811} & -0.009 & -0.012 & 0.95 & 0.59 & 0.85 & 0.85 & $\times1$ & +5.1 \\
\texttt{L21.n4049} & -0.000 & -0.011 & 0.99 & 0.97 & 0.52 & 0.51 & $\times2$ & +0.5 \\
\texttt{L19.n3754} & -0.005 & -0.009 & 0.76 & 0.66 & 0.76 & 0.76 & $\times1$ & +7.0 \\
\texttt{L22.h1.svd117} & +0.003 & -0.007 & 0.95 & -- & 0.31 & 0.29 & $\times1$ & -0.8 \\
\texttt{L19.n2738} & -0.004 & -0.006 & 0.84 & 0.60 & 0.83 & 0.83 & $\times1$ & +4.8 \\
\texttt{L24.n4543} & +0.001 & -0.004 & 0.63 & -- & 0.43 & 0.51 & $\times1$ & +1.9 \\
\texttt{L20.n1390} & +0.006 & -0.004 & 0.75 & -- & -- & -- & $\times1$ & +4.2 \\
\texttt{L21.n14165} & +0.002 & -0.003 & 0.89 & -- & 0.13 & 0.12 & $\times1$ & +0.9 \\
\texttt{L22.h1.c60} & +0.003 & -0.002 & 0.74 & -- & -- & -- & $\times4$ & -1.0 \\
\texttt{L21.n798} & +0.001 & -0.002 & 0.61 & -- & 0.22 & 0.20 & $\times1$ & +3.7 \\
\texttt{L21.n517} & +0.001 & -0.002 & 0.90 & -- & 0.31 & 0.27 & $\times4$ & +1.7 \\
\texttt{L28.h27.c120} & +0.001 & -0.002 & 0.86 & -- & 0.35 & 0.33 & $\times1$ & +0.0 \\
\texttt{L20.n14044} & -0.000 & -0.002 & 0.46 & 0.93 & 0.54 & 0.37 & $\times1$ & -0.7 \\
\texttt{L22.h1.svd116} & +0.002 & -0.002 & 0.68 & -- & -- & -- & $\times4$ & -0.5 \\
\texttt{L22.svd323} & +0.000 & -0.002 & 0.68 & -- & 0.39 & 0.30 & $\times2$ & +0.3 \\
\texttt{L28.h27.c34} & +0.001 & -0.001 & 0.80 & -- & -- & -- & $\times4$ & -1.0 \\
\texttt{L28.h27.c43} & +0.001 & -0.001 & 0.37 & -- & -- & -- & $\times1$ & -1.2 \\
\addlinespace[2pt]
\multicolumn{9}{l}{\textsc{Gemma-2-9b-it}}\\[1pt]
\texttt{L38.n7088} & -0.195 & -0.653 & 0.92 & 0.77 & 0.65 & 0.71 & $\times1$ & +34.3 \\
\texttt{L41.n8771} & -0.049 & -0.092 & 0.87 & 0.65 & 0.77 & 0.80 & $\times1$ & +4.8 \\
\texttt{L38.n854} & -0.129 & -0.081 & 0.87 & 0.39 & 1.29 & 1.40$^{\dagger}$ & $\times1$ & +4.9 \\
\texttt{L39.n10210} & -0.087 & -0.064 & 0.93 & 0.42 & 1.18 & 1.29$^{\dagger}$ & $\times1$ & +28.8 \\
\texttt{L39.n10085} & -0.398 & -0.047 & 0.69 & 0.11 & 4.70 & 3.21$^{\dagger}$ & $\times1$ & +8.6 \\
\texttt{L32.n8200} & -0.069 & -0.043 & 0.18 & 0.38 & -- & -- & $\times1$ & +4.0 \\
\texttt{L36.n1119} & +0.048 & +0.038 & 0.34 & 0.44 & 1.13 & 1.01$^{\dagger}$ & $\times1$ & +4.2 \\
\texttt{L41.n3789} & -0.091 & -0.034 & 0.59 & 0.27 & 1.85 & 1.49$^{\dagger}$ & $\times1$ & +6.7 \\
\texttt{L34.n4784} & -0.023 & -0.018 & 0.98 & 0.44 & 1.13 & 1.13$^{\dagger}$ & $\times1$ & +1.4 \\
\texttt{L39.n12353} & +0.048 & +0.017 & 0.96 & 0.27 & 1.86 & 1.70$^{\dagger}$ & $\times1$ & +4.6 \\
\texttt{L30.n761} & +0.090 & +0.017 & 0.61 & 0.16 & 3.09 & 1.85$^{\dagger}$ & $\times1$ & +5.2 \\
\texttt{L33.n1466} & -0.048 & -0.014 & 0.16 & 0.22 & 2.24 & 1.63$^{\dagger}$ & $\times1$ & +2.3 \\
\texttt{L40.h11.svd18} & -0.025 & -0.011 & 0.72 & 0.30 & 1.66 & 1.62$^{\dagger}$ & $\times1$ & +1.9 \\
\texttt{L40.h7.c105} & -0.008 & -0.006 & 0.99 & 0.40 & 1.24 & 1.34$^{\dagger}$ & $\times2$ & +2.9 \\
\texttt{L40.h7.c177} & -0.009 & -0.004 & 0.82 & 0.32 & 1.56 & 1.65$^{\dagger}$ & $\times2$ & +3.2 \\
\rowcolor{black!8}\texttt{L34.h15.c22} & +0.011 & -0.002 & 0.50 & -- & -- & -- & $\times1$ & -1.1 \\
\rowcolor{black!8}\texttt{L31.h3.c137} & +0.011 & -0.001 & 0.18 & -- & -- & -- & $\times1$ & +3.1 \\
\addlinespace[2pt]
\multicolumn{9}{l}{\textsc{Qwen2.5-7B-Instruct}}\\[1pt]
\texttt{L27.n13033} & -1.790 & -1.162 & 0.97 & 0.39 & 1.27 & 1.14$^{\dagger}$ & $\times1$ & +18.7 \\
\texttt{L27.n4228} & -0.356 & -0.409 & 0.96 & 0.53 & 0.93 & 0.95 & $\times1$ & +18.3 \\
\texttt{L24.n8308} & +0.171 & +0.086 & 0.36 & 0.34 & 1.49 & 1.28$^{\dagger}$ & $\times1$ & +12.9 \\
\texttt{L23.n8341} & -0.058 & -0.063 & 0.90 & 0.52 & 0.96 & 0.98 & $\times1$ & +7.4 \\
\texttt{L27.n15192} & +0.149 & +0.049 & 0.97 & 0.25 & 2.00 & 1.50$^{\dagger}$ & $\times1$ & +7.1 \\
\texttt{L25.n11824} & +0.068 & +0.045 & 0.88 & 0.40 & 1.26 & 1.26$^{\dagger}$ & $\times1$ & +5.9 \\
\texttt{L25.n4929} & -0.048 & -0.043 & 0.61 & 0.47 & 1.05 & 1.12$^{\dagger}$ & $\times1$ & +7.3 \\
\texttt{L24.n12848} & -0.046 & -0.033 & 0.21 & 0.42 & 1.20 & 0.98$^{\dagger}$ & $\times1$ & +6.1 \\
\texttt{L25.n8444} & -0.003 & -0.029 & 0.77 & 0.90 & 0.55 & 0.43 & $\times1$ & +8.3 \\
\texttt{L23.n14476} & -0.028 & -0.026 & 0.79 & 0.48 & 1.05 & 1.16$^{\dagger}$ & $\times1$ & +8.9 \\
\texttt{L26.h17.svd117} & +0.105 & +0.007 & 0.07 & 0.06 & -- & -- & $\times1$ & -0.4 \\
\texttt{L24.n9321} & +0.013 & +0.004 & 0.37 & 0.21 & 2.33 & 1.39$^{\dagger}$ & $\times1$ & +1.5 \\
\rowcolor{black!8}\texttt{L23.h27.c82} & +0.015 & -0.003 & 0.24 & -- & -- & -- & $\times1$ & -0.7 \\
\texttt{L26.h20.svd17} & -0.007 & -0.002 & 0.72 & 0.18 & 2.81 & 1.97$^{\dagger}$ & $\times4$ & -0.1 \\
\bottomrule
\end{tabular}
\end{table*}

\begin{table*}[!htbp]\centering\footnotesize\renewcommand{\arraystretch}{0.9}\setlength{\tabcolsep}{4pt}
\addtocounter{table}{-1}
\caption{\textbf{The coupling law, per direction} (continued).}
\begin{tabular}{lrrrcrrrcr}
\toprule
direction & $\mathrm{own}_r$ & $\gamma_r$ & $R^2$ & inh. & $k^\star$ & $k_{\mathrm{obs}}$ & lin. & $z$ \\
\midrule
\multicolumn{9}{l}{\textsc{Mistral-7B-Instruct-v0.3}}\\[1pt]
\texttt{L31.n8773} & -0.393 & -0.398 & 0.98 & 0.50 & 0.99 & 0.96 & $\times1$ & +18.0 \\
\texttt{L31.n13669} & -1.850 & -0.208 & 0.58 & 0.10 & 4.96 & 1.88$^{\dagger}$ & $\times1$ & +14.3 \\
\texttt{L28.n4079} & -0.239 & -0.158 & 0.55 & 0.40 & 1.25 & 1.22$^{\dagger}$ & $\times1$ & +13.4 \\
\texttt{L24.n4907} & +0.095 & -0.093 & 0.90 & -- & -- & -- & $\times1$ & +13.7 \\
\rowcolor{black!8}\texttt{L22.h20.svd0} & +0.302 & -0.083 & 0.44 & -- & -- & -- & $\times1$ & +9.1 \\
\rowcolor{black!8}\texttt{L29.n8705} & +0.251 & -0.048 & 0.11 & -- & -- & -- & $\times1$ & +5.8 \\
\texttt{L24.n5072} & +0.070 & +0.042 & 0.75 & 0.38 & 1.33 & 1.14$^{\dagger}$ & $\times1$ & +14.5 \\
\texttt{L30.n11434} & -0.048 & -0.032 & 0.88 & 0.40 & 1.26 & 1.47$^{\dagger}$ & $\times1$ & +6.4 \\
\texttt{L22.h23.svd2} & -0.086 & -0.023 & 0.46 & 0.21 & 2.35 & 2.55$^{\dagger}$ & $\times1$ & +7.8 \\
\texttt{L25.h8.svd45} & +0.028 & +0.017 & 0.92 & 0.38 & 1.32 & 1.19$^{\dagger}$ & $\times1$ & +2.8 \\
\texttt{L22.n8735} & -0.080 & -0.016 & 0.35 & 0.17 & 3.02 & 2.17$^{\dagger}$ & $\times1$ & +2.5 \\
\texttt{L22.n5778} & -0.012 & +0.015 & 0.89 & -- & 0.11 & 0.07 & $\times1$ & +1.2 \\
\texttt{L30.n498} & +0.039 & +0.011 & 0.50 & 0.21 & 2.33 & 3.74$^{\dagger}$ & $\times1$ & +6.9 \\
\texttt{L30.n11755} & +0.004 & +0.010 & 0.74 & 0.71 & 0.70 & 0.66 & $\times1$ & +3.0 \\
\rowcolor{black!8}\texttt{L24.n1021} & -0.013 & +0.007 & 0.93 & -- & -- & -- & $\times1$ & +4.2 \\
\texttt{L28.n9060} & +0.017 & +0.005 & 0.30 & 0.22 & 2.24 & 2.40$^{\dagger}$ & $\times3$ & +0.4 \\
\rowcolor{black!8}\texttt{L22.h20.c93} & +0.014 & -0.004 & 0.44 & -- & -- & -- & $\times1$ & +3.5 \\
\texttt{L22.n1070} & -0.028 & -0.004 & 0.11 & 0.11 & 4.46 & 2.04$^{\dagger}$ & $\times1$ & +6.9 \\
\texttt{L27.n11713} & +0.006 & +0.002 & 0.08 & 0.23 & 2.13 & 1.59$^{\dagger}$ & $\times1$ & +2.5 \\
\bottomrule
\end{tabular}
\end{table*}

\subsection{Fit diagnostic figures}
\label{app:fit-diagnostics}

\begin{figure}[!htbp]
\centering
\includegraphics[width=\textwidth]{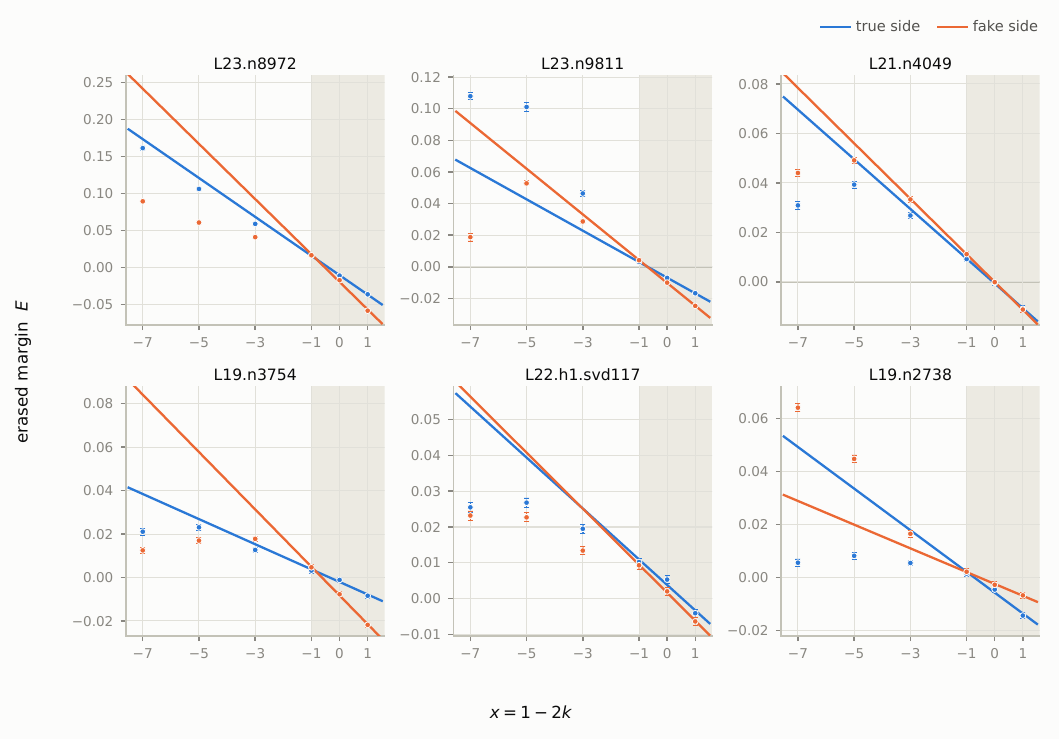}
\caption{{\textbf{Affine fits for the six
strongest-coupling downstream directions of Llama-3-8B-Instruct.}} Each panel plots the measured erased
margin $E$ against the dose regressor $\lambda = 1-2k$, separately for the
true-prompt side (blue) and the fake-prompt side {(orange). Markers are} the
archived per-dose means with their standard errors, so the six markers per
colour correspond to $k = 0, \tfrac12, 1, 2, 3, 4$. The shaded band marks the
on-manifold window $\lambda \in [-1,+1]$ (equivalently $k \leq 1$) that the
{headline fit uses. Each side's straight line is fitted on that window and \emph{extrapolated} across the full range. The two lines have their own slopes, so the panels show per-class fits rather than the shared-slope fit of Eq.~\ref{eq:law}, and drawing them separately makes a sign disagreement between the sides visible. The axis label $x=1-2k$ is $\lambda$.}}
\label{fig:fitdiag-llama}
\end{figure}

\begin{figure}[!htbp]
\centering
\includegraphics[width=\textwidth]{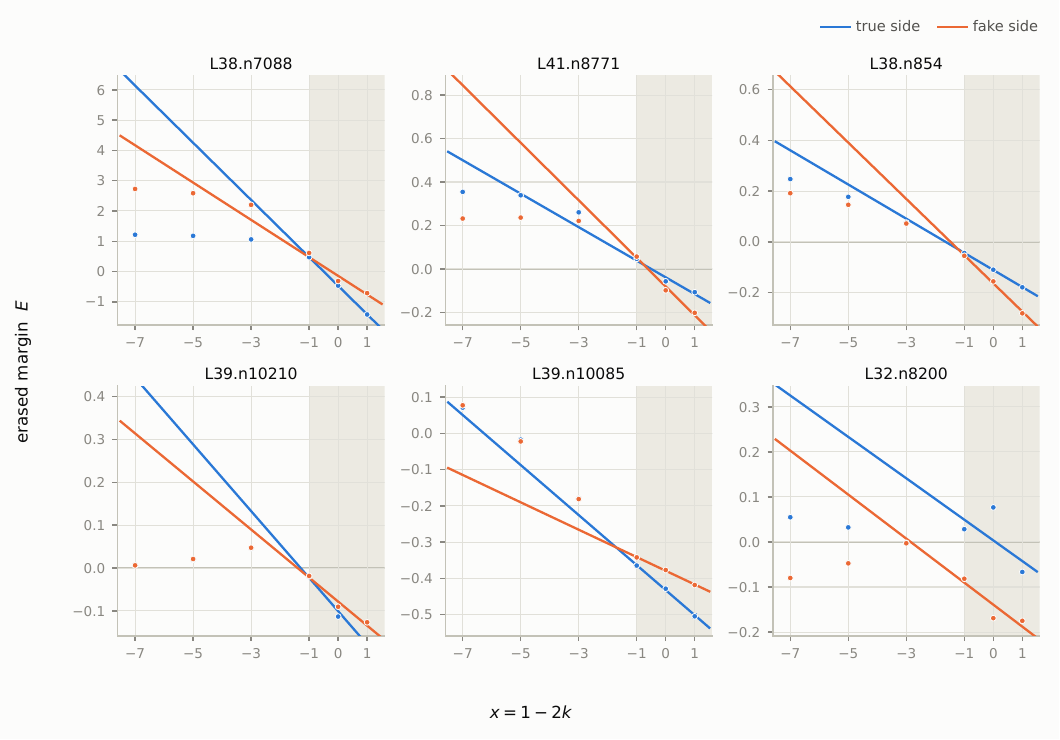}
\caption{{\textbf{Affine fits for the six strongest-coupling
downstream directions of Gemma-2-9B-it.}} Axes, colours, shading and fitted line as in
Figure~\ref{fig:fitdiag-llama} ($\lambda = 1-2k$). {The vertical
scale differs by roughly an order of magnitude between panels, since the coupling
strengths $|\gamma_r|$ vary widely within a single model. This spread is why
the wiring} correlations in the main text are reported on ranks rather than
raw values.}
\label{fig:fitdiag-gemma}
\end{figure}

\begin{figure}[!htbp]
\centering
\includegraphics[width=\textwidth]{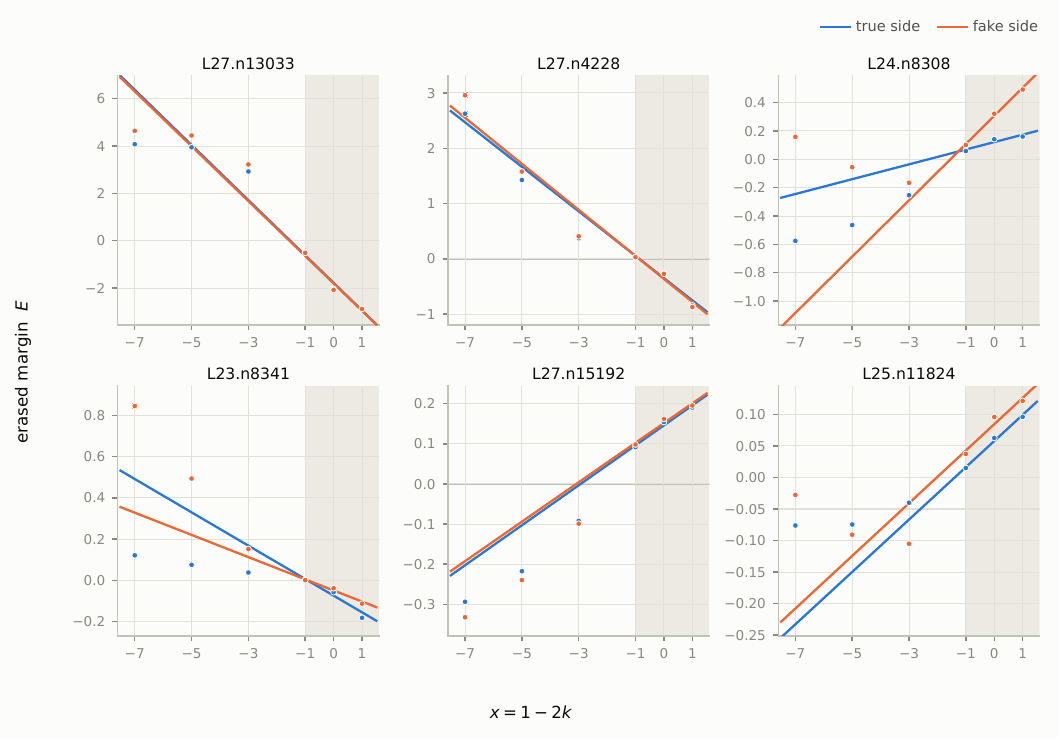}
\caption{{\textbf{Affine fits for the six
strongest-coupling downstream directions of Qwen2.5-7B-Instruct.}} Axes, colours, shading and fitted
line as in Figure~\ref{fig:fitdiag-llama} ($\lambda = 1-2k$).}
\label{fig:fitdiag-qwen}
\end{figure}

\begin{figure}[!htbp]
\centering
\includegraphics[width=\textwidth]{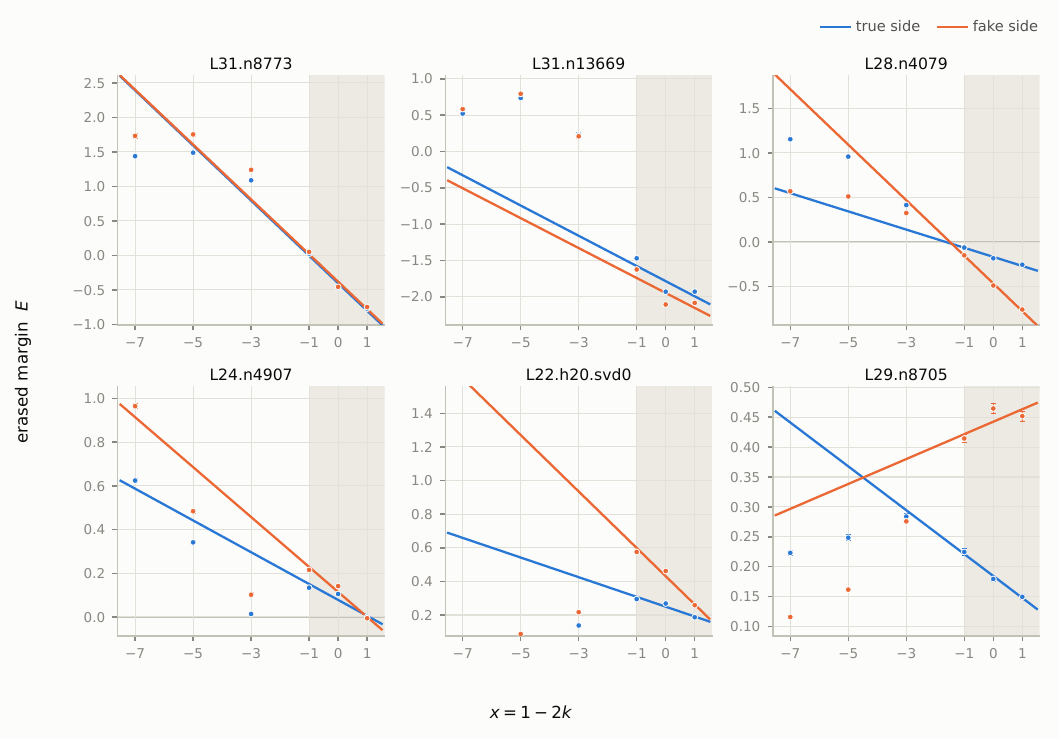}
\caption{{\textbf{Affine fits for the six
strongest-coupling downstream directions of Mistral-7B-Instruct-v0.3.}} Axes, colours, shading and fitted
line as in Figure~\ref{fig:fitdiag-llama} ($\lambda = 1-2k$).}
\label{fig:fitdiag-mistral}
\end{figure}

{Figures~\ref{fig:fitdiag-llama} to \ref{fig:fitdiag-mistral} plot, for the six}
strongest-coupling downstream directions per model, the measured erased
margin $E$ against the dose regressor $\lambda = 1-2k$, on-manifold window
shaded, {each side's window fit extrapolated as a straight line}
. Every per-direction statistic ($\gamma_r$,
its standard error, $R^2$, the saturation onset, and the BH-FDR
certification) is tabulated in Table~\ref{tab:coupling}, so the panels are
deliberately unannotated.

\subsection{Exemplar fits and the half-rung holdout}
\label{app:coupling-holdout}
{Tables~\ref{tab:law-exemplars-full} and~\ref{tab:window-robustness} give the exemplar fits and the effect of adding the half rungs.}

\begin{table}[!htbp]
\centering
\small
\caption{\textbf{The exemplar fits of main text Figure~{\ref{fig:law-exemplars}}, and where the post-fit rungs landed.} {For each exemplar direction the table gives the two} per-class intercepts, their difference $\Delta\mathrm{own}_r$, the shared slope with its standard error, the weighted $R^2$, and the slope refitted with the half rungs included. \emph{track} is how far the two parallel lines sit from their own class's fitted points, as a fraction of the exemplar's full data span. {The two half-rung columns give the out-of-sample landing as the largest deviation} of the four half-rung means from the line of their own class, in SEs and as a fraction of span. That line was fitted before those rungs existed.}
\label{tab:law-exemplars-full}
\resizebox{\textwidth}{!}{\begin{tabular}{llrrrrrrrrr}
\toprule
& & \multicolumn{3}{c}{intercepts} & \multicolumn{3}{c}{shared slope} & \multicolumn{2}{c}{half rungs} & \\
\cmidrule(lr){3-5}\cmidrule(lr){6-8}\cmidrule(lr){9-10}
Model & Direction & $\mathrm{own}^{T}$ & $\mathrm{own}^{F}$ & $\Delta\mathrm{own}$ & $\gamma_r$ (SE) & $R^2$ & $\gamma_r^{\,5\mathrm{pt}}$ & max $|z|$ & \% span & track \\
\midrule
Llama-3-8B & \texttt{L23.n8972} & $-0.0111$ & $-0.0190$ & $+0.0080$ & $-0.0319\;(0.0007)$ & $0.969$ & $-0.0328$ & $6.1$ & $1.4\%$ & $3.3\%$ \\
Gemma-2-9B & \texttt{L38.n7088} & $-0.3318$ & $-0.1034$ & $-0.2285$ & $-0.7272\;(0.0051)$ & $0.944$ & $-0.7284$ & $17.9$ & $7.2\%$ & $8.7\%$ \\
Qwen2.5-7B & \texttt{L27.n13033} & $-1.7789$ & $-1.8004$ & $+0.0215$ & $-1.1619\;(0.0153)$ & $0.973$ & $-1.1542$ & $5.6$ & $2.4\%$ & $3.7\%$ \\
Mistral-7B-v0.3 & \texttt{L31.n8773} & $-0.4038$ & $-0.3815$ & $-0.0223$ & $-0.3986\;(0.0119)$ & $0.980$ & $-0.3962$ & $2.1$ & $2.1\%$ & $2.9\%$ \\
\bottomrule
\end{tabular}}
\end{table}

\begin{table}[!htbp]
\centering
\small
\caption{\textbf{The affine window survives doubling its interior sampling.} {With the two half rungs $\lambda=\pm\tfrac12$ added ($478$ to $598$ pairs each, same population as the ladder), the columns give agreement between} $\gamma_r$ fitted on five rungs and on the original three, the median slope shift, and two Occam columns (median $|\gamma|$ change and misfit removed when a quadratic term is added). $R^2$ is the per-class variance in $E$ explained by the affine law over the five rungs, pooled over directions. {The last column is a holdout test in which the law's parameters, fitted on the original three rungs alone, are scored against the two half-rung means the fit never saw} ($324$ held-back means). The same test run on the pooled single-intercept form the split intercepts replaced gives $0.86$, $0.59$, $0.87$ {and $-0.41$, so out of sample} the pooled form is worse than predicting each direction-class's own mean in Mistral.}
\label{tab:window-robustness}
\resizebox{\textwidth}{!}{\begin{tabular}{lrrrrrrrrr}
\toprule
& & \multicolumn{3}{c}{$\gamma$: 5 rungs vs 3} & \multicolumn{1}{c}{shift} & \multicolumn{2}{c}{$+$quadratic} & & \\
\cmidrule(lr){3-5}\cmidrule(lr){6-6}\cmidrule(lr){7-8}
Model & $n$ & $r$ & $\rho_{|\gamma|}$ & sign & med.\,$|\Delta\gamma|/\mathrm{SE}$ & med.\,$|\Delta\gamma|/|\gamma|$ & misfit rem. & $R^2$ & OOS $R^2$ \\
\midrule
Llama-3-8B & 20 & $0.99957$ & $0.982$ & $1.00$ & $0.33$ & $0.6\%$ & $5\%$ & $0.981$ & $0.905$ \\
Gemma-2-9B & 22 & $0.99998$ & $0.998$ & $1.00$ & $0.39$ & $0.9\%$ & $4\%$ & $0.974$ & $0.803$ \\
Qwen2.5-7B & 16 & $0.99999$ & $0.997$ & $1.00$ & $0.51$ & $2.5\%$ & $9\%$ & $0.965$ & $0.904$ \\
Mistral-7B-v0.3 & 23 & $0.99973$ & $0.999$ & $1.00$ & $0.26$ & $1.3\%$ & $3\%$ & $0.913$ & $0.653$ \\
\bottomrule
\end{tabular}}
\end{table}

\FloatBarrier

\section{Tests of the Coupling Law}
\label{app:law-tests}

This appendix carries the evidence for \S\ref{sec:law}. It establishes that
the coupling $\gamma_r$ is resolved for most of the bench and does not move
under any {refit we tried, that the affine} form with one intercept per class is
the right leading-order description on $|\lambda|\le1$, predicting doses it
{never saw, and that the description} stops at full inversion. It does not
establish that the affine form is selected over other smooth forms, and
\S\ref{app:law-form-test} says why. Every number comes from a committed
{artifact, and none is hand-computed.}

\subsection{Fit and population}
\label{app:law-fitting}

The dose is $\lambda=1-2k$, where $k$ is the fraction of the swap toward the
{counterfactual, with $k=0$ clean} ($\lambda=+1$), $k=\tfrac12$ neutral
($\lambda=0$), $k=1$ full inversion {($\lambda=-1$), and $k\in\{2,3,4\}$}
overshoot. Eq.~\ref{eq:law} is fitted by weighted least squares on the three
on-manifold rungs $\lambda\in\{+1,0,-1\}$, both classes at once, one intercept
per class and one shared slope, with weights $1/\sigma^2$ from the per-rung
standard error of the mean over per-prompt paired differences. The window was
fixed before any fit and is the same for {every direction, while the half rungs}
$\lambda=\pm\tfrac12$ and the overshoot rungs are held out.
Table~\ref{tab:app-setup} gives the population. The class-$F$ margin enters
$E$ with its sign flipped {so that on both classes a positive $E$ means
the direction moved} the verdict the way the swap pushes it.

\begin{table}[!htbp]
\centering\small
\begin{tabular}{@{}lrrr@{}}
\toprule
Model & Downstream directions & Correction family & Prompt pairs \\
\midrule
Llama-3-8B-Instruct      & 20 & 20 & 532 \\
Gemma-2-9B-it            & 22 & 22 & 583 \\
Qwen2.5-7B-Instruct      & 16 & 22 & 478 \\
Mistral-7B-Instruct-v0.3 & 23 & 33 & 598 \\
\midrule
Total & 81 & 97 & \\
\bottomrule
\end{tabular}
\caption{\textbf{The fitted population.} Directions downstream of a core
{enter the fits, and the correction} family is the full bench, which in Qwen and
Mistral includes 6 and 10 upstream placement controls. The same prompt pairs
are measured at every dose.}
\label{tab:app-setup}
\end{table}

\subsection{Certification}
\label{app:law-certification}

A direction is certified when $|\gamma_r/\mathrm{se}(\gamma_r)|$ survives
{Benjamini-Hochberg at $q=0.05$} over its family's full bench. The gate tests
{the slope alone, and $R^2$ plays} no part. Under the pooled-intercept fit 68 of 81
directions certify, 52 counterweights ($\gamma_r<0$) and 16 relays
{($\gamma_r>0$). Under the split-intercept} fit 69 certify, the same 68 plus
Gemma \texttt{L31.n4305} ($p=0.047$ pooled, $0.026$ split). \S\ref{sec:law}
quotes the 68, which pass both gates.

{The 13 that do not certify fail a significance gate rather than a linearity test}
(Table~\ref{tab:app-exceptions}). Seven have slopes indistinguishable from
{zero, and six sit narrowly outside the correction. $R^2$ explains none of them.
Llama} \texttt{L28.h27.c0} (0.59), Gemma \texttt{L31.h3.c72} (0.53) and
Mistral \texttt{L30.h3.c8} (0.55) fit reasonably and still do not resolve a
slope, because on three to six points $R^2$ is a weak diagnostic independent of
whether $\gamma_r$ is resolved. An uncertified direction is not shown to be
{uncoupled, and its role is unresolved at this sample size.}

\begin{table}[!htbp]
\centering\small
\begin{tabular}{@{}lrl@{}}
\toprule
Status & Count & $p$ (pooled fit) \\
\midrule
Slope indistinguishable from zero        & 7 & $0.37$--$0.99$ \\
Excluded at the margin of the correction & 6 & $0.031$--$0.077$ \\
\bottomrule
\end{tabular}
\caption{\textbf{The 13 uncertified directions.}}
\label{tab:app-exceptions}
\end{table}

The gate is conservative in two ways. The correction family is the full
bench, so in Qwen and Mistral it includes the upstream placement controls
(22 and 33 tests rather than 16 and 23). And the borderline decisions are
sensitive to the rounding of the {archived statistics (\S\ref{app:law-tests}).
Pushing every} $\sigma$ to its rounding extreme moves the certified count to
69 or 70, and every direction that flips (\texttt{L30.h3.c8},
\texttt{L38.h15.c111}, \texttt{L30.h13.c170}) is already among the borderline
six. {The coefficients are stable, with $\gamma_r$ over the certified set shifting by a}
median $0.04\%$.

\subsection{One slope, two intercepts}
\label{app:law-form}

Fitting each class with its own free slope and correlating the two across
the 81 directions gives {Pearson $r=0.954$. Because} $|\gamma_r|$ spans orders of
magnitude and a few large couplings carry $r$, the rank correlation,
Spearman $\rho=0.834$, is the scale-free statement.

{Splitting the intercept by class makes the law predictive}
(Table~\ref{tab:app-split}). In sample it raises the median per-direction $R^2$
in every family, {from between $0.50$ and $0.76$ to between $0.82$ and $0.91$.} Out of sample, on the
half rungs of \S\ref{app:law-halfrungs}, the single-intercept form predicts
Mistral worse than each direction's own mean would {($R^2=-0.41$), while the split}
form, fitted on the same three rungs with no new data, turns that into a usable
forecast. We report medians across directions because a variance-weighted
$R^2$ is dominated by a few large-coupling directions.

\begin{table}[!htbp]
\centering\small
\begin{tabular}{@{}lrrrr@{}}
\toprule
& \multicolumn{2}{c}{in-sample median $R^2$} & \multicolumn{2}{c}{held-out $R^2$} \\
\cmidrule(lr){2-3}\cmidrule(l){4-5}
Model & split (reported) & single & split (reported) & single \\
\midrule
Llama   & 0.874 & 0.755 & 0.905 & 0.858 \\
Gemma   & 0.911 & 0.599 & 0.803 & 0.586 \\
Qwen    & 0.821 & 0.665 & 0.904 & 0.867 \\
Mistral & 0.903 & 0.499 & 0.653 & $-0.406$ \\
\bottomrule
\end{tabular}
\caption{\textbf{The split intercept, in sample and out of sample.} {The left columns give the
median per-direction} $R^2$ of the affine law on $\lambda\in\{+1,0,-1\}$ under
the two {intercept forms. The right columns give the variance-weighted} $R^2$ of the frozen
three-rung line on the 324 held-back half-rung class means, which the line
never saw.}
\label{tab:app-split}
\end{table}

\subsection{Doses the fit never saw}
\label{app:law-halfrungs}

We ran $\lambda=\pm\tfrac12$ on the same population after the three-rung fits
were frozen and scored each frozen line against the 324 held-back class means
(80, 88, 64 and 92 by family). {Each score is therefore a forecast. The
variance-weighted held-out $R^2$ is $0.65$ to $0.91$ by family
(Table~\ref{tab:app-split}), and this is the out-of-sample number, rather than
the wider-window in-sample figure of $0.91$ to $0.98$.}

The coefficient is stable under the new data. Refitting with the half rungs
included shifts $\gamma_r$ by a median {$1.7$ to $2.1\%$} of its magnitude in three
families and $7.2\%$ in Llama, in every case {$0.26$ to $0.51$} of its own
standard error, with the sign preserved on all 81 directions. Adding a free
quadratic term moves $\gamma_r$ by a {further $0.6$ to $2.5\%$}, again below its
standard error.

The weighted $R^2$ describes the largest couplings, so we screened every
certified direction carrying at least $1\%$ of its family's held-out weight for
a held-out $R^2$ below $0.30$. {Only Mistral \texttt{L31.n13669} qualifies. It is a
counterweight} with $\gamma_r=-0.21$ that fits in sample at $R^2=0.72$, carries
$12\%$ of Mistral's weight, and predicts its held-out doses no better than its
mean ($R^2=-0.01$).

Table~\ref{tab:law-exemplars} gives the four exemplars of
Figure~\ref{fig:law-exemplars}, chosen from the top two to six fits by $R^2$
among each family's {certified directions. All four} are MLP neurons and
counterweights. The hollow markers in the figure are the half-rung means.

\begin{table}[!htbp]
\centering\small
\begin{tabular}{@{}llrrrrrr@{}}
\toprule
& & \multicolumn{2}{c}{intercepts} & \multicolumn{2}{c}{shared slope} & \multicolumn{2}{c}{half rungs} \\
\cmidrule(lr){3-4}\cmidrule(lr){5-6}\cmidrule(l){7-8}
Model & Direction & $\mathrm{own}^{T}_r$ & $\mathrm{own}^{F}_r$ &
$\gamma_r$ (SE) & $R^2$ & $\gamma^{\,5\text{pt}}_r$ & \% span \\
\midrule
Llama-3-8B      & \texttt{L23.n8972}  & $-0.011$ & $-0.019$ & $-0.032\ (0.0007)$ & $0.97$ & $-0.033$ & $1.4$ \\
Gemma-2-9B      & \texttt{L38.n7088}  & $-0.332$ & $-0.103$ & $-0.727\ (0.005)$  & $0.94$ & $-0.728$ & $7.2$ \\
Qwen2.5-7B      & \texttt{L27.n13033} & $-1.779$ & $-1.800$ & $-1.162\ (0.015)$  & $0.97$ & $-1.154$ & $2.4$ \\
Mistral-7B-v0.3 & \texttt{L31.n8773}  & $-0.404$ & $-0.382$ & $-0.399\ (0.012)$  & $0.98$ & $-0.396$ & $2.1$ \\
\bottomrule
\end{tabular}
\caption{\textbf{The exemplar fits, and where the held-out rungs landed.}
Per-class intercepts and shared slope from the three-rung fit with its
{weighted $R^2$, the slope refitted with the half rungs included, and the}
largest deviation of the four half-rung means from their class's line, as a
fraction of the direction's data span.}
\label{tab:law-exemplars}
\end{table}

\subsection{What the holdout does and does not show}
\label{app:law-form-test}

The half rungs lie between the training rungs, so predicting them is
interpolation, and a predictor that assumes no law can be scored on the same
points (Table~\ref{tab:app-formtest}). Straight interpolation between the
bracketing measured rungs beats the affine line by a factor of four in
$\chi^2$, and a quadratic through the three training points {by six, and the same}
ordering holds when the neutral rung is withheld (\S\ref{app:law-neutral}).
Curvature inside the window is therefore resolvable at the precision of the
standard errors. {It is also small. The quadratic term moves $\gamma_r$ by
$0.6$ to $2.5\%$, below its standard error, and the affine line still captures
$65$ to $91\%$ of} the held-out variance. The holdout establishes that the dose
response is smooth and that the affine line tracks it to leading order with a
coefficient that does not depend {on the form, but it does not select the affine
form over its smooth rivals. A test that did would have to extrapolate} beyond the
window, and \S\ref{app:law-stops} shows why that is not available.

\begin{table}[!htbp]
\centering\small
\begin{tabular}{@{}lrrrr@{}}
\toprule
Withheld & points & affine line $\chi^2$ & bracket interpolation $\chi^2$ & quadratic $\chi^2$ \\
\midrule
half rungs $\lambda=\pm\tfrac12$ & 324 & 7501 & 1929 & 1188 \\
neutral rung $\lambda=0$         & 162 & 3221 &  530 &  390 \\
\bottomrule
\end{tabular}
\caption{\textbf{Three predictors on the same held-out points.} The affine
line is the frozen fit {(three rungs for the half rungs, the four outer rungs for}
the neutral rung). Bracket interpolation joins the two measured rungs on either
side of the withheld one and {assumes no law. The quadratic passes} exactly
through the training rungs. $\chi^2$ is in units of the per-rung standard
{error, and lower is better.}}
\label{tab:app-formtest}
\end{table}

\subsection{Leave one rung out}
\label{app:law-neutral}

On the five-rung ladder we withheld each rung in turn, refit on the other four,
and scored the prediction at the withheld rung {(Table~\ref{tab:neutral-rung},
with 162 class means per row)}. The three interior rungs are predicted to
{$1.36$ to $1.64$ standard errors and a median $8$ to $9\%$ of each direction's data
span, and the two endpoints} to about $3.0$ SE and $17\%$ of span. Extrapolating past
the last anchor costs twice what interpolating does, which is the leverage of
any line, and it is independent support for fitting on $|\lambda|\le1$ rather
than a wider window. The single-intercept form is worse at every rung but
$-1$, where both forms extrapolate.

\begin{table}[!htbp]
\centering\small
\begin{tabular}{@{}lrrrrl@{}}
\toprule
Withheld rung & median $|z|$ & single-intercept $|z|$ & median \% of span & frac.\ $|z|>2$ & \\
\midrule
$+1$          & 2.97 & 4.14 & 17.9 & 0.62 & endpoint \\
$+\tfrac12$   & 1.39 & 2.36 &  9.1 & 0.36 & interior \\
$0$ (neutral) & 1.36 & 2.19 &  9.0 & 0.36 & interior \\
$-\tfrac12$   & 1.64 & 1.94 &  8.4 & 0.41 & interior \\
$-1$          & 3.04 & 2.73 & 16.5 & 0.59 & endpoint \\
\bottomrule
\end{tabular}
\caption{\textbf{Leave-one-rung-out.} Each row refits on the other four
rungs and scores the prediction at the withheld rung, in standard errors
of the withheld measurement and as a fraction of the direction's data span.}
\label{tab:neutral-rung}
\end{table}

\subsection{Where the law stops}
\label{app:law-stops}

\begin{figure}[!htbp]
\centering
\includegraphics[width=\textwidth]{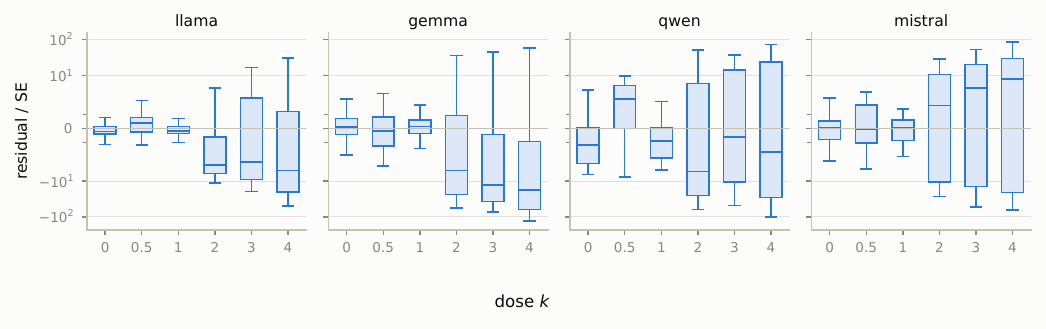}
\caption{\textbf{Standardised residuals from the on-manifold fit, by dose, for
all downstream directions in each model.} Each box aggregates the residuals
$(E - \hat{E})/\mathrm{SE}$ over every downstream direction and both verdict
sides {at that dose. Whiskers are} the usual $1.5\times$IQR range and outliers are
suppressed. The vertical axis is symmetric-logarithmic, linear within
$\pm 2$ SE and logarithmic beyond, so that the tightly clustered residuals at
$k \leq 1$ and the heavy tails at the stress doses are both resolvable in one
frame. Residuals at $k = 0, \tfrac12, 1$ are flat and centred on zero by construction {(those are the three points the fit uses),} so the informative
content is the abrupt fanning at $k = 2, 3, 4$, which is the same saturation
seen per direction in {Figures~\ref{fig:fitdiag-llama} to \ref{fig:fitdiag-mistral}}
and is why the headline $\gamma_r$ is estimated on the $k \leq 1$ window rather
than the full ladder.}
\label{fig:fitdiag-residuals}
\end{figure}

Figure~\ref{fig:fitdiag-residuals} plots the standardised residuals of the
on-manifold fit at every dose, including the three overshoot rungs the fit never
saw. Inside the window the {residuals are small, and beyond it} they grow with dose
as the response saturates and the line overpredicts. For each direction we
record the last dose at which the measurement stays within two standard errors
{of its line, and for all four exemplars} it is $k=1$, full inversion. This is why the
headline fit uses $k\le1$ and not the six-dose ladder, and why no extrapolative
test of the affine form is available on it. Gemma \texttt{L38.n7088} is shown
with its held-out points in Figure~\ref{fig:law-exemplars} because its curvature
is the best resolved on the bench (first of 81 by the $t$-statistic of a freed
{quadratic term), though not the largest.}

\section{{Out-of-Domain Test on GPT-2 Small and Indirect-Object Identification}}
\label{app:ioi}

\subsection{Setup}

The results so far were obtained on one task family, in four models of
comparable size, over a bench of components that we selected ourselves. The
natural concern is that the coupling coefficient is an artefact of those
choices. Indirect-object identification changes all three at once, and none
of the changes is ours to make. The model is GPT-2 small, two orders of
magnitude smaller than the models above. The task is syntactic name binding
rather than factual verification. The component bench is Wang et al.'s
published IOI circuit together with the copy-suppression head \texttt{L10.h7}
identified by McDougall et al., so the assignment of heads to roles, and in
particular which heads the intervention can reach and which it cannot, was
settled before this work began.

The intervention and the estimator are the ones used throughout. A prompt pair
is a single IOI template instantiated twice with the two names exchanged, so
that the margin $D_{AB} = \mathrm{logit}(A) - \mathrm{logit}(B)$ is faithful on
one member of the pair and contrastive on the other, exactly as $D_{TF}$ is on
the factual task, and the erased margin $E$ is built from it by the same
definition. The core is the name-mover pair \texttt{L9.h6}\,$+$\,\texttt{L9.h9},
swapped at doses $k \in \{\tfrac14, \tfrac12, \tfrac34, 1, 2, 3, 4\}$, so that
$\lambda = 1 - 2k$ runs over $\{+\tfrac12, 0, -\tfrac12, -1, -3, -5, -7\}$,
with $300$ prompt pairs at every rung.

The published circuit gives the law two groups to predict on. Ten heads lie in
layers $10$ and $11$, where the swap can {reach them, and ten lie in layer} $8$ and
earlier, where it cannot. If $\gamma_r$ measures what a head inherits from the
core, it should resolve in the first group and vanish in the second. The two
remaining heads of the circuit, \texttt{L9.h7} and \texttt{L9.h0}, sit in the
core's own layer, where a swap of \texttt{L9.h6} and \texttt{L9.h9} cannot
reach them, and are excluded.

\subsection{Seven of the ten reachable heads follow the law}

\begin{figure}[t]
\centering
\includegraphics[width=\textwidth]{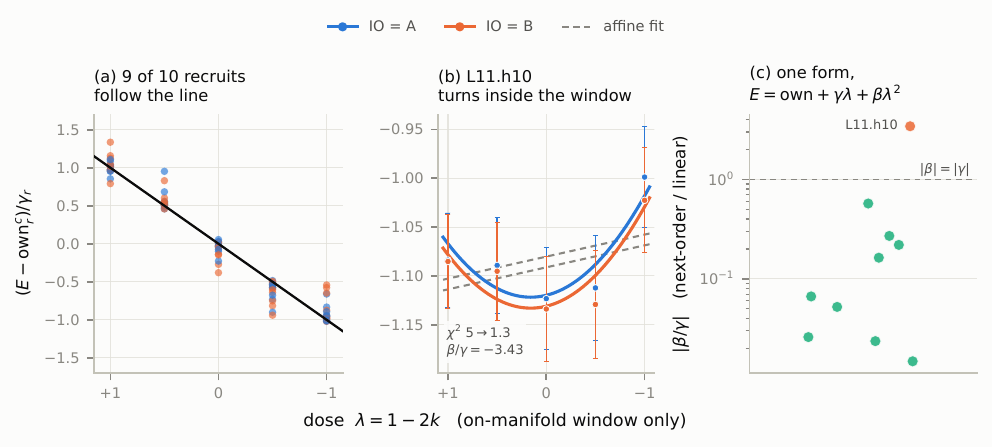}
\caption{\textbf{Which heads require a next-order term.} {The panels show only reachable recruits,
the heads above the core's layer.} Adding $\beta_r\lambda^2$ and taking
$|\beta_r/\gamma_r|$ divides them cleanly. Nine of ten sit at or below $0.572$
{and follow the line \emph{(a)}. One, \texttt{L11.h10} at $3.430$, sits well}
above it, a sixfold gap with nothing inside \emph{(c)}. Its parabola turns
inside the window \emph{(b)}, and its $\gamma_r$ does not certify, so the
quadratic is free to dominate a ratio whose denominator is unresolved. Built
from the clean-reference ladder.}
\label{fig:ioi-exceptions}
\end{figure}

\begin{figure}[!htbp]
\centering
\includegraphics[width=0.85\linewidth]{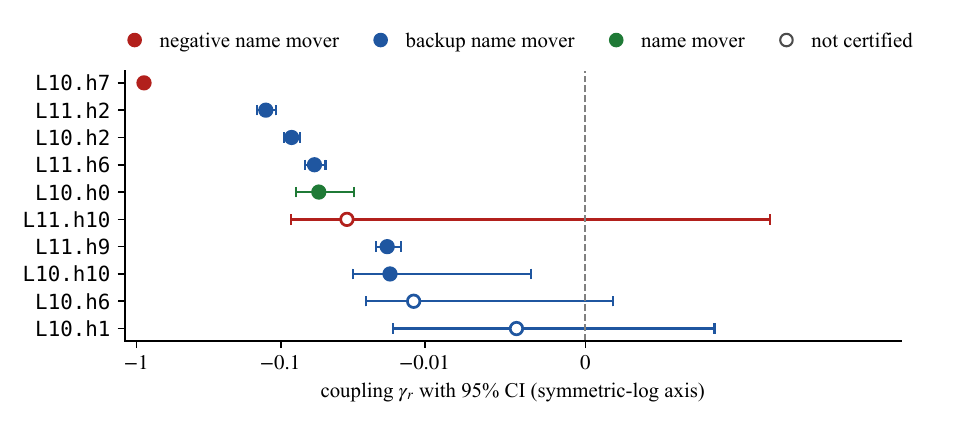}
\caption{\textbf{Coupling of the ten reachable IOI heads.} $\gamma_r$ with
$95\%$ intervals {from Table~\ref{tab:ioi-heads}. Filled markers are
significant after Benjamini-Hochberg} correction at $q = 0.05$. Every point
estimate is negative, and \texttt{L10.h7}, the copy-suppression head,
dominates.}
\label{fig:ioi-gamma}
\end{figure}

Each reachable head is fitted as on the factual ladders, with one intercept
per class and one shared slope $\gamma_r$, on the on-manifold window
$|\lambda| \le 1$ (Figure~\ref{fig:ioi-exceptions}a). A head follows the law if
$\gamma_r$ {is nonzero after Benjamini-Hochberg} correction at $q = 0.05$ and
the response is affine, in that a freed quadratic term $\beta_r\lambda^2$ is
not required, $|\beta_r/\gamma_r| \le 1$.
Seven of the ten reachable heads do (Table~\ref{tab:ioi-heads},
Figure~\ref{fig:ioi-gamma}), with a median weighted $R^2$ of $0.971$ across
the seven. Two of the remaining three, \texttt{L10.h1} and \texttt{L10.h6},
have $|\gamma_r| \le 0.012$, within two standard errors {of zero, so the swap does}
not measurably reach them, and their response is a flat line. The third,
\texttt{L11.h10}, has a curved response, and the next subsection shows why.

\begin{table}[!htbp]
\centering\small
\caption{\textbf{The coupling law on the ten reachable IOI heads.} Shared
slope $\gamma_r$ with its standard error, fitted on the on-manifold window.
Roles are Wang et al.'s. Every certified head is a counterweight, with
$\gamma_r < 0$.}
\label{tab:ioi-heads}
\begin{tabular}{llrrc}
\toprule
Head & Role & $\gamma_r$ & (SE) & cert. \\
\midrule
\texttt{L10.h7}  & negative name mover & $-0.8864$ & $0.0211$ & yes \\
\texttt{L11.h2}  & backup name mover   & $-0.1269$ & $0.0095$ & yes \\
\texttt{L10.h2}  & backup name mover   & $-0.0841$ & $0.0055$ & yes \\
\texttt{L11.h6}  & backup name mover   & $-0.0584$ & $0.0048$ & yes \\
\texttt{L10.h0}  & name mover          & $-0.0545$ & $0.0120$ & yes \\
\texttt{L11.h10} & negative name mover & $-0.0348$ & $0.0253$ & --  \\
\texttt{L11.h9}  & backup name mover   & $-0.0183$ & $0.0019$ & yes \\
\texttt{L10.h10} & backup name mover   & $-0.0175$ & $0.0072$ & yes \\
\texttt{L10.h6}  & backup name mover   & $-0.0120$ & $0.0070$ & --  \\
\texttt{L10.h1}  & backup name mover   & $-0.0043$ & $0.0063$ & --  \\
\bottomrule
\end{tabular}
\end{table}

Two features of the table stand out. Every head that follows the law is a
counterweight, {with $\gamma_r < 0$, and there are no relays, while the factual}
bench divides $52$ to $16$. A circuit built to move a single name into the
output position couples to its own core only in opposition to it. And the
coupling is dominated by one head. \texttt{L10.h7}, the copy-suppression head,
reaches $|\gamma_r| = 0.886$, seven times the next largest value and twenty
times the median over the reachable heads. The strongest coupling on the bench
falls on the head that prior work independently identified as the circuit's
negative feedback element.

\begin{figure}
    \centering
    \includegraphics[width=\linewidth]{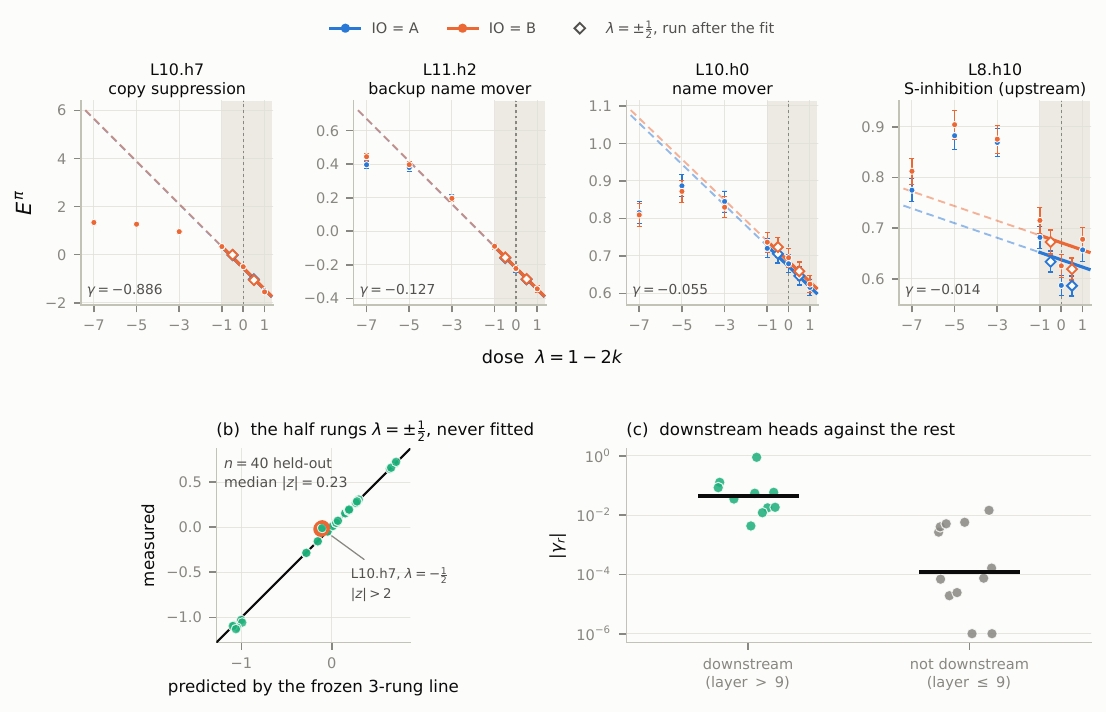}
    \caption{\textbf{The coupling law transfers to GPT-2 small on IOI.}
{\emph{(a)} Dose response of four heads, solid inside the window}
$|\lambda|\le1$ where the law is fitted, dashed outside. \emph{(b)} The 40
held-out means at $\lambda=\pm\tfrac12$ {against the frozen three-rung line, with
the two beyond $2$\,SE ringed.} \emph{(c)} $|\gamma_r|$ for heads the core
swap can reach (above layer~9) and cannot (at or below it).}
    \label{fig:ioi-law}
\end{figure}

The window is read off the data rather than imposed on it. Outside
$|\lambda| \le 1$ {the response saturates. Eight of the ten} reachable heads
leave the two-standard-error band around the fitted line before
$\lambda = -7$, four of them by $\lambda = -3$, and the only two that never
leave it are the flat heads \texttt{L10.h1} and \texttt{L10.h6}. This is the
same saturation that motivates restricting to $k \le 1$ on the factual ladders.
Inside the window the line predicts. The half rungs at $\lambda = \pm\tfrac12$
were run after the fits had been frozen and {never entered them. Over the forty}
held-out class means the median $|z|$ is $0.23$, and only two exceed two standard errors, both at \texttt{L10.h7} for $\lambda = -\tfrac12$, the head with by far the steepest slope (Figure~\ref{fig:ioi-law}b). Because these rungs lie inside the fitted window {this is interpolation. It shows that the line tracks the response but does not show that it beats every smooth alternative.}

The law also predicts where the coefficient should vanish, and it does. No
head outside the downstream set certifies (Figure~\ref{fig:ioi-law}c). Six of
these twelve do not move at all under the intervention, with
$\max|E^{\pi} - E^{0}| < 10^{-3}$ across the ladder. Among the six that do,
whose clean effects reach $E^{0} = +0.68$, the median $|\gamma_r|$ is
$4.5 \times 10^{-3}$, against $4.5 \times 10^{-2}$ {for the downstream heads, a
separation of roughly tenfold, on a division of heads that was fixed by the
published circuit \citep{wang2022interpretability}.}

\subsection{The exception}

\begin{figure}[!htbp]
\centering
\includegraphics[width=\textwidth]{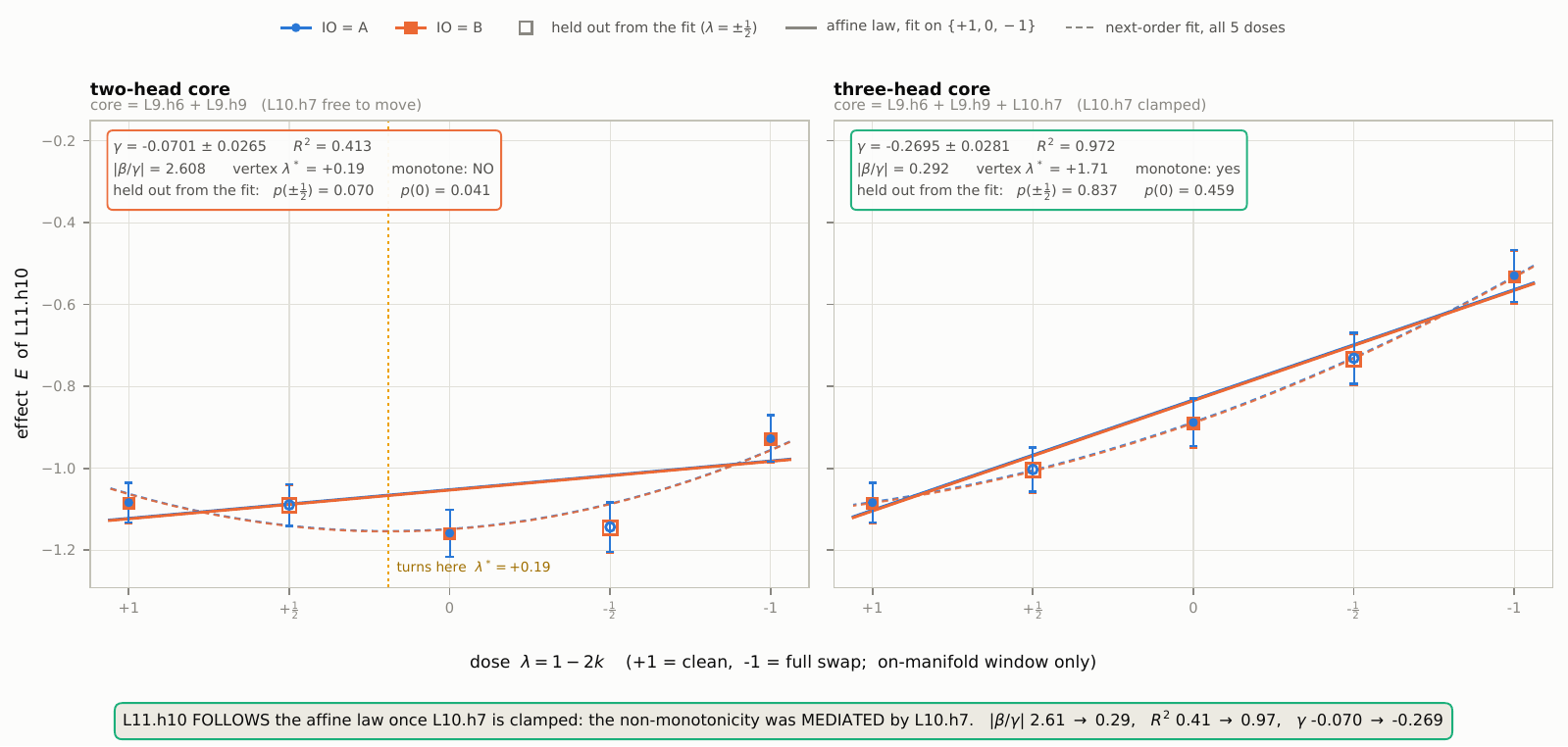}
\caption{\textbf{\texttt{L11.h10}'s exception is a missing core term.} Left,
under the two-head core the on-manifold response {falls and then rises,
with $R^2 = 0.413$, $|\beta_r/\gamma_r| = 2.608$ and vertex at} $\lambda^{*} = +0.19$.
Right, with \texttt{L10.h7} clamped into the core {the response is a line,
with $R^2 = 0.972$, $|\beta_r/\gamma_r| = 0.292$, vertex at $+1.71$ and held-out points}
no longer rejected ($p = 0.837$ and $0.459$, against $0.070$ and $0.041$). The
coefficient nearly quadruples, from $-0.0701$ to $-0.2695$. Filled markers are
{the three fitted doses, and open markers} at $\lambda = \pm\tfrac12$ are held out.
The clean effects $E^{0}$ are identical between the two ladders.}
\label{fig:ioi-l11h10-law}
\end{figure}

{\texttt{L11.h10}, the second of the two negative name movers, is the one
reachable head with a curved response. Under the two-head core its response
falls and then rises inside the window, with $R^2 = 0.413$, a quadratic term
larger than the linear one, $|\beta_r/\gamma_r| = 2.608$, and a parabola that
turns at} $\lambda^{*} = +0.19$ (Figure~\ref{fig:ioi-l11h10-law},
left). For the other nine reachable heads the ratio is below one.

The curvature has a mechanism. \texttt{L11.h10} reads the core largely through
\texttt{L10.h7}, the other negative name mover, which is free to move under the
two-head swap and is itself strongly curved. If that is the cause, clamping
\texttt{L10.h7} into the core should remove it. It does
{(Figure~\ref{fig:ioi-l11h10-law}, right). The ratio} falls from $2.608$ to
$0.292$, $R^2$ rises from $0.413$ to $0.972$, the vertex moves from $+0.19$ to
$+1.71$, outside the window, the raw means become monotone, and the two
held-out points, at $p = 0.070$ and $p = 0.041$ before, are at $p = 0.837$ and
$p = 0.459$ after. The coefficient itself nearly quadruples, from $-0.0701$ to
{$-0.2695$. While \texttt{L10.h7} was free, it had been absorbing} most of
\texttt{L11.h10}'s response to the core, and holding it at the counterfactual
releases that response as a line. The change is specific to \texttt{L11.h10},
since every other shared recruit moves by a factor between $0.73$ and $1.08$,
and the two ladders share the same clean effects $E^{0}$. {The exception
therefore traces to a term missing from the core, and the law holds once the
mediator is held.}

\subsection{The coupling is visible in the weights}

\begin{figure}[!htbp]
\centering
\includegraphics[width=\textwidth]{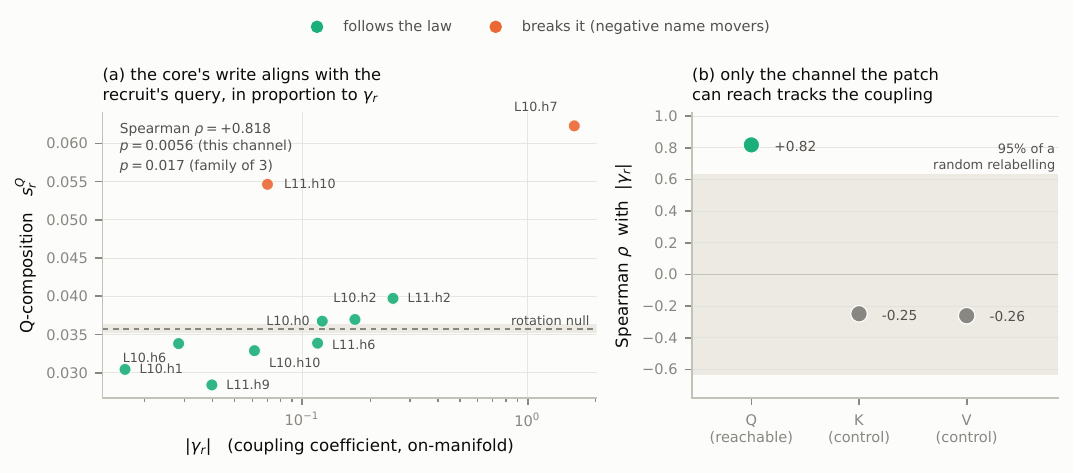}
\caption{\textbf{The coupling is legible in the query weights alone.}
Alignment between the core's write direction and each reachable head's read
rows, computed from the checkpoint with no prompts and no forward pass,
against the measured $|\gamma_r|$. The query channel carries the association
($\rho = +0.818$, {$p = 0.0056$, rising to $+0.917$, $p = 0.0012$, with} \texttt{L11.h10}
excluded as mediated), while the key and value channels are null ($-0.249$,
$-0.261$). Family-wise $p$ across the three channels is $0.0170$.}
\label{fig:ioi-wiring}
\end{figure}

Because the components on this bench are attention heads, the read side of
each recruit decomposes into query, key and value channels, and the weights
can be asked which of them carries the coupling. For each of the ten reachable
heads we compute, from the checkpoint alone, the alignment between the core's
write direction and the recruit's read rows on each channel, and correlate it
against the measured $|\gamma_r|$. The query channel carries the association,
at a Spearman correlation of $+0.818$ with $p = 0.0056$, while the key and
value channels are null at $-0.249$ and {$-0.261$. The family-wise} $p$ across
the three channels is $0.0170$. Excluding \texttt{L11.h10}, which the previous
subsection shows to be reading through a mediator rather than directly, raises
the query correlation to $+0.917$ with $p = 0.0012$
(Figure~\ref{fig:ioi-wiring}).

The informative part is the channel. The core is a pair of name movers, and
what a downstream head inherits from them is where to attend, which is a
{query-side quantity. The query channel carrying the whole correlation while
the other two carry nothing is the pattern the mechanism predicts. With ten
recruits this is a correspondence well above chance rather than a predictor.}

\subsection{Where ablation falls on the dose axis}

The dose axis absorbs ablation of the core as a special case. A clamp
$a := v$ on the core sits at $\lambda(v) = 2(m - v)/\Delta$, with
$m = (a^{T} + a^{F})/2$ and $\Delta = a^{F} - a^{T}$, {so mean ablation amounts to
a distribution of doses, one per prompt pair.}

\begin{figure}[!htbp]
\centering
\includegraphics[width=\textwidth]{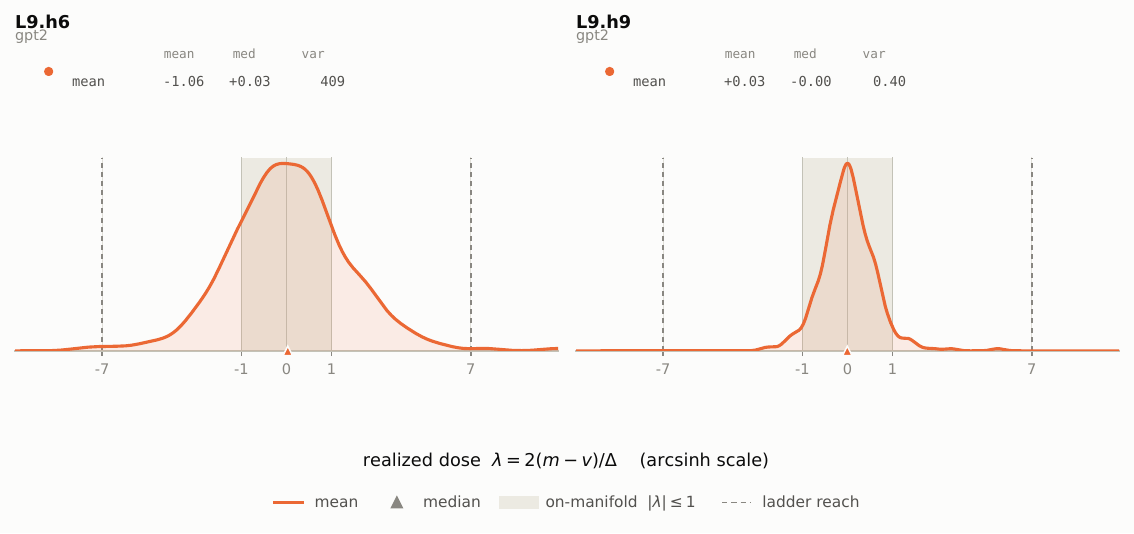}
\caption{\textbf{Mean ablation of the core, placed on the dose axis.} Each
prompt pair puts the clamp at its own dose $\lambda(v) = 2(m - v)/\Delta$.
Both core name movers centre on the neutral dose, with median realised doses
of $+0.03$ (\texttt{L9.h6}) and {$-0.00$ (\texttt{L9.h9}). \texttt{L9.h9}}
places $8\%$ of pairs outside the on-manifold window (shaded) and
\texttt{L9.h6} $40\%$. The realised dose is a ratio, so the median rather than
the mean is the reliable summary.}
\label{fig:ioi-core-mean-dose}
\end{figure}

For the two core name movers that distribution is centred on the neutral dose,
as the balanced design requires, with median realised doses of $+0.03$ for
\texttt{L9.h6} and $-0.00$ for \texttt{L9.h9}. {They differ in spread.
Only} $8\%$ of pairs place \texttt{L9.h9} outside the on-manifold window,
against $40\%$ for \texttt{L9.h6} (Figure~\ref{fig:ioi-core-mean-dose}).
Mean-ablating the core is therefore, on average, the neutral dose, but for a
substantial fraction of prompts it lands where the law no longer applies, and
the fraction depends on which head is ablated. This is the same reason the
factual ladders are read on the dose axis rather than under deletion.

\subsection{Limitations}

Three qualifications apply. {The placement control leans on a single dose. The}
$\lambda = +1$ rung carries over a third of the fit weight for the reachable
heads, and refitting without it is enough for one unreachable head,
\texttt{L8.h10}, to begin certifying while two reachable heads stop. Six of
the ten unreachable heads never move, so the honest separation between the two
groups is the eightfold figure above, not the three orders of magnitude the
raw medians suggest. And the held-out doses interpolate rather than
extrapolate, so they establish smoothness, not the affine form. The run also
{exists under a second reference convention, the one the factual ladders use.
There all} ten reachable heads certify and the held-out prediction is looser,
with a median $|z|$ of $0.41$. We report the convention the figures were built
from, and the qualitative conclusions hold under both.

Finally, the bench is small. Ten heads in one model on one task can test
whether the coefficient behaves as the law says, not how far it generalises.
Within those limits, on a model, task and component set chosen by other
authors, the coefficient resolves where the intervention can reach and
vanishes where it cannot, is largest on the head prior work names as the
circuit's negative feedback element, and its one failure becomes affine once
the mediating head is clamped with the core.

\FloatBarrier

\section{{The Wiring Statistic as a Weight-Space Predictor of Coupling}}
\label{app:wiring}

\paragraph{Definition of the wiring alignment.}
Let $w_c = g^{\mathrm{post}}_{l_c}\odot W_{\mathrm{down}}^{(l_c)}[:,j_c]$ be the residual-stream write of verdict core $c$, where $g^{\mathrm{post}}$ is the post-feed-forward RMSNorm gain (Gemma-2 only, and $1$ elsewhere). For a direction $r$ at layer $l_r$, let $g$ be the gain of the RMSNorm at the input of the block $r$ reads from, and let $\mathcal{R}_r$ be its read rows in residual coordinates with $g$ folded in. For an MLP neuron $j$ these are $g\odot W_{\mathrm{gate}}[j,:]$ and $g\odot W_{\mathrm{up}}[j,:]$. For an MLP-SVD mode $k$ they are $g\odot(u_k^{\top}W_{\mathrm{gate}})$ and $g\odot(u_k^{\top}W_{\mathrm{up}})$, with $u_k=\sigma_k v_k$ the $k$-th scaled right singular vector of $W_{\mathrm{down}}$, a first-order surrogate. For OV channel $c$ of head $h$ the read row is $g\odot W_V[\kappa(h)\,d_{\mathrm{head}}+c,:]$, with $\kappa(h)$ the grouped-query value head, and for OV-SVD mode $k$ of head $h$ it is $g\odot\sigma_k\rho_k$, with $\rho_k$ the $k$-th right singular vector of $W_O^{(h)}W_V^{(\kappa(h))}$. With $\mathcal{C}_r=\{c \mid l_c<l_r\}$ the cores that write strictly before $r$ reads,
\[
a_r=\max_{c\in\mathcal{C}_r}\ \max_{v\in\mathcal{R}_r}\frac{|\langle v,w_c\rangle|}{\lVert v\rVert\,\lVert w_c\rVert},
\qquad
z_r=\frac{a_r-\hat\mu_r}{\hat\sigma_r},
\]
where $\hat\mu_r$ and $\hat\sigma_r$ are the mean and population standard deviation of the same statistic over $N=2{,}000$ random readers, drawn with replacement from the same layer for each direction and scored against the same $\mathcal{C}_r$. For MLP families a reader is a uniformly random neuron (both its gate and up rows), and for OV families it is a uniformly random $W_V$ row. For single-unit directions the direction's own row is excluded. For the two SVD families ($11$ of the $81$ correlated directions) the random reader is a single neuron or $W_V$ row, so their $z_r$ is less well calibrated. $z_r$ is a standardized score against an empirical, non-Gaussian null, and we do not convert it to a $p$-value. Directions with $\mathcal{C}_r=\emptyset$, the placement controls of Table~\ref{tab:upstream-controls}, are scored against all cores and excluded from every correlation. Weights are loaded in bfloat16 and cosines computed in float64, and the null uses \texttt{numpy.random.default\_rng(0)}. Using the raw $a_r$ in place of $z_r$ changes each per-model $\rho$ by at most $0.015$. Table~\ref{tab:onehop-perm} lists the per-model tests behind the Fisher combination of \S\ref{sec:coupling_mag_ground_in_weights}.

For the GPT-2 IOI arm the read side of an attention head is a map rather than a vector, so Figure~\ref{fig:ioi-wiring} uses the query-composition score $\lVert W_O^{\mathrm{core}}W_Q^{r}W_K^{r\top}\rVert_F / (\lVert W_O^{\mathrm{core}}\rVert_F\,\lVert W_Q^{r}W_K^{r\top}\rVert_F)$ and correlates the raw score with $|\gamma_r|$. A $z$-score against $200$ random orthogonal rotations of $W_O^{\mathrm{core}}$ gives the same $\rho=+0.818$.

\begin{table}[!htbp]
\centering
\small
\caption{\textbf{Per-model single-hop test.} Spearman $\rho$ between the weights-only alignment $z_r$ and $|\gamma_r|$ over downstream directions, with a one-sided within-model permutation $p$ ($H_1\colon \rho>0$, $20{,}000$ random re-pairings, $p=(k+1)/(N+1)$, floor $5.0\times10^{-5}$). These four values enter the Fisher combination $\chi^2_8=63.3$, $p\approx1\times10^{-10}$, an upper bound because Gemma sits at the floor.}
\label{tab:onehop-perm}
\begin{tabular}{lrrr}
\toprule
Model & $n$ & $\rho(z_r,|\gamma_r|)$ & one-sided perm.\ $p$ \\
\midrule
Llama-3-8B      & 20 & $0.555$ & $0.006$ \\
Gemma-2-9B      & 22 & $0.823$ & $5.0\times10^{-5}$ (floor) \\
Qwen2.5-7B      & 16 & $0.815$ & $2.5\times10^{-4}$ \\
Mistral-7B-v0.3 & 23 & $0.729$ & $2.5\times10^{-4}$ \\
\bottomrule
\end{tabular}
\end{table}

\subsection{Family breakdown}
\label{app:wiring-family}
{Table~\ref{tab:family-breakdown} splits the certified directions by component family.}

\begin{table}[!htbp]
\centering
\small
\caption{\textbf{The coupling law resolved by component family.} The $68$ certified downstream directions {span four families beyond MLP neurons. For each family the columns give the number benched, downstream and certified, the median weighted $R^2$ of the three-rung fit, the median $|\gamma_r|$, and the split into counterweights ($\gamma_r<0$) and relays ($\gamma_r>0$). The last column is the family-agnostic robustness check, giving the median slope shift when the two half rungs are added and the number of certified directions that keep their sign.}}
\label{tab:family-breakdown}
\resizebox{\textwidth}{!}{\begin{tabular}{llrrrrrrr}
\toprule
& & \multicolumn{3}{c}{count} & & & & \\
\cmidrule(lr){3-5}
Family & & bench & down & cert. & $\widetilde{R}^2$ & $\widetilde{|\gamma|}$ & cw\,/\,relay & $\widetilde{|\Delta\gamma|}/\mathrm{SE}$ (sign) \\
\midrule
\texttt{mlp\_neuron} & MLP neuron & 67 & 54 & 49 & $0.758$ & $0.0286$ & 35\,/\,14 & $0.39$ (49/49) \\
\texttt{ov\_neuron} & OV channel & 16 & 16 & 10 & $0.621$ & $0.0023$ & 10\,/\,0 & $0.29$ (10/10) \\
\texttt{ov\_svd} & OV SVD & 11 & 10 & 8 & $0.698$ & $0.0091$ & 6\,/\,2 & $0.30$ (8/8) \\
\texttt{mlp\_svd} & MLP SVD & 3 & 1 & 1 & $0.680$ & $0.0016$ & 1\,/\,0 & $0.07$ (1/1) \\
\midrule
\textbf{All} & & 97 & 81 & 68 & \multicolumn{4}{l}{sign preserved on all 68 certified directions} \\
\bottomrule
\end{tabular}}
\end{table}

\subsection{{The placement control on directions the core cannot reach}}
\label{app:wiring-placement}
{Tables~\ref{tab:upstream} and~\ref{tab:upstream-controls} score the directions the core cannot reach.}

\begin{table}[!htbp]
\centering
\small
\caption{\textbf{Alignment without causal reach.} Bench directions that sit at or above the layer of every certified core, so that no core writes strictly before them. That their coefficients do not move under the clamp ($\Delta a$ rel.\ $=0$ exactly) is layerwise causality rather than a finding, {and verifies the pipeline. The control lies} in the alignment column. Two of these directions are among the most strongly wired in the study {(}\texttt{L22.n4538} at $z={+}14.1$ and \texttt{L19.n11056} at $z={+}10.1$, above all but a handful of the downstream directions of Table~\ref{tab:coupling}{),} and neither responds. {Llama and Gemma contribute no rows, because their benches lie below the cores.}}
\label{tab:upstream}
\begin{tabular}{llccrcr}
\toprule
direction & family & layer & clean & $z$ & $\Delta a$ rel. & $|R|$ \\
\midrule
\multicolumn{7}{l}{\textsc{Qwen2.5-7B-Instruct} \quad\small cores at L22, L24}\\[1pt]
\texttt{L22.n4538} & mlp neuron & 22 & anti & $+14.06$ & $0.0000$ & $0.0063$ \\
\texttt{L22.n9609} & mlp neuron & 22 & contribute & $+6.55$ & $0.0000$ & $0.2087$ \\
\texttt{L20.h1.svd6} & ov svd & 20 & contribute & $+4.19$ & $0.0000$ & $0.1968$ \\
\texttt{L22.n13025} & mlp neuron & 22 & anti & $+2.54$ & $0.0000$ & $0.0381$ \\
\texttt{L11.svd690} & mlp svd & 11 & null & $-0.10$ & $0.0000$ & $0.0047$ \\
\texttt{L21.svd2451} & mlp svd & 21 & mixed & $-1.05$ & $0.0000$ & $0.0082$ \\
\addlinespace[2pt]
\multicolumn{7}{l}{\textsc{Mistral-7B-Instruct-v0.3} \quad\small cores at L19, L20}\\[1pt]
\texttt{L19.n11056} & mlp neuron & 19 & mixed & $+10.12$ & $0.0000$ & $0.0017$ \\
\texttt{L16.n1170} & mlp neuron & 16 & contribute & $+3.08$ & $0.0000$ & $0.0599$ \\
\texttt{L16.n9991} & mlp neuron & 16 & null & $+2.48$ & $0.0000$ & $0.0007$ \\
\texttt{L18.n10672} & mlp neuron & 18 & null & $+2.38$ & $0.0000$ & $0.0009$ \\
\texttt{L15.n11835} & mlp neuron & 15 & contribute & $+1.54$ & $0.0000$ & $0.0199$ \\
\texttt{L16.n5090} & mlp neuron & 16 & null & $+1.28$ & $0.0000$ & $0.0010$ \\
\texttt{L17.n11073} & mlp neuron & 17 & null & $+1.08$ & $0.0000$ & $0.0002$ \\
\texttt{L17.n8203} & mlp neuron & 17 & null & $+0.25$ & $0.0000$ & $0.0013$ \\
\texttt{L18.n8574} & mlp neuron & 18 & mixed & $+0.22$ & $0.0000$ & $0.0003$ \\
\texttt{L17.n13852} & mlp neuron & 17 & null & $-0.22$ & $0.0000$ & $0.0009$ \\
\bottomrule\end{tabular}\end{table}

\begin{table}[!htbp]
\centering
\small
\caption{\textbf{The placement control, scored against the fitted trend.} The same $16$ directions as Table~\ref{tab:upstream}, placed against its own model's {downstream trend $\log_{10}|\gamma_r| = a + b\,z_r$. \emph{Fold} is} how far below that line it falls, so $>\!1$ means less coupling than its alignment alone would buy. 13 of 16 fall below the line on the three-point fit the coupling law uses (sign test $p = 0.011$) and 13 of 16 on the full-ladder fit. Qwen \texttt{L22.n9609} is the only direction that {clears its trend under both fits. It is a strong verdict unit that sits beside the core rather than after it.}}
\label{tab:upstream-controls}
\setlength{\tabcolsep}{5pt}
\begin{tabular}{llrrrrrr}
\toprule
& & & \multicolumn{2}{c}{three-point fit} & \multicolumn{2}{c}{full ladder} & \\
\cmidrule(lr){4-5}\cmidrule(lr){6-7}
Model & Direction & $z_r$ & $|\gamma_r|$ & fold & $|\gamma_r|$ & fold & $\Delta a_r/|a_r|$ \\
\midrule
Qwen2.5-7B & \texttt{L22.n4538} & $+14.06$ & $0.00335$ & $67.7$ & $0.00066$ & $285.0$ & $0.0000$ \\
Qwen2.5-7B & \texttt{L22.n9609} & $+6.55$ & $0.05401$ & $0.26$ & $0.01732$ & $0.73$ & $0.0000$ \\
Qwen2.5-7B & \texttt{L20.h1.svd6} & $+4.19$ & $0.00397$ & $1.5$ & $0.01404$ & $0.39$ & $0.0000$ \\
Qwen2.5-7B & \texttt{L22.n13025} & $+2.54$ & $0.00526$ & $0.62$ & $0.00270$ & $1.1$ & $0.0000$ \\
Qwen2.5-7B & \texttt{L11.svd690} & $-0.10$ & $0.00129$ & $0.95$ & $0.00017$ & $7.0$ & $0.0000$ \\
Qwen2.5-7B & \texttt{L21.svd2451} & $-1.05$ & $0.00070$ & $1.2$ & $0.00059$ & $1.4$ & $0.0000$ \\
Mistral-7B-v0.3 & \texttt{L19.n11056} & $+10.12$ & $0.00020$ & $165.8$ & $3.9\!\times\!10^{-5}$ & $685.2$ & $0.0000$ \\
Mistral-7B-v0.3 & \texttt{L16.n1170} & $+3.08$ & $0.00195$ & $2.2$ & $0.00368$ & $0.67$ & $0.0000$ \\
Mistral-7B-v0.3 & \texttt{L16.n9991} & $+2.48$ & $0.00059$ & $6.3$ & $1.0\!\times\!10^{-5}$ & $199.3$ & $0.0000$ \\
Mistral-7B-v0.3 & \texttt{L18.n10672} & $+2.38$ & $0.00043$ & $8.2$ & $5.5\!\times\!10^{-5}$ & $35.0$ & $0.0000$ \\
Mistral-7B-v0.3 & \texttt{L15.n11835} & $+1.54$ & $0.00102$ & $2.8$ & $0.00122$ & $1.2$ & $0.0000$ \\
Mistral-7B-v0.3 & \texttt{L16.n5090} & $+1.28$ & $5.6\!\times\!10^{-5}$ & $46.6$ & $0.00010$ & $13.3$ & $0.0000$ \\
Mistral-7B-v0.3 & \texttt{L17.n11073} & $+1.08$ & $0.00029$ & $8.5$ & $2.5\!\times\!10^{-5}$ & $50.1$ & $0.0000$ \\
Mistral-7B-v0.3 & \texttt{L17.n8203} & $+0.25$ & $0.00029$ & $6.6$ & $1.8\!\times\!10^{-5}$ & $53.4$ & $0.0000$ \\
Mistral-7B-v0.3 & \texttt{L18.n8574} & $+0.22$ & $0.00035$ & $5.5$ & $4.0\!\times\!10^{-6}$ & $236.3$ & $0.0000$ \\
Mistral-7B-v0.3 & \texttt{L17.n13852} & $-0.22$ & $0.00010$ & $16.5$ & $2.1\!\times\!10^{-5}$ & $38.0$ & $0.0000$ \\
\bottomrule
\end{tabular}
\end{table}

\subsection{What the wiring statistic does and does not predict}
\label{app:wiring-limits}

The correlation between $z_r$ and $|\gamma_r|$ (Figure~{\ref{fig:weight-visibility}},
Spearman $\rho\in[0.55,0.82]$ per family, Fisher-combined
$p\approx10^{-10}$) licenses a bounded claim. Four limits apply.

\paragraph{Ordinal.} $z_r$ ranks $|\gamma_r|$ {across directions but does not
predict} its numerical value. The relationship is monotone, not calibrated.

\paragraph{Unsigned.} The statistic is built from alignment magnitudes and
carries no sign. Whether a direction is a counterweight ($\gamma_r<0$) or a
relay ($\gamma_r>0$) {must be measured by intervention. The checkpoint
}anticipates how strongly $r$ is coupled to the core, not in which direction.

\paragraph{Causal reach.} Directions placed before the core in the forward
pass cannot read its write and are structurally excluded (hollow markers in
{Figure~\ref{fig:weight-visibility} and §\ref{app:wiring-placement})}. The correlation is a
statement about downstream directions only.

\paragraph{Direct pathway only.} $z_r$ scores the alignment of the core's
write vector with $r$'s read rows, i.e.\ coupling through the value pathway.
Coupling routed through attention patterns is invisible to it.

\paragraph{Entanglement with the clean effect.} $z_r$ correlates with the
clean-run causal magnitude $|E_r^{0}|$ ($\rho\in[0.68,0.86]$) about as
strongly as with $|\gamma_r|$. On this bench {the two cannot be separated, since
under} Eq.~{\ref{eq:affine_law}}, $E_r^{0}=\mathrm{own}_r+\gamma_r$ at $\lambda=+1$,
so any statistic that tracks one tracks the other. Refitting $\gamma_r$ on the
intervened rungs alone, $\lambda\in\{0,-1\}$, removes the clean run from the
fit ($\rho\in[0.43,0.83]$, combined $p<10^{-9}$) but not the boundary
condition. We therefore do not claim that the weights encode the slope
$\gamma_r$ in isolation. The claim we do make is that the static checkpoint
anticipates the magnitude of the model's causal coupling, and that
$\gamma_r$ consequently indexes a property of the trained network rather
than of the intervention protocol.
\FloatBarrier

\section{{Derivations Showing That Every Clamp Is a Dose}}
\label{app:deletion}
\label{app:realised}
Throughout, a pair $p$ has clean values $a_T, a_F$ with
$m = \tfrac12(a_T + a_F)$ and $\Delta = a_F - a_T$ on the true member. On the
false member $\Delta$ changes sign and $m$ does not. On the presented member
the clean value sits at $\lambda = +1$, the partner's at $-1$, and the
midpoint at $0$. Main text Eq.~\ref{eq:law} is assumed on the on-manifold window.

\paragraph{Any clamp is a dose.}
The dosed swap sets $a_c := m - \tfrac{\lambda}{2}\Delta$. Read the other way,
a clamp $a_c := v$ is the dose
\begin{equation}
  \lambda(v) = \frac{2(m - v)}{\Delta},
  \label{eq:clampdose}
\end{equation}
and by Eq.~\ref{eq:law} a downstream direction contributes
$E_r(v) = \mathrm{own}_r + \gamma_r\,\lambda(v)$. This is a change of
variable, not a new claim. Every deletion method is a choice of $v$, and
Eq.~\ref{eq:clampdose} says where that choice lands on the axis.

\paragraph{Where each deletion lands.}
Each deletion method clamps $a_c := v$ with $v$ drawn from a distribution
$D$ with mean $\mu_D$ and standard deviation $\sigma_D$. A fixed value is a
distribution with $\sigma_D = 0$. By Eq.~\ref{eq:clampdose}, the dose is
\begin{equation}
  \lambda(v) = \frac{2(m - v)}{\Delta}, \quad v \sim D,
  \qquad
  \mathbb{E}[\lambda] = \frac{2(m - \mu_D)}{\Delta},
  \qquad
  \mathrm{sd}[\lambda] = \frac{2\sigma_D}{|\Delta|} =: \rho_D .
  \label{eq:randomdose}
\end{equation}
The mean dose is inside the window when $\mu_D$ lies between $a_T$ and
$a_F$. Each draw moves a direction's contribution by $|\gamma_r|\rho_D$
around its mean.
\begin{itemize}
  \item \emph{Zero ablation} ($D$ is a point mass at $0$) lands at
  $\lambda = 2m/\Delta$. This is neutral only when the pair's midpoint $m$
  is zero, and task-specialised units rarely rest at zero
  (Figure~\ref{fig:deletion_dose_a}).
  \item \emph{Mean ablation} ($D$ is a point mass at the reference-set mean
  $\bar a$) lands at $2(m - \bar a)/\Delta$. With a fraction $q$ of the
  presented class in the reference set, it averages $2q - 1$ over pairs,
  which is $0$ for a balanced set. On any single pair, though, it is large
  whenever the pair's contrast $\Delta$ is small. In our data it reaches
  $|\lambda| = 5.1$, beyond any dose the ladder tests.
  \item \emph{Resample ablation} ($D$ is the donor pool's activation
  distribution) lands wherever the pool puts it. In a task-matched,
  class-balanced pool, {$D$ has two modes. A same-class donor} gives a dose
  near $+1$ (no change) and an opposite-class donor gives a dose near $-1$
  (a full swap). The dose is then a coin flip between the two ends of the
  window, {with variance at least $1$, and we measure $1.35$ to $2.48$ across}
  cores. In an open-distribution pool, $\mu_D$ is the unit's generic-text
  mean $b$. The average dose is then $2(m - b)/\Delta$, which is zero
  ablation plus noise when $b \approx 0$.
\end{itemize}

\section{Beyond a Binary Contrast}
\label{app:general}

{The results of \S3 to \S7 use one property of the task. Its class signal} on the
core is one-dimensional, so every clamp is a single dose that every
downstream direction agrees on. Under the linear model of Eq.~{\ref{eq:law}}, the
following hold for any number of classes and any number of core directions.
Proofs are two lines each and are omitted.

\begin{proposition}[Dose]
Any clamp of the core is still {a dose for each downstream direction. That
direction's} contribution moves linearly with how far the clamp displaces the
core along the axis the direction reads. With more than two classes the dose
{is specific to the reader, and with two} it is the shared $\lambda$ of \S3.
\end{proposition}

\begin{proposition}[Constant clamps]
A clamp that ignores the prompt's class (zero, mean, resample, fixed-donor
patching, fixed-strength steering) leaves every downstream direction with the
same core-derived input on every class. It can therefore remove the
class-differential part of the direction's contribution and shift its
common level, but it cannot reverse that part {or create a new one. The only
}class information left in the direction is what it gathers by other routes.
\end{proposition}

\begin{proposition}[Release]
Consequently, under any constant clamp a counterweight's contribution rises
by an amount proportional to its coupling and to the distance of the clamp
from the clean point, and a relay's falls by the same. Directions rank by
$|\gamma_r|$, a quantity read from the checkpoint before any ablation, and
draw-to-draw variation under a random donor is the coupling times the
donor's spread rather than noise. Reversal requires an intervention that
moves each class separately, as the dosed swap of Eq.~{\ref{eq:swap}} does.
\end{proposition}

These statements are exact on the on-manifold window, where linearity is
certified (\S5), and are assumptions beyond it. They do not describe a
direction whose clean coupling is near zero and which activates only under
{ablation. That would be genuine gating, and nothing here} rules it out.
Whether the same accounting explains self-repair in settings other than IOI,
the sensitivity of circuit discovery to ablation type, or partial recovery
after refusal or unlearning interventions is left to future work.

\FloatBarrier

\section{{A Conjecture That Every Design Is a Point in the \texorpdfstring{$(u,v)$}{(u,v)} Plane}}
\label{app:conjecture}

\paragraph{The plane.}
Any intervention on a core, followed by an interchange on a downstream
direction, {is described by two doses, $u$ for} the presented prompt and $v$ for
the donor, each in units of its own pair's contrast. Under the linear model
{the readout is $E_r(u,v)=\mathrm{own}_r+\gamma_r\,\tfrac{u+v}{2}$, a function} of $(u+v)/2$ alone. The designs in
use are lines in this plane. The driven swap is the diagonal $u = v$, the
clean model its point $(+1,+1)$, and a constant clamp the anti-diagonal
$v = -u$, at a position the data set through $\lambda(v)$. Standard activation
patching with an unpatched donor is the line $v = +1$. Every published
protocol commits to a line and to a position on it, and outside the driven
swap the position is chosen by the activations, not by the experimenter. {This is
the consequence \S\ref{sec:deletion} reports. On the anti-diagonal}
the coupling cancels per class under the interchange readout and in the class
sum under any readout, so a clamp can release standing coupling but cannot
reverse it.

\paragraph{More than two classes.}
With $C$ counterfactual classes the dose becomes barycentric coordinates in
the affine hull of the class-conditional activations $\{a_c\}_{c=1}^{C}$, and
a constant clamp lands at coordinates the data set. The binary {class-sum cancellation of \S\ref{sec:deletion} does not survive verbatim}, because
$\sum_c \lambda_c(v)$ is no longer independent of $v$. A weaker statement
does. After any constant clamp the intervened coordinate has zero
between-class variance, so a linear reader of it loses its {class-differential
component and retains} only a common-mode term proportional to
$\bar a - v$. Common mode is outside the estimand of \S\ref{sec:test}. Hence,
for any $C$, a constant clamp can attenuate a class-differential signal and
shift its common mode, but cannot induce a class-inconsistent signal. This is
{the invariant that generalises. Reversal requires} a class-dependent
intervention, and no constant clamp is one.

\paragraph{Conjecture.}
\emph{Wherever (a)~the intervened coordinate carries a class-differential
signal, (b)~the checkpoint contains downstream directions with standing
negative gain on it, and (c)~the readout is per-class or class-pooled, a
constant-clamp ablation records a rise in those directions that is monotone
in $|\gamma_r|$, and the rise is reported as compensation.} {Condition (b) is
not exotic, since components} that write against the output recur across models
{\citep{mcdougall2023copysuppressioncomprehensivelyunderstanding,gurnee2024universalneuronsgpt2language,stolfo2024confidenceregulationneuronslanguage}},
and the counterweights of this paper are present in base checkpoints
(App.~{\ref{app:pretraining}}). We list settings {in decreasing order of how closely
(a) to (c) are met.} Self-repair under mean and resample ablation
\citep{wang2022interpretability, mcgrath2023hydraeffectemergentselfrepair,
rushing2024explorationsselfrepairlanguagemodels}, {where (a) to (c) hold by}
construction and the IOI case is measured in App.~\ref{app:ioi}. The
sensitivity of automated circuit discovery to ablation type, already noted by
{\citet{conmy2023automatedcircuitdiscoverymechanistic}}, which on this account is the dose varying with the
clamp. Ablation of a refusal direction, a zero clamp of a verdict core, whose
partial recovery elsewhere would be release of standing safety counterweights
(\S{\ref{sec:conclusion}}). Probe-based steering, which scores a counterweight
as a truth feature and steers it the wrong way. And the relearnability of
unlearned behaviour, where a clamp on a knowledge direction leaves every
coupled direction at its resting sum. Only the first has evidence in this
{paper. For the others, the preconditions appear to hold.}

\paragraph{Assumptions, and where the conjecture would fail.}
Linearity is certified on $|\lambda| \le 1$ and {saturates beyond, and the two
}copy-suppression heads of IOI compose sublinearly (App.~\ref{app:ioi}). The
conjecture as stated {needs less than linearity, only a response monotone}
in the class-differential input, which is why we state it in terms of a rise
rather than a coefficient. The wiring statistic scores the direct value
pathway and is blind to coupling routed through attention patterns, so a
compensator on that route would be missed. The LayerNorm rescaling share of
\citet{rushing2024explorationsselfrepairlanguagemodels} is controlled here
(App.~{\ref{app:renorm}}) and not in the settings above. And the conjecture
is falsified in any setting by a compensator whose clean-model coupling is
near zero and which {activates under ablation. That would be genuine gating,
which nothing in this paper rules out elsewhere. Before ablating, the conjecture yields the dose the clamp will}
deliver, from the unit's pair statistics, and a rank order of which directions
will appear to compensate, from the checkpoint.
\clearpage
\FloatBarrier

\bibliographystyle{unsrtnat}
\bibliography{iclr2027_conference}

\end{document}